\documentclass{article} 
\usepackage[preprint]{goodfire/goodfire}

\usepackage{microtype}
\usepackage{hyperref}
\usepackage{url}
\usepackage{booktabs}
\usepackage{array}
\usepackage{graphicx}
\usepackage{ulem}
\usepackage{amsmath}
\usepackage{amssymb}
\usepackage{tikz}
\usepackage{bm}

\usepackage{subcaption}
\usepackage{enumitem}
\usepackage{multirow}
\usetikzlibrary{positioning}
\usetikzlibrary{arrows.meta}

\usepackage{algorithm}
\usepackage{algorithmic}

\newcommand{\inputnumber}{offset}
\newcommand{\inputconcept}{input concept}
\newcommand{\outputconcept}{output concept}

\usepackage{lineno}

\hypersetup{colorlinks=true, citecolor=primary, linkcolor=primary, urlcolor=primary}

\definecolor{primary}{HTML}{7B2D26} 
\definecolor{secondary}{HTML}{B8973A} 

\providecommand{\aut}[1]{\textbf{#1}}
\providecommand{\af}[1]{{\small #1}}

\providecommand{\afn}[1]{\textcolor{primary}{$^{#1}$}}

\newcommand{\costar}{\textcolor{secondary}{\boldsymbol{\star}}}
\newcommand{\colead}{\textcolor{secondary}{\boldsymbol{\dagger}}}

\newcommand{\goodfiremark}{\raisebox{0.5pt}{\hspace{0.5mm}\includegraphics[height=6pt]{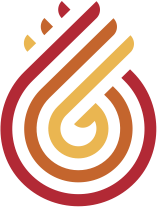}}}

\newcommand{\goodfireaff}{%
  \includegraphics[height=14pt]{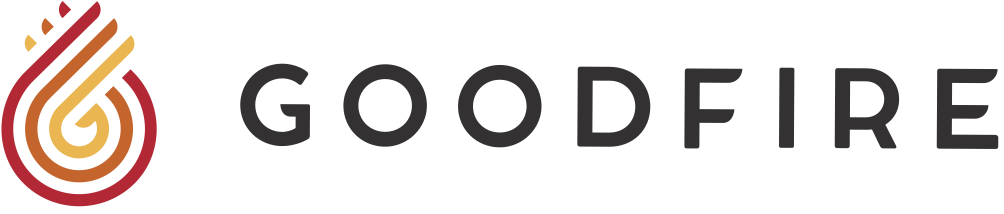}
}

\newcommand{\authorentry}[2]{\aut{#1}\afn{#2}}
\newcommand{\afflabel}[2]{\afn{#1}\af{#2}}
\newcommand{\authorsep}{\quad}
\newcommand{\repolink}[1]{%
  {\small
    \href{#1}{\raisebox{-2.8pt}{\includegraphics[height=10pt]{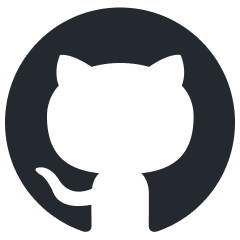}}%
    \texttt{\textcolor{secondary}{#1}}}
  }%
}

\newtoggle{goodfireonly}
\togglefalse{goodfireonly}  
\newcommand{\extline}[1]{\iftoggle{goodfireonly}{}{#1}}
\newcommand{\extaff}[1]{\iftoggle{goodfireonly}{}{#1}}

\newcommand{\paperauthors}{%
\authorentry{Sheridan Feucht}{\costar\goodfiremark\extaff{,a}} \authorsep
\authorentry{Tal Haklay}{\costar\goodfiremark\extaff{,b}} \\
\vspace{2pt}
\authorentry{Usha Bhalla}{\goodfiremark\extaff{,c}} \authorsep
\authorentry{Daniel Wurgaft}{\goodfiremark\extaff{,d}} \authorsep
\authorentry{Can Rager}{\goodfiremark} \authorsep \\
\vspace{2pt}
\authorentry{Raphaël Sarfati}{\goodfiremark} \authorsep
\authorentry{Jack Merullo}{\goodfiremark} \authorsep
\authorentry{Thomas McGrath}{\goodfiremark} \authorsep
\authorentry{Owen Lewis}{\goodfiremark} \\
\vspace{2pt}
\authorentry{Ekdeep Singh Lubana}{\colead\goodfiremark} \authorsep
\authorentry{Thomas Fel}{\colead\goodfiremark} \authorsep
\authorentry{Atticus Geiger}{\colead\goodfiremark}
\vspace{3pt}\\
\textcolor{secondary}{$^{\boldsymbol{\star}}$}\af{Equal contribution}  \authorsep
\vspace{3mm}
\textcolor{secondary}{$^{\boldsymbol{\dagger}}$}\af{Equal senior contribution} \\
\goodfireaff \\
\extline{%
  \afflabel{a}{Northeastern University} \authorsep
  \afflabel{b}{Technion IIT} \authorsep
  \afflabel{c}{Harvard University} \authorsep
  \afflabel{d}{Stanford University}
}
\vspace{2mm}\\
\repolink{~https://github.com/goodfire-ai/arithmetic-wild}
\vspace{-4mm}
}
\author{\paperauthors}

\title{Arithmetic in the Wild: Llama uses Base-10 Addition to Reason About Cyclic Concepts}

\newcommand{\orig}{o}
\newcommand{\counter}{c}
\newcommand{\llama}{Llama-3.1-8B}

\begin{document}

\maketitle

\vspace{-10pt}
\begin{abstract}
Does structure in representations imply structure in computation? We study how \llama\ reasons over cyclic concepts (e.g., \textit{``what month is six months after August?''}). Even though \llama's representations for these concepts are circularly structured, we find that instead of directly computing modular addition in the period of the cyclic concept (e.g., 12 for months), the model re-uses a generic addition mechanism across tasks that operates independently of concept-specific geometry. First, it computes the sum of its two inputs using base-10 addition (\textit{six + August}=14). Then, it maps this sum back to cyclic concept space (14$\rightarrow$February). We show that \llama\ uses task-agnostic Fourier features to compute these sums---in fact, these features have periods that respect standard base-10 addition, e.g., 2, 5, and 10, rather than the cyclic concept period (e.g., 12 for months). Furthermore, we identify a sparse set of 28 MLP neurons re-used across all tasks (approximately 0.2\% of the MLP at layer 18) that can be partitioned into disjoint clusters, each computing the sum for a Fourier feature with a different period. Our work highlights how an interplay between causal abstraction and feature geometry can deepen our mechanistic understanding of LMs.

\end{abstract}
\section{Introduction}

What is the algorithmic role of representation geometry in language models?
We investigate cyclic concepts such as weekdays and months as a case study, which LMs represent using circular geometry
\citep{engelscircles, model2025,karkada2026symmetrylanguagestatisticsshapes, prieto2026correlations, park2025}. 
Because small transformers trained on modular addition operate over a periodic basis~\citep{nanda2023modular, zhong2023pizza,furuta2024interpreting}, which also fits a circular geometry~\citep{morwani2024feature}, a natural hypothesis is that LMs compute answers to questions like ``\texttt{what month is six months after August?}'' using a similar modular arithmetic algorithm.
However, we find that this is not the case: \llama~\citep{grattafiori2024llama}, in fact, uses a base-10 addition mechanism for cyclic tasks, converting the resulting sum back to circular representations in late layers.

\begin{figure}
    \centering
    \includegraphics[width=\linewidth]{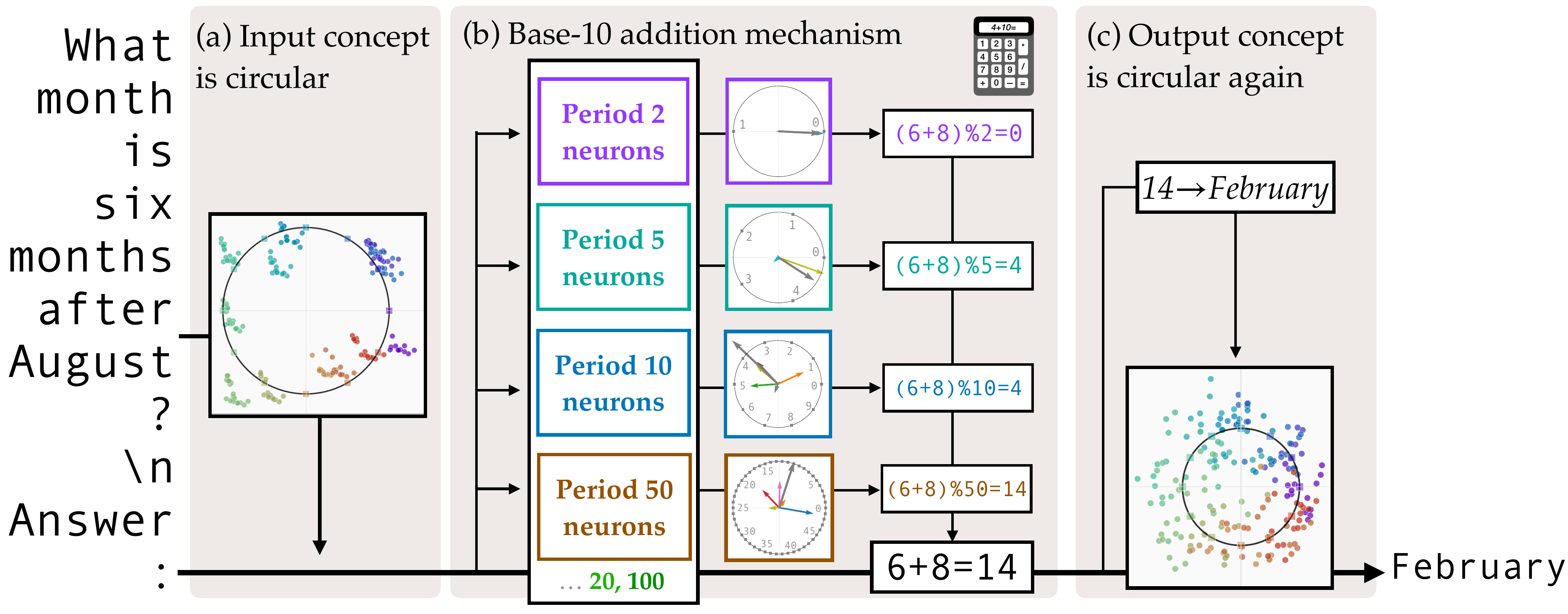}
    \caption{\llama\ calculates \textit{six months after August} with a standard addition mechanism that is used for numbers, months, weekdays, and hours. (a) Cyclic concepts are represented with circular geometry at the input token position \citep{engelscircles}. (b) The model computes addition in a base-10 Fourier number space \citep{kantamnenihelix, fu2026convergentevolutiondifferentlanguage} using the same neurons for all tasks. We discover 28 MLP neurons forming distinct clusters, where each cluster computes the sum for a specific periodicity (circle radii scaled for legibility). (c) This sum is mapped back to cyclic concept space in late layers.}
    \label{fig:figure-1}
\end{figure}

Using causal analysis \citep{vig2020, geiger2025causalabstractionunderpinscomputational, Geiger2026, Mueller2026}, we isolate a base-10 addition mechanism that \llama\ uses to solve cyclic tasks (Figure~\ref{fig:figure-1}). This mechanism, which is re-used for months, weekdays, hours, \textit{and} standard addition, computes a sum in numerical space (e.g., \textit{six + August = 6 + 8 = 14}) that is mapped to concept space in subsequent layers (14$\rightarrow$February). Because this addition mechanism is shared across tasks, we can patch between standard addition prompts (\textit{a+b=}) and cyclic prompts with predictable results.

We analyze these numerical representations by probing for ``Fourier features'', i.e., representations of sinusoidal functions that when composed together, by the Fourier transform, can be used to represent continuous maps~\citep{fourier1807theorie}. Prior work has found that models represent numbers using Fourier features with base-10 periodicity, specifically, $T\in\{2,5,10,20,50,100\}$ \citep{kantamnenihelix, zhoufourier, zhou2025fone, fu2026convergentevolutiondifferentlanguage}, but we show that these features are also used to calculate addition for months, weekdays, and hours. 
We isolate a sparse set of 28 MLP neurons that perform addition across all tasks: these neurons write to the Fourier features we found, and can be partitioned into clusters of neurons~\citep{yanivheuristics, hanagreaterthan, gurnee2023finding} that compute the sum for each period. Our work provides an in-depth understanding of a shared addition mechanism that re-uses the same geometry across all tasks, regardless of the structure of the output domain.

\section{Causal Abstraction over Cyclic Tasks}\label{sec:das}
\paragraph{Task setup.} We focus on months of the year (\textit{what is six months after August?}), weekdays (\textit{what day is three days after Friday?}), and 24-hour time (\textit{It is currently 13:00. What time will it be in four hours?}).
We evaluate \llama\ on every combination of \inputconcept\ and \inputnumber; offsets range from $1$ to $2p$, where $p$ is the cycle length of the concept (e.g., for months, $p$=12).
We also include a standard addition task \textit{a+b=} with $a,b\in \{1, \dots, 100\}$.
Restricted to in-cycle offsets (\inputnumber\ $\leq p$), \llama\ achieves 82\% accuracy on \texttt{months}, 92\% on \texttt{weekdays}, and 97\% on \texttt{hours}.
Within this range, most errors come from prompts where the answer must ``loop around'' to the start of the cycle (e.g., \textit{two months after December}; see App.~\ref{app:performance} for details). We also find that \llama\ is surprisingly unable to perform explicit modular arithmetic for modulo 7, 12, and 24 (App.~\ref{app:explicit-mod}). 

\paragraph{Approach.}
Distributed alignment search (DAS; \citealt{Geiger2023FindingAB}) is an optimization procedure that finds a subspace containing information relevant for LM behavior, represented as an abstract causal variable $V$ in a causal model $\mathcal{A}$. 
We use DAS to localize such variables to low-rank subspaces of the residual stream. Let $d_{\text{model}}$ be the residual stream dimension and $k$ be the dimension of the subspace we want to find, which is a hyperparameter. Then, we learn a low-rank matrix $\mathbf{R} \in \mathbb{R}^{d_{\text{model}} \times k}$ with orthonormal columns 
(with the neural network $\mathcal{N}$ frozen) such that patching within the subspace defined by $\mathbf{R}$ from a counterfactual to an original prompt ($\counter\rightarrow\orig$) causes the model to predict a counterfactual output for $\orig$. The training objective is:

\vspace{-5pt}
\begin{equation}
\label{eq:commute}
\mathsf{CrossEntropy}\bigl(\mathcal{N}_{\mathbf{R} \leftarrow \mathcal{N}(\counter)}(\orig),\mathcal{A}_{V \leftarrow \mathcal{A}(\counter)}(\orig)\bigl),
\end{equation}

where the abstract causal model $\mathcal{A}$ defines our expectation of what should happen when we patch $V$ for any $(\counter, \orig)$. See App.~\ref{app:causal} for full definitions.

For example, we can train DAS on the \texttt{months} task to localize a $k$-dimensional subspace of the residual stream that encodes the \inputconcept\ at a particular layer and token position. We measure success with \textit{interchange intervention accuracy} (IIA), which in this case indicates whether patching within this subspace from e.g., \textit{six months after \textbf{October}}$\rightarrow$\textit{two months after \textbf{April}} causes the model to output \textit{December} (two + \textbf{October}). To patch within the subspace spanned by $\mathbf{R}$, we replace the original hidden state $\mathbf{h}_{\orig}$ with a new state
\begin{equation} \label{eq:das}
\mathbf{h} = \mathbf{h}_\orig + \mathbf{R}(\mathbf{R}^T \mathbf{h}_\counter  - \mathbf{R}^T \mathbf{h}_\orig ),
\end{equation}
where $\mathbf{h}_\counter$ is the hidden state for the counterfactual prompt $\counter$.
We use DAS to localize subspaces for the \inputconcept, \inputnumber, and \outputconcept\ at the last token position\footnote{
We choose the last token position based on initial intervention experiments from App.~\ref{app:interchanges}.}, training on the residual stream after each attention/MLP sublayer. All tasks are trained separately.
We sweep across subspace dimensions $1\leq k\leq8$ for \texttt{months} and \texttt{weekdays}, and $1\leq k\leq16$ for \texttt{hours} and \texttt{addition}, reporting results for the dimension with the best performance (empirically, this is usually the largest $k$). From this point on, we refer to the subspace with the best test IIA as simply ``the DAS subspace'' for a given task and causal variable. See training details in App.~\ref{app:das}. 

\begin{figure}
    \centering
\includegraphics[width=\linewidth]{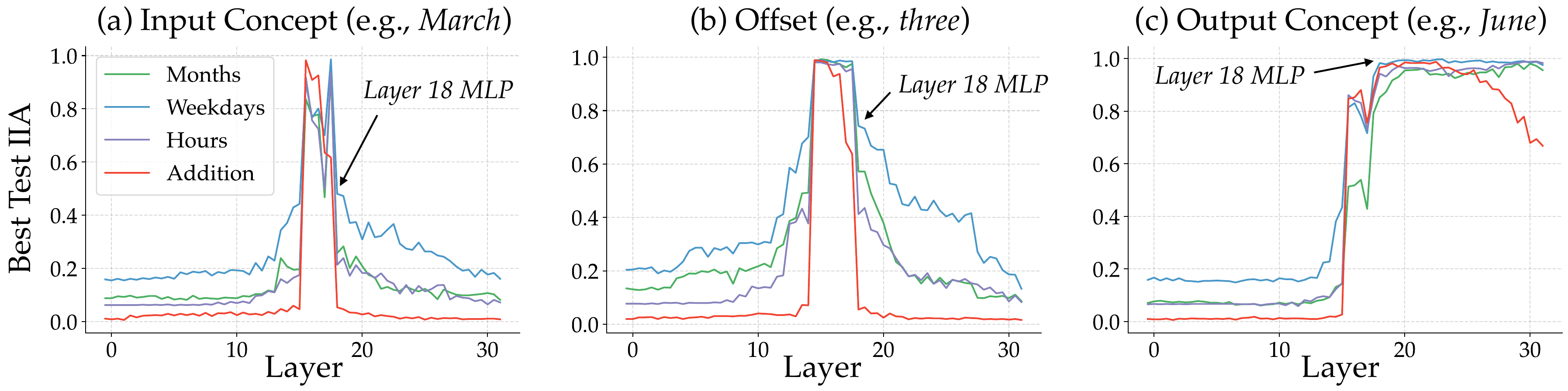}
    \caption{DAS results provide strong evidence that \inputconcept\ and \inputnumber\ information are combined to produce the \outputconcept\ in layer 18 at the last token position. (a) \textbf{Input concept} interchange accuracy (IIA) reaches near-100\% immediately \textit{before} the layer 18 MLP is applied; afterwards, IIA drops, indicating that this information has been ``consumed.'' (b) \textbf{Offset} variables are copied to the last token position at layer 15 before being ``consumed'' after layer 18, when IIAs drop. (c) \textbf{Output concept} information materializes after layer 18, where IIA reaches 100\%. We show the best subspace dimensions for each task ($k=8$ for \texttt{months} and \texttt{weekdays}, and $k=16$ for \texttt{hours} and \texttt{addition}). See App.~\ref{app:das} for full results.}
    \label{fig:main-das-iia}
\end{figure}

\paragraph{Evidence for a shared mechanism.}
Figure~\ref{fig:main-das-iia} shows results for the best subspace for each task across sublayers. Despite DAS being trained separately for each task, results are remarkably similar across domains. For cyclic tasks, \inputconcept\ and \inputnumber\ variables can be causally isolated with $>$95\% accuracy in the input to MLP 18. Immediately \textit{after} this MLP, we cannot cleanly intervene on input arguments, e.g., IIA for input month drops by about 80 points, implying that the model has started to compute the final output. This pattern persists even for the \texttt{addition} task, which is not cyclic. Overlap between subspaces for the same variable across tasks also peaks in similar layers (App.~\ref{app:principal-angles}). \textit{Taken together, these results point to a shared mechanism at layer 18 in the last token position for all tasks.}

\begin{figure}
\centering
\includegraphics[width=\linewidth]{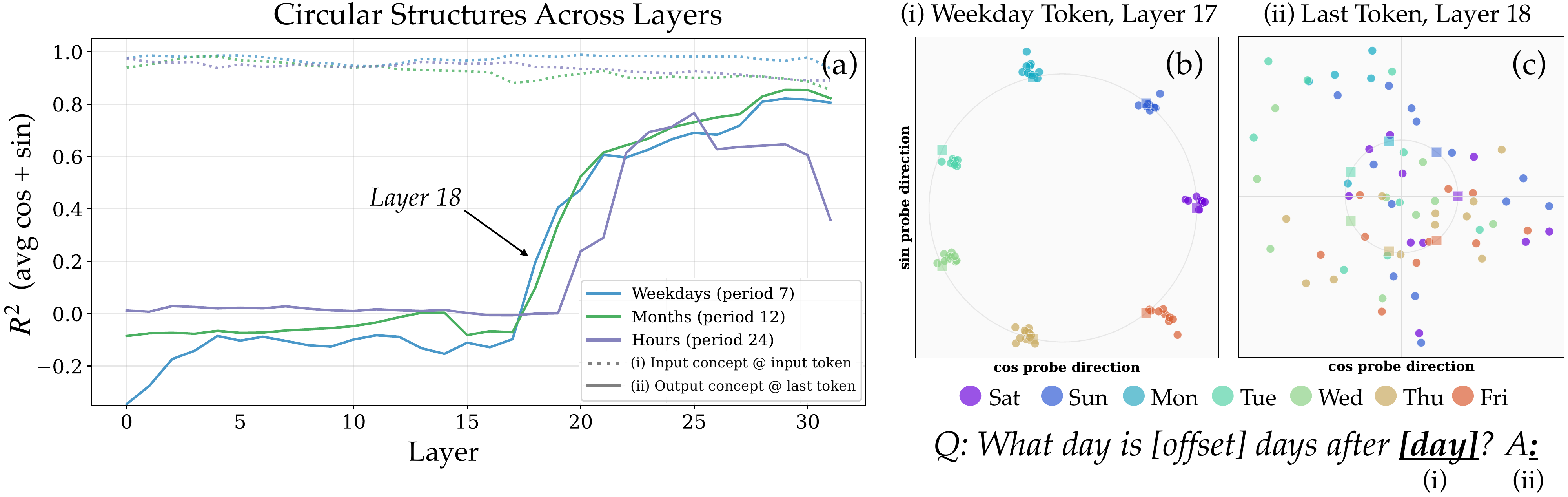}
\caption{Probes reveal that circular structure \citep{engelscircles} is not present in layer 18 at the final token, where the input concept and offset are combined. (a) We train circular probes for (i) the \inputconcept\ at the input concept token position and (ii) the \outputconcept\ at the final token position. Circular structure for (i) is consistently recovered across layers; see (b) for an example of the weekdays circle at layer 17. In contrast, circular structure for (ii) emerges only in later layers. Notably, at layer 18, we do not yet see circular structure for the \outputconcept, despite results in Section~\ref{sec:das} indicating that the inputs have already been combined at this stage (see (c)).
}
\label{fig:natural_circles}
\end{figure}

\paragraph{Where is the circle?}
We replicate the experiment from \cite{engelscircles} and train ``circular probes'' for each task to recover the circular geometry that \llama\ uses to represent cyclic concepts: in particular, we search for these circular structures in a PCA-reduced activation subspace within the top five principal components (see App.~\ref{app:circular_probes} for details). If \llama\ computed e.g., \textit{three days after Wednesday} by somehow rotating along the weekday circle discovered by \cite{engelscircles}, we would expect to find this structure in the layers relevant for computation.
Although we can recover circular structure at the \inputconcept\ token position across layers (Figure~\ref{fig:natural_circles}), we cannot reliably probe for circular structure at the \textit{final} token position until layers 22-25, even though Figure~\ref{fig:main-das-iia}c shows that the \outputconcept\ has started to emerge by layer 18. This suggests that the causal mechanism at layer 18 does \textit{not} rely on task-specific circular geometry.

\section{The Shared Mechanism is Addition}\label{sec:cross-task}

Given the similarity of DAS results between \texttt{addition} and cyclic tasks (Section~\ref{sec:das}), one hypothesis is that \llama\ uses base-10 addition for cyclic tasks.
Under this hypothesis, the \outputconcept\ subspace in a forward pass for \textit{6+8=} should contain the same information as it does for \textit{six months after August}. We show that this is in fact true: we can reliably patch between \texttt{addition} prompts and cyclic tasks with predictable effects, suggesting that cyclic tasks are represented as numbers around layer 18.

\begin{figure}[t]
    \centering
\includegraphics[width=\linewidth]{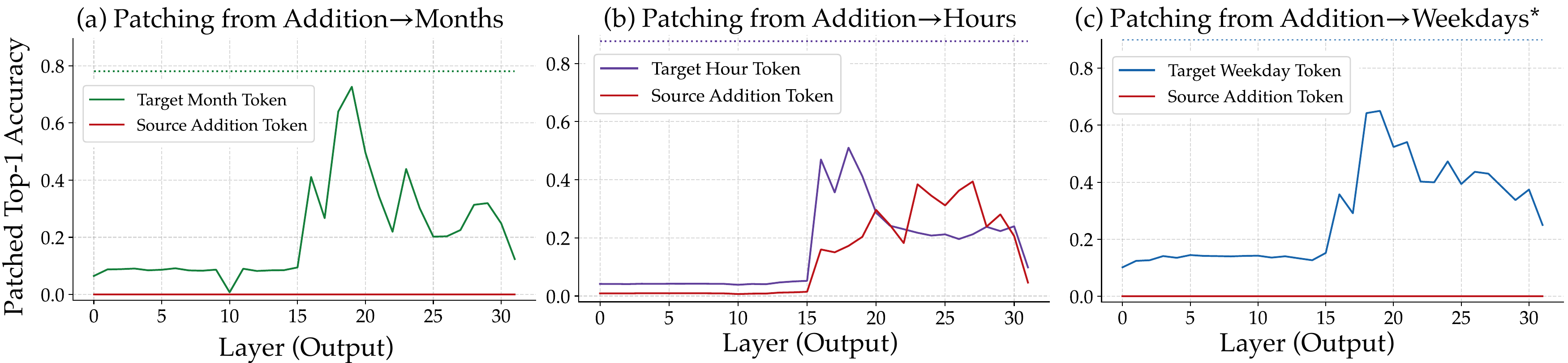}
    \caption{Patching from \texttt{addition} to cyclic tasks within the shared subspace of both tasks shows that base-10 numerical representations at layer 18 are converted into cyclic concepts. (e.g., \textit{6+8=14$\rightarrow$February}).
    Patching at layer 18 consistently causes the model to output the predicted concept. This patch does not cause the model to output the source number token (red line), except for \texttt{hours}, which also uses number tokens. Dotted lines indicate clean model performance for equivalent number ranges. We enumerate weekdays based on results from other experiments; see App.~\ref{app:weekdays}.
   }
    \label{fig:addition-to-everything}
\end{figure}

\paragraph{Patching from addition to cyclic tasks.}
First, we patch from \texttt{addition} into cyclic tasks, with the expectation that the sum from the addition task will be decoded into a concept, e.g., patching from \textit{6+8=} to any \texttt{months} prompt should cause the model to output \textit{February}, as (6+8) mod 12 = 2 and February is the second month. 
At every layer, we patch within the union of the \texttt{addition} and target output DAS subspaces.
Figure~\ref{fig:addition-to-everything} shows that for all three cyclic tasks, \llama\ converts the \texttt{addition} sum to a cyclic concept with at least 40\% accuracy. 
For \texttt{months}, performance is comparable to the model's accuracy in a clean run. 
These results imply that \llama\ uses numerical representations in middle layers for cyclic tasks.

\begin{figure}[t]
    \centering
    \includegraphics[width=\linewidth]{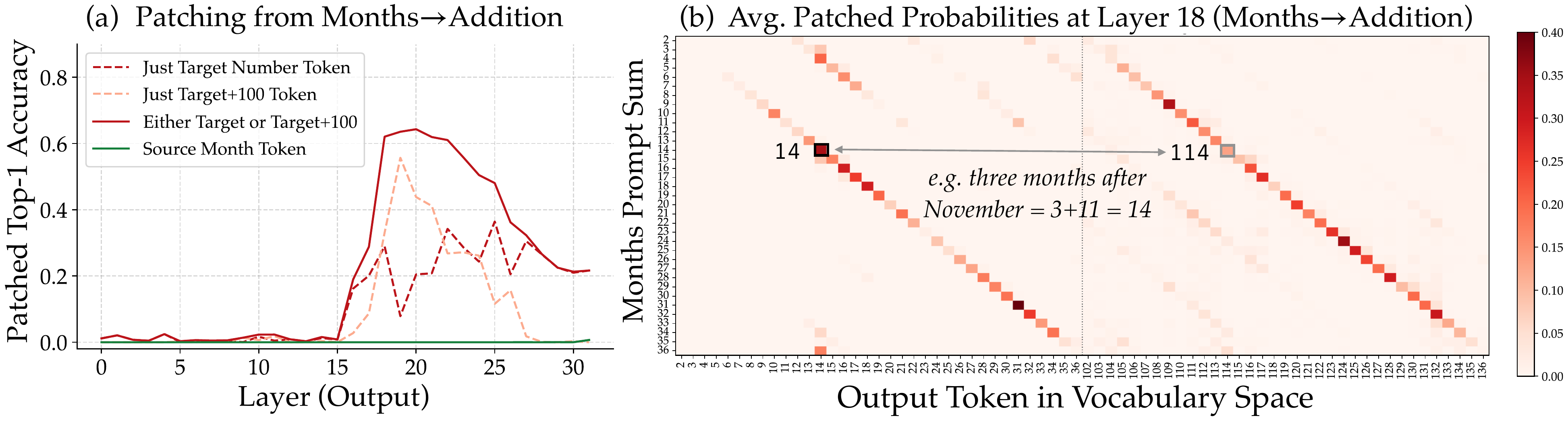}
    \caption{Patching from \texttt{months} $\rightarrow$ \texttt{addition} within the shared subspace of both tasks shows that \llama\ represents cyclic concepts using base-10 numerical representations at layer 18 (e.g., patching from \textit{six months after August} into \texttt{addition} prompts causes the model to output 14, suggesting that the model computed \textit{6+8} as an intermediate step). We observe a surprising +100 echo, where the expected sum is sometimes output as 114 as well as 14.
(a) Including +100 echoes, this intervention causes \llama\ to output the predicted sum in over 60\% of examples, without ever causing the model to output a month. (b) We observe a clean diagonal, implying that the model distinguishes between, e.g., 9 and 21 at this point (even though they both eventually map to \textit{September}). We show similar results for other cyclic tasks in App.~\ref{app:cross-task}.}
    \label{fig:months-to-addition}
\end{figure}

\paragraph{Patching from cyclic tasks to addition.}\label{sec:expose-sum}
Next, we perform the opposite experiment: we patch from cyclic tasks into standard addition prompts. We expect that patching from \textit{What is six months after August?} to any prompt \textit{a+b=} will cause \llama\ to output the number \textit{14}, since August is the eighth month and 6+8=14. 
Again, we patch within the union of per-task output spaces at each layer. 
The clean diagonal we observe in Figure~\ref{fig:months-to-addition}b shows that prompts with the same output month but different pre-modulo sums (e.g., 3,15$\rightarrow$March) always cause the model to output \textit{different} numbers under patching. Even for \texttt{months} prompts that the model answers incorrectly, like \textit{twenty months after October}, patching into \texttt{addition} reveals that the model had computed the correct sum (e.g., 30) in 63\% of interventions. This means that for many \texttt{months} prompts, \llama\ can internally compute the sum, but then struggles to map that sum to the correct output month. Notably, even though we are patching from \texttt{months}, this intervention never increases the model's predicted probability for a month. These results hold for \texttt{hours} and \texttt{weekdays} (App.~\ref{app:cross-task}), implying that numerical representations are indeed computed in cyclic forward passes.

Figure~\ref{fig:months-to-addition}b reveals an ``echo'' pattern: patching into an addition context also raises the probability of 100 + the target token, occasionally surpassing the target token itself (e.g., patching \textit{six months after August} increases the probability of 14 and 114). The same pattern arises for \texttt{weekdays} and \texttt{hours}. A possible explanation is that Llama-3.1-8B never needs to represent the hundreds place when reasoning over cyclic concepts, since the relevant sums are always small, consistent with the finding that larger Fourier periodicities are causally unimportant for these tasks (App.~\ref{app:steering}).

\section{Fourier Probes Trained on Addition Can Steer Cyclic Tasks}\label{sec:fourier}

If the model uses addition for these cyclic tasks, prior work suggests it may use ``Fourier features'' \citep{nanda2023modular, zhong2023pizza, zhoufourier, kantamnenihelix,heinzerling2024monotonic}. 
We first replicate results from prior work, training probes on the \texttt{addition} task that best recover periods $T\in\{2,5,10,20,50,100\}$ \citep{DBLP:conf/naacl/LevyG25, kantamnenihelix,fu2026convergentevolutiondifferentlanguage, gould2023successor,zhu2025linear,kadlcik2025pretrained}.
Then, we show that these Fourier probes generalize to cyclic tasks by using them to steer outputs for \texttt{months}, \texttt{weekdays}, and \texttt{hours}. Because probes do not always find causal features \citep{belinkov2022probing, sharkey2025open}, the fact that these probes trained on addition have a causal effect on unseen tasks is strong evidence for the importance of Fourier features in this setting.

\begin{figure}[t]
    \centering
\vspace{-4mm}
\includegraphics[width=0.9\linewidth]{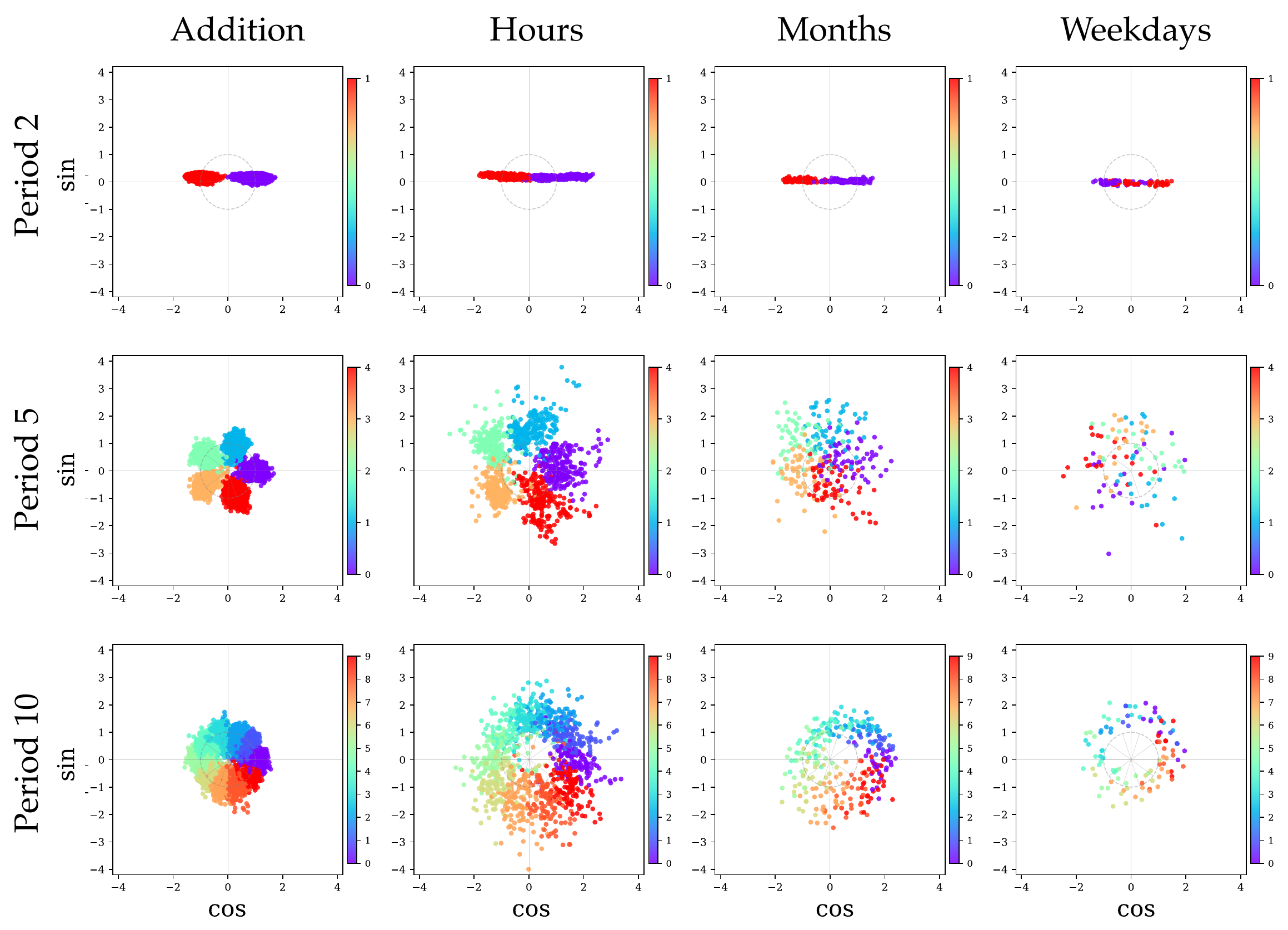}
    \caption{
Applying Fourier probes trained on the \texttt{addition} task to layer 18 activations reveals circular structures with base-10 periodicities across all tasks; here, we show $T\in\{2,5,10\}$. Although hours, months, and weekdays have their own natural periods (e.g., 24, 12, and 7), the model represents intermediate sums in these tasks using a base-10 system.
    }
\label{fig:circles}
\end{figure}

\paragraph{Training Fourier probes.} 
Let \(\textbf{h}_{a + b}^{(l)}\) be a hidden state at layer $l$ for the last token in the prompt \textit{a+b=}. 
For each period $1\leq T\leq 150$, we train two affine probes to follow sine and cosine harmonics by minimizing the following losses:
\begin{equation}
    \mathsf{MSE}\Bigg( \langle\mathbf{w}_{\sin}^{(l,T)} \mathbf{h}^{(l)}_{a + b}\rangle + b_{\sin}^{(l,T)}, \sin\!\left(\tfrac{2\pi(a+b)}{T}\right)\Bigg),  \;
    \mathsf{MSE}\Bigg(\langle\mathbf{w}_{\cos}^{(l,T)} \mathbf{h}^{(l)}_{a+b}\rangle + b_{\cos}^{(l,T)},\cos\!\left(\tfrac{2\pi(a+b)}{T}\right)\Bigg).
\end{equation}
Here, \(\mathbf{w}_{\sin}^{(l,T)}, \mathbf{w}_{\cos}^{(l,T)} \in \mathbb{R}^{d}\) and \(b_{\sin}^{(l,T)}\), \(b_{\cos}^{(l,T)}\) are scalar biases. For example, given the prompt \textit{8+5=13}, our probes would be trained to recover the gold labels 
$\sin(13(2\pi/T))$
and 
$\cos(13(2\pi/T))$
from \(\mathbf{h}^{(l)}_{8+5}\). When we train these probes for the \texttt{addition} task, we find that periods $T \in \{2,5,10,20,50,100\}$ can be reliably recovered. These periods emerge in middle layers, in agreement with our causal analysis. See App.~\ref{app:fourier_training} for full results. 
Applying probes to layer 18 activations reveals circular structures with base-10 periodicities across all tasks (Figure~\ref{fig:circles}). Although the hours, months, and weekdays tasks have natural periods (e.g., 24, 12, and 7), the model represents the intermediate sums in these tasks using a base-10 system. 

\begin{figure}[t]
    \centering
\vspace{-4mm}    
\includegraphics[width=\linewidth]{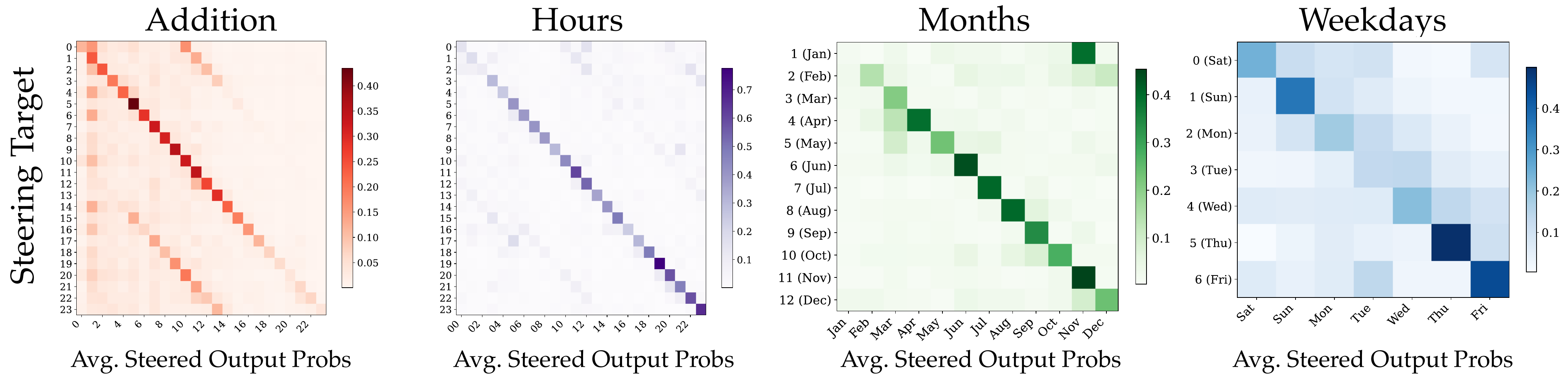}
    \caption{
Steering with Fourier probes at layer 18 shows that Fourier features identified for the \textit{addition} task are causal across all tasks. For each original prompt, we steer towards a numeric target $n'$ by modifying the Fourier features to encode that value. Each row shows the average output distribution after steering all prompts toward a specific target. 
A strong diagonal indicates a successful intervention.
    }
\label{fig:steering}
\end{figure}

\paragraph{Steering on cyclic tasks with Fourier probes.}
\label{sec:fourier_causal}
To show that these features are causally important, we use the Fourier probes trained on \texttt{addition} to steer outputs for cyclic tasks. In essence, we use the probe directions to set the model's pre-modulo sum $n$ to some counterfactual $n'$, expecting the model to then map the counterfactual sum $n'$ to concept space. 
For example, for the prompt \textit{What is four months after March?}, the original sum is $n$=7. If we steer to a target sum of $n'$=6, we expect the model to output \textit{June} instead of \textit{July}. 

Let $\mathbf{h}^{(l)}_{\mathsf{Cyclic}(a, b)}$ be the residual stream representation at layer $l$ for the last token in a cyclic task prompt.
To steer the model to a target sum \(n'\) for this prompt, we modify the residual stream so that its Fourier coefficients match the target phase for each period $T$. For each probe with period \(T\), we compute the new desired phase $\theta_T^{n'}=n'(2\pi/T)$ and cache the empirically observed radius $r_T$ from the original forward pass:
\[
\hat{s} = \mathbf{w}_{\sin}^{(l,T)} \mathbf{h}_{\mathsf{Cyclic}(a,b)}^{(l)} + b_{\sin}^{(l,T)},\qquad
\hat{c} = \mathbf{w}_{\cos}^{(l,T)} \mathbf{h}_{\mathsf{Cyclic}(a,b)}^{(l)} + b_{\cos}^{(l,T)},\qquad r_{T} = \sqrt{\hat{s}^2 + \hat{c}^2}.
\]
Then, we construct a steered activation with scaling factor $\alpha$ that increases the radius proportional to the original values:
\begin{equation}
\tilde{\mathbf{h}}^{(l)}_{\mathsf{Cyclic}(a,b)} \leftarrow \mathbf{h}^{(l)}_{\mathsf{Cyclic}(a,b)} 
+ \Bigg(\alpha\, r_T\sin(\theta^{n'}_T) - \hat{s}\Bigg)\,\hat{\mathbf{w}}_{\sin}^{(l,T)}
+ \Bigg(\alpha r_{T} \cos(\theta^{n'}_T) - \hat{c}\Bigg)\hat{\mathbf{w}}_{\cos}^{(l,T)},
\end{equation}
where $\hat{\mathbf{w}}_{\sin}^{(l,T)},\hat{\mathbf{w}}_{\cos}^{(l,T)}$ are normalized vectors for the probes. To determine which $T$ to steer on for each task, we measure overlap of Fourier probes with DAS subspaces; this means that e.g., period $T=50$ is not used for the weekday task. For details, see App.~\ref{app:steering}.

Figure~\ref{fig:steering} shows the resulting output distributions for steering every task at layer 18. Each row corresponds to the mean output distribution after steering all prompts in the task toward a specific target value; successful steering should yield a strong diagonal pattern, indicating that probability mass shifts to the intended target. In all experiments, we use a scaling factor of \(\alpha = 10\), which controls the strength of the intervention. The need for this high $\alpha$ suggests that these features alone may not fully override downstream computation.

\section{Decomposing The Shared MLP Addition Module into Subcircuits}\label{sec:neuron-analysis}

If \llama\ represents the output sum using Fourier features, how do its parameters interact with this geometry? 
In this section, we study a small set of 28 MLP neurons\footnote{See~\cite{merrill2023tale} that study competition between sparse and dense subnetworks during grokking and offers a complementary perspective on why sparse arithmetic circuits could emerge.} at layer~18 that use Fourier features to compute the sum of two numbers.

\begin{figure}
    \centering
\vspace{-4mm}    
\includegraphics[width=\linewidth]{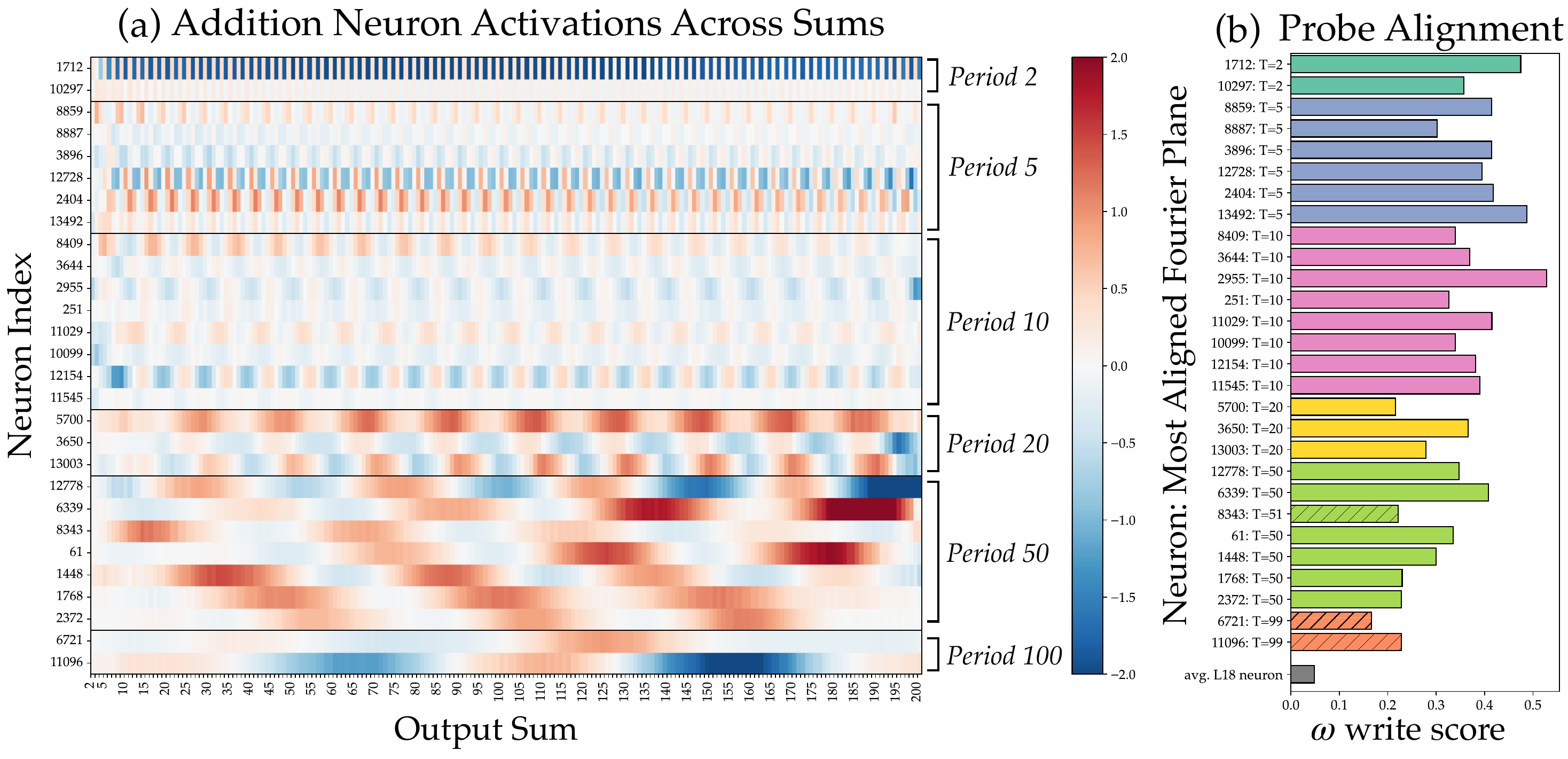}
    \caption{Addition neurons are sparse and can be grouped by the Fourier periodicities found in Section~\ref{sec:fourier}. (a) Activations of neurons in $\mathcal{N}_{\text{add}}$ averaged across prompts with the same output sum; colormap is clipped to [-2,2] for visibility. (b) For each neuron, we report the pair of Fourier probes that this neuron is most aligned with across all $2\leq T\leq150$. Every neuron writes to the Fourier plane from Section~\ref{sec:fourier} that corresponds to its activation pattern across examples. As a baseline (gray), we report maximum Fourier plane alignment averaged across every MLP neuron at L18.}
    \label{fig:ribbons}
\end{figure}

\subsection{Identifying Addition Neurons}\label{sec:identify-neurons}
We focus on the MLP at layer 18, which is a gated MLP \citep{Shazeer2020GLUVI}. Thus, if $d_{\text{mlp}}$ is the MLP dimension (number of neurons) and $d_{\text{model}}$ is the residual stream dimension, then for a hidden state $\mathbf{h}\in\mathbb{R}^{d_{\text{model}}}$ and $\mathbf{W}_{\text{gate}},\mathbf{W}_{\text{up}},\mathbf{W}_{\text{down}}\in\mathbb{R}^{d_{\text{mlp}}\times d_{\text{model}}}$ , the MLP is defined as
\begin{equation}
    \text{MLP}(\mathbf{h}) = (\text{SiLU}(\mathbf{W}_{\text{gate}}\mathbf{h}) \odot \textbf{W}_{\text{up}}\mathbf{h})\textbf{W}_{\text{down}}
\end{equation}
where $\odot$ is element-wise multiplication. We refer to a single neuron $n_i$ as the collection of three vectors: the gate, up, and down weights $\mathbf{g}_i,\mathbf{u}_i,\mathbf{d}_i\in\mathbb{R}^d$ that correspond to index $i$. 

We want to find neurons that write to causally-important subspaces found by DAS; i.e., that are causally important for computing the sum~\citep{geva2021feedforward,arora2025language, hanagreaterthan}. For each task, we look for MLP neurons at layer 18 that write to the DAS \outputconcept\ subspace for that task, spanned by $\mathbf{R}_{\text{task}}\in\mathbb{R}^{d\times k}$ from Section~\ref{sec:das}. For a given neuron's down weight $\mathbf{d}_i$, we calculate
\begin{equation}\label{eq:write-score}
    \omega_{\text{task}}(\mathbf{d}_i)=||\mathbf{R}_{\text{task}}\mathbf{d}_i||/||\mathbf{d}_i||.
\end{equation}

Empirically, only a small number of MLP neurons at layer 18 have high $\omega$. We plot the distribution of $\omega$ for each task in Figure~\ref{fig:neuron-hist}; with a threshold of $\omega>0.4$, we identify 16 neurons for \texttt{months}, 15 neurons for \texttt{weekdays}, 26 for \texttt{hours}, and 28 for \texttt{addition}. All neurons important for cyclic tasks are a strict subset of those important for \texttt{addition}, except for a single outlying \texttt{hours} neuron, which we ignore. See App.~\ref{app:neuron-selection} for details. 
For every pair of tasks, Pearson correlation of $\omega$ scores is high, with $r\geq0.70$ ($p=0.0$). 
\vspace{-5pt}
\paragraph{Neuron ablations.} We focus on the set of 28 addition neurons in layer 18 that satisfy our threshold, $\mathcal{N}_{\text{add}}=\{n_i|\omega_{\text{addition}}(\mathbf{d}_i)>0.4\}$. 
When these neurons are zero-ablated, model accuracy decreases significantly (e.g., from 95\%$\rightarrow$24\% for \texttt{addition}). Despite making up only 0.2\% of MLP neurons in layer 18, these neurons explain most of the computation in this sublayer: ablating all other layer 18 MLP neurons except for $\mathcal{N}_{\text{add}}$ retains most of the model's performance (e.g., 95\%$\rightarrow$86\% for \texttt{addition}). We also find that these neurons are causally important for unseen prompt templates; see Table~\ref{tab:l18-neuron-ablation} for full ablations.

\paragraph{Partitioning neurons according to Fourier features.}
First, we visualize the average activation of each neuron $n_i\in\mathcal{N}_{\text{add}}$ for every possible output sum in the \texttt{addition} task in Figure~\ref{fig:ribbons}a, finding that neurons fire periodically across sums; all have similar activation patterns for \texttt{months}, \texttt{weekdays}, and \texttt{hours} (Figure~\ref{fig:unclipped-ribbons}).
We can partition these neurons into disjoint clusters that activate with period $T\in\{2,5,10,20,50,100\}$. In fact, each cluster of neurons writes to Fourier directions with the same period: we calculate how much each $\mathbf{d}_i$ overlaps with the span of Fourier probes $\mathbf{w}^{(T)}_{\sin},\mathbf{w}^{(T)}_{\cos}$ across values of $T$ (see Eq.~\ref{eq:write-score}). Figure~\ref{fig:ribbons}b shows that each cluster of neurons activating with period $T$ also writes to Fourier features with the same period. Measuring absolute cosine similarity of $\mathbf{d}_i$, we observe that these clusters are perfectly orthogonal to each other (Fig.~\ref{fig:cluster-neurons}). 

\paragraph{Larger periods are possibly missing.}
Although these neurons are crucial to explain layer 18 computation, we are likely missing neurons with larger periods. When we zero-ablate all other neurons at layer 18 except for $\mathcal{N}_{\text{add}}$, the small drop in accuracy from 95\%$\rightarrow$86\% comes mostly from examples where both summands are large (Figure~\ref{fig:only-addition-errors}), suggesting that our threshold may exclude some neurons with large $T$.

\subsection{How Addition Neurons Compute the Sum}

\begin{figure}
    \centering 
\includegraphics[width=\linewidth]{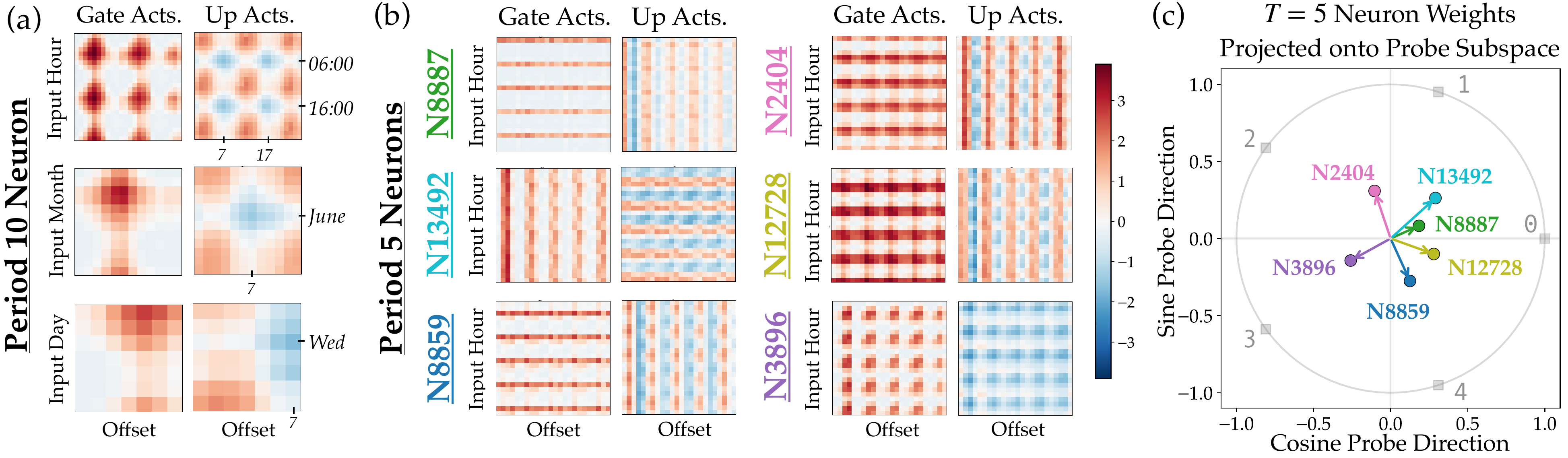}
    \caption{Addition neuron activations organized by \inputconcept\ and \inputnumber, separated into gate and up activations. (a) Activations for a period 10 neuron $n_{8409}$. This neuron has the same activation pattern across all cyclic tasks: e.g., its up projection is negative for \textit{seven hours after 06:00} as well as \textit{seven months after June}. Because it is period 10, it also activates for \textit{seventeen hours after 16:00}. (b) Activations for all period 5 neurons for the \texttt{hours} task. (c) Down projection rows for all period 5 neurons, projected onto the Fourier plane for $T=5$.}
    \label{fig:neurons-main} 
\end{figure}

\begin{figure}[t]
    \centering
\includegraphics[width=\linewidth]{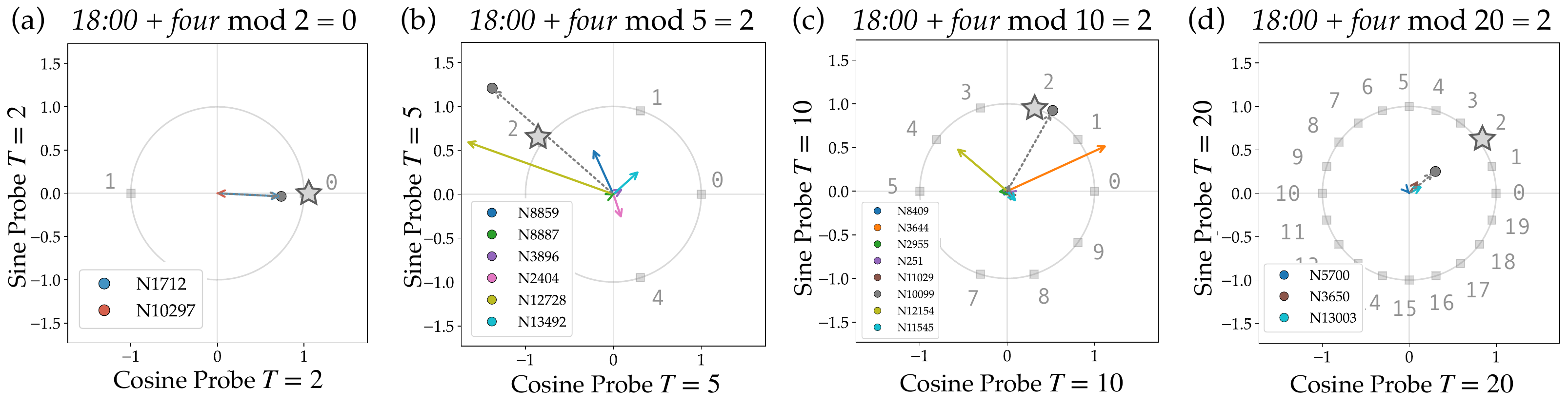}
    \caption{The model computes the sum \textit{18+4=22} on multiple orthogonal planes, each encoding a different modulo. We visualize all addition neurons for periods $T\in\{2,5,10,20\}$ projected onto their respective Fourier planes: arrows indicate each neuron's down projection row $\mathbf{d}_i$ scaled by its activation for the prompt \textit{four hours after 18:00}. The gray dotted line indicate the sum of these vectors, and gray stars indicate the ground truth label.}
    \label{fig:actual-forward-pass}
\end{figure}

First, we analyze activation patterns for addition neurons. We arrange activations in grid format, where each square corresponds to a neuron's activation for a single prompt. Rows correspond to prompts with the same \inputconcept, while columns correspond to prompts with the same \inputnumber. We separate activations into $\mathbf{W}_{\text{gate}}$ and $\mathbf{W}_{\text{up}}$ activations, before they are element-wise multiplied. Observe that all neurons have similar activation patterns across cyclic tasks: for example, Figure~\ref{fig:neurons-main}a shows a consistent pattern for $n_{8409}$ for \texttt{hours}, \texttt{months}, and \texttt{weekdays}. Additionally, if the cluster that a neuron belongs to is period $T$, it fires with periodicity $T$ over its inputs (e.g., $n_{8409}$ fires for \textit{06:00} and \textit{16:00}). See App.~\ref{app:neurons-all-tasks}.

Second, we observe two types of neuron activation patterns: \textit{split} patterns and \textit{mixed} patterns. If a neuron has a \textit{split} activation pattern, this means that it is using its gate vector $\mathbf{g}_i$ to read from one summand and its up vector $\mathbf{u}_i$ to read from the other. We hypothesize that this split behavior helps the model take advantage of element-wise multiplication to combine information from two disparate subspaces. For example, in Figure~\ref{fig:neurons-main}b, $n_{8887}$ has a \textit{split} activation pattern for cyclic tasks, where its gate activation fires whenever the \inputconcept\ is a multiple of five (e.g., \textit{15:00}, horizontal stripes), and its up activation fires whenever the \inputnumber\ mod~5~=~1 (e.g., \textit{in six hours, in eleven hours}, vertical stripes). \textit{Mixed} neurons use both vectors to read both input arguments. Figure~\ref{fig:neurons-main}a shows a neuron with mixed activation patterns: its gate and up activations respond to both summands. We find that approximately 17/28 neurons have split behavior for cyclic tasks (App.~\ref{app:all-neurons}). For addition, all neurons have mixed activation patterns, possibly related to the fact that IIA for \texttt{addition} at this sublayer is only about 60\%, indicating that input arguments are more ``fused.''

Third, we observe that calculation of the sum is performed in a distributed manner, where neurons with the same periodicity combine to predict the sum within that periodicity. For example,  Figure~\ref{fig:neurons-main}c shows every period 5 neuron's down projection row projected to the $T=5$ Fourier plane alongside its activations: in a forward pass, these down projection rows are scaled by their respective activations such that the sum of these vectors ``points to'' the correct output. Figure~\ref{fig:actual-forward-pass} shows real neuron activations for periods $T\in\{2,5,10,20\}$ for the prompt \textit{four hours after 18:00}. Within each period, neurons work together, summing to a vector with an angle encoding the output $4+18\pmod{T}$.

\begin{figure}[t]
    \centering
    \includegraphics[width=\linewidth]{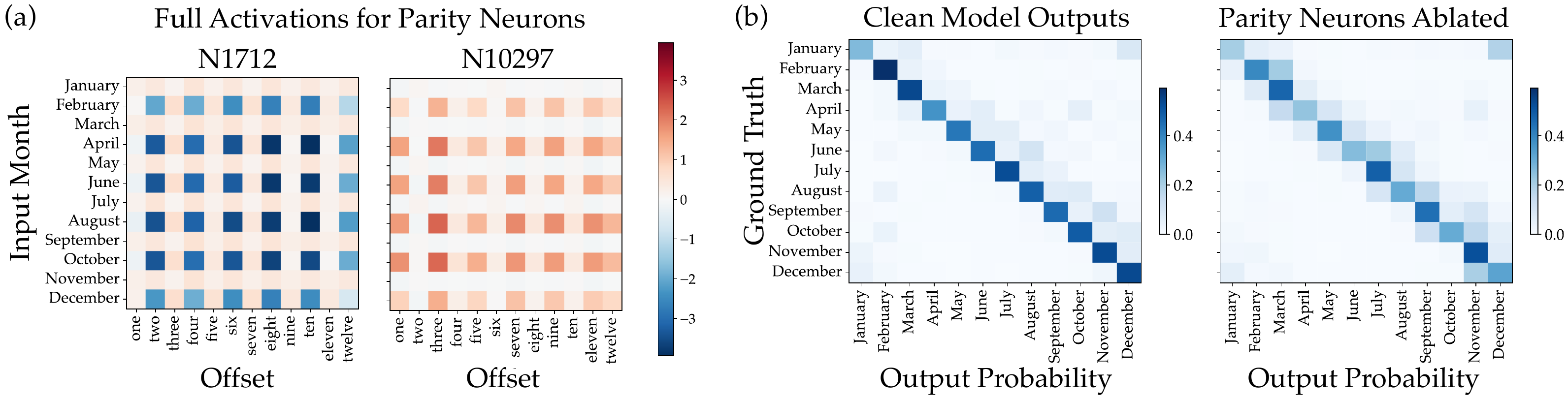}
    \caption{Parity neurons ($T=2$) sharpen model predictions. (a) Full activations ($\text{gate} \cdot \text{up}$) for both parity neurons on the \texttt{months} task. These neurons write in opposite directions for alternating prompts. (b) Zero-ablating just these two neurons shifts probability towards neighboring outputs with the wrong parity, ``blurring'' the model's predictions.}
    \label{fig:parity}
\end{figure}

\paragraph{Why use several periods?} One might wonder why \llama\ needs to represent smaller periods at all if, e.g., $T=100$ could theoretically differentiate between numbers up to 100. As \cite{zhoufourier} explain, smaller periods help to sharpen the model's representations. Take $T=2$ as an example. We find two parity neurons that play complementary roles, writing in opposite directions: $n_{10297}$ is an ``odd neuron'' that fires for \textit{even+odd=odd}, while $n_{1712}$ is an ``even neuron'' that fires only for \textit{even+even=even}.\footnote{Cosine similarity of $\mathbf{d}_{10297},\mathbf{d}_{1712}$ in the output subspace for \texttt{addition} is 0.997, but activations for $n_{1712}$ are always negative; thus, these two neurons write in opposite directions.}
Figure~\ref{fig:parity}b shows that if we zero-ablate \textit{only} these two neurons at the last token position, probability mass shifts away from the correct answer and towards neighboring answers, ``blurring'' the model's outputs.

\section{Related Work}

\paragraph{Representation geometry.} Neural network representations have been shown to encode abstract concepts along intricate, nonlinearly curved geometries across architectures and modalities~\citep{thomasrabbithull, csordas2024recurrent, lubana2025priorstimemissinginductive, costa2025flathierarchicalextractingsparse, shafran2025decomposingmlpactivationsinterpretable, saxe2019mathematical, maheswaranathan2019universality, parkhierarchical, park2025, shai2026transformerslearnfactoredrepresentations, pearce2025tree, gurnee2025when, yocum2025neural, brenner2026grid, piotrowski2024constrained, shai2024transformers, kantamnenihelix, song2023uncovering, doshi2026bi}.
As recent work has started to concretize in models of scale~\citep{karkada2026symmetrylanguagestatisticsshapes, prieto2026correlations, korchinski2025emergence}, similar to prior work in toy scenarios~\citep{saxe2019mathematical, shai2024transformers, arora-etal-2018-linear, parkhierarchical}, these geometric structures are an artifact of data statistics---a model optimally encoding the data distribution ought to organize a concept along specific geometries, as claimed by these works and corroborated empirically by literature.
However, despite the richness of this literature, there has been minimal work justifying the causal role this representation geometry plays in behavior, i.e, over the output space of the model.
Our work takes a step towards filling this gap by developing a precise account of a task known to show geometrically structured representations in toy scenarios, i.e., modular addition~\citep{nanda2023modular, morwani2024feature}, by analyzing it as arithmetic over concepts known to show circular geometries in model representations (e.g., weekdays and months)~\citep{engelscircles, karkada2026symmetrylanguagestatisticsshapes, park2025, nishi2024representation}.

\paragraph{Causal analysis of neural networks.}
We build on prior interpretability research grounded in the theory of causality \citep{hume1748enquiry, Pearl1999, Spirtes}. 
In particular, causal mediation \citep{pearl2001directindirecteffects, vig2020, Mueller2026} has been used to characterize the flow of information in neural networks and causal abstraction \citep{Rubinstein2017, BeckersHalpern2019,geiger2021, geiger2025ca, Geiger2026} has been used to uncover algorithmic processes.
Empirically, we used a variety of different intervention have been applied to neural network hidden representations, including ablations \citep{li2017understandingneuralnetworksrepresentation, cammarata2020thread:,Elazar-etal-2020, linear_adversarial_concept_Eraser, Ravfogel2023, kernelized_concept_Eraser, LEACE, geva2023dissectingrecallfactualassociations, meng2022locating,Meng23}, steering \citep{Giulianelli:2018,bau2018identifyingcontrollingimportantneurons, bau2018gandissectionvisualizingunderstanding,Besserve20,subramani2022extracting,antverg2022pitfallsanalyzingindividualneurons, marks2023geometry}, and interchange interventions \citep{vig2020, geiger-etal-2020-neural, finlayson-etal-2021-causal,Davies:2023, DBLP:conf/emnlp/StolfoBS23, Guerner2023CausalProbing,wang2023interpretability,Todd2024,arora2024causalgym, huang-etal-2024-ravel, feng2024how, mueller2025mibmechanisticinterpretabilitybenchmark,prakash2025languagemodelsuselookbacks, gur-arieh-etal-2025-enhancing, grant2025emergentsymbollikenumbervariables, Diego2024, sutter2025nonlinearrepresentationdilemmacausal}.

\paragraph{Interpretability on mathematical reasoning in LMs.} We build on prior work aimed at understanding how LMs perform arithmetic. 
\cite{hanagreaterthan} and \cite{boundlessdas} study greater-than operations, while \cite{nanda2023modular} and \cite{zhong2023pizza} study modular arithmetic in small transformers. 
Several works isolate specific components responsible for arithmetic \citep{yanivheuristics, DBLP:conf/emnlp/StolfoBS23, zhang2024interpretingimprovinglargelanguage,quirke2024understanding,quirke2025subtraction, baileemultiplication}. 
Of these works, several have identified base-10 representations of numbers \citep{DBLP:conf/naacl/LevyG25, kantamnenihelix, zhoufourier, zhou2025fone}, although we provide a new perspective by unifying these representations with non-numerical concepts.
More critically, we identify the computational basis, i.e., the concrete neurons that underlie these geometries and the corresponding computation performed over it in a model of scale. 
While prior work has partially shown such results, to our knowledge, these studies have been limited to toy scenarios that primarily show a correlation between geometry and the precise algorithm a model implements to perform a task~\citep{gopalani2024abrupt, gopalani2025happens, edelman2023pareto, morwani2024feature, kunin2025alternating, marchetti2026sequential} or focused primarily on representation geometry, i.e., does not demonstrate how these geometries are implemented or used in computation~\citep{zhou2025fone, zhoufourier, kantamnenihelix}.

\section{Conclusion}

In this work, we discovered a generic addition mechanism that is re-used across several domains. Although prior work shows that transformers can compute modular addition in one step, we show that \llama\ implements a two-step approach: it first uses a base-10 addition mechanism to compute a sum using Fourier features before mapping that number back to a concept-specific period (e.g., 7, 12, or 24). 
Furthermore, we were able to isolate a sparse set of 28 neurons writing to specific Fourier features of different periods.
Our work highlights the interplay between causal analysis and geometric understanding: causal analysis enabled us to localize subspaces containing information used by the model for computation, and analyzing the geometry of the representations within those subspaces provided deeper insight into how these computations are implemented.

\section*{Limitations} 
\paragraph{\textbf{Models.}} In this work, we conduct an in-depth analysis of how \llama\  reasons about cyclic concepts. While prior work suggests that some of the phenomena we observe---such as Fourier features \citep{zhoufourier,kantamnenihelix} or circular representations \citep{engelscircles}---also appear in other models, we cannot assume that the specific mechanisms identified in this work generalize beyond \llama.

\paragraph{\textbf{Data.}} Our analysis is limited to three types of cyclic concepts (hours, days, and months), while many other cyclic structures exist and remain unexplored. Moreover, we consider only a specific class of reasoning tasks—specifically, addition-based problems. Other forms of cyclic reasoning, such as backward offsets (e.g., “what month is 4 months before July?”) or distance queries (e.g., “If it’s July and I have an appointment in September, how many months away is that?”), are outside the scope of this study.

\paragraph{\textbf{Analysis.}} We focus primarily on how \llama\ performs the addition step within cyclic domains. Although we observe that intermediate sums are mapped to the appropriate output concept, we do not analyze \textit{how} the model computes this mapping, and leave this stage for future work.

\section*{Acknowledgments} 

We would like to thank Thomas Icard, Vasudev Shyam, Jing Huang and Ren Makino for helpful discussions throughout the course of this project. Additionally, we thank Yonatan Belinkov, David Bau, Arjun Khurana, and Andy Arditi for feedback on our initial drafts.

\newpage

\bibliography{colm2026_conference}
\bibliographystyle{colm2026_conference}

\clearpage
\appendix
\section{Task Setup}

\subsection{Task Definitions}
We define four tasks: three natural modular arithmetic tasks (\texttt{weekdays}, \texttt{months}, and \texttt{hours}), and one control \texttt{addition} task. For each task, we evaluate on every combination of \inputconcept\ (e.g., \textit{Monday}, \textit{January}, \textit{00:00}) and \inputnumber\ (e.g., \textit{one}, \textit{two}, etc.). 

\begin{table}[h]
\caption{Task templates and possible values for \textbf{input concept} and \textbf{offset}. All \textbf{output} variables are drawn from the same set as \textbf{input concept} variables. We assume that Monday=1 for \texttt{weekdays} and January=1 for \texttt{months}, and take the integer version of the hour for \texttt{hours}.}
\begin{tabular}{@{}l>{\raggedright}p{4.5cm}>{\raggedright}p{2.2cm}>{\raggedright}p{2.2cm}ll@{}}
\toprule
Task & Template & \tt concept & \tt offset & $n$ & Acc. \\ \midrule
\tt months    & \it Q: What month is \{offset\} months after \{concept\}?\textbackslash{}nA:                                                                                                        & January ...December   & one... twenty-four & 288 & 65.3\%            \\
\tt weekdays  & \it Q: What day is \{offset\} days after \{concept\}?\textbackslash{}nA:                                                                                                              &  Monday ...Sunday    & one... fourteen & 98 & 71.4\%           \\
\tt hours     & \it Q: In 24-hour time, it is now \{concept\}:00. What time will it be in \{offset\} hours?\textbackslash{}n A: In 24-hour time, it will be  & 00, 01, ... 22, 23    & one... forty-eight & 1152 & 70.8\%            \\
\tt addition  & \it a+b= & \texttt{a} $\in$ [1, ..., 100] & \texttt{b} $\in$ [1, ..., 100]      & 10k & 97.2\%           \\
 \bottomrule
\end{tabular}
\label{tab:task-templates}
\end{table}

\subsection{Model Performance}\label{app:performance}
Performance for small number ranges is acceptable, but begins to break down as \inputnumber\ increases. We first evaluate Llama-3.1-8B performance on each task broken down by whether the model has to perform a modulo operation, assuming that Monday is the first weekday and January is the first month. Table~\ref{tab:task-performance} shows that most failures come from prompts for which the model has to take a modulo, or ``wrap'' back around the concept circle (e.g., \textit{What month is four months after October}). This suggests that, even if concepts are structured as a circle, the first concept is still privileged in some sense. 

\begin{table}[h]
\centering
\caption{\llama\ accuracy for all tasks, overall and broken down in two manners. We assume that Monday is the first weekday. \textbf{By Number Range}: Performance worsens as \inputnumber\ increases. \textbf{By Pre-Modulo Sum}: the model has almost perfect accuracy for examples where the answer does not ``wrap'' around (e.g., \textit{three months after June}, $3+6\leq 12$), but performs poorly for examples where the answer does ``wrap'' (e.g., \textit{three months after November}, $3+11>12$).}
\begin{tabular}{@{}llllll@{}}          
\toprule         
\multirow{2}{*}{\textbf{Task}} & \multicolumn{2}{c}{\textbf{By Number Range}} & \multicolumn{3}{c}{\textbf{By Pre-Modulo Sum}} \\
\cmidrule(lr){2-3} \cmidrule(l){4-6}
& $1\rightarrow p$ & $p\rightarrow 2p$ & All & sum $\leq m$ & sum $> m$ \\
\midrule
\texttt{months}   & 81.9\% & 48.6\% & 65.3\% & 100\%  & 55.0\% \\
\texttt{weekdays} & 91.8\% & 51.0\% & 71.4\% & 95.2\% & 64.9\% \\
\texttt{hours}    & 97.0\% & 44.6\% & 70.8\% & 100\%  & 60.6\% \\
\texttt{addition} & ---    & ---    & 97.2\% & ---    & ---    \\
\bottomrule   
\end{tabular}                               
\label{tab:task-performance}
\end{table}

\begin{table}[h]                                  
\centering
\caption{\llama\ accuracy for cyclic tasks broken down by pre-modulo sum range, separately and and cumulatively. The model has difficulty ``taking the modulo'' of larger numbers. We assume that Monday is the first weekday.}
\begin{tabular}{@{}lllllll@{}}   
\toprule                          
\multirow{2}{*}{\textbf{Task}} & \multicolumn{3}{c}{\textbf{Separate}} & \multicolumn{3}{c}{\textbf{Cumulative}} \\
\cmidrule(lr){2-4} \cmidrule(l){5-7}                                     
& $[1,p]$ & $[p,2p]$ & $[2p,3p]$ & ${\leq}p$ & ${\leq}2p$ & ${\leq}3p$ \\
\midrule
\texttt{months}   & 100.0\% & 68.1\% & 30.8\% & 100.0\% & 78.1\% & 65.3\% \\
\texttt{weekdays} & 95.2\%  & 87.8\% & 25.0\% & 95.2\%  & 90.0\% & 71.4\% \\
\texttt{hours}    & 100.0\% & 81.1\% & 17.8\% & 100.0\% & 87.6\% & 70.8\% \\
\bottomrule   
\end{tabular}                               
\label{tab:task-performance-range}                                
\end{table}

\begin{figure}
    \centering
    \includegraphics[width=\linewidth]{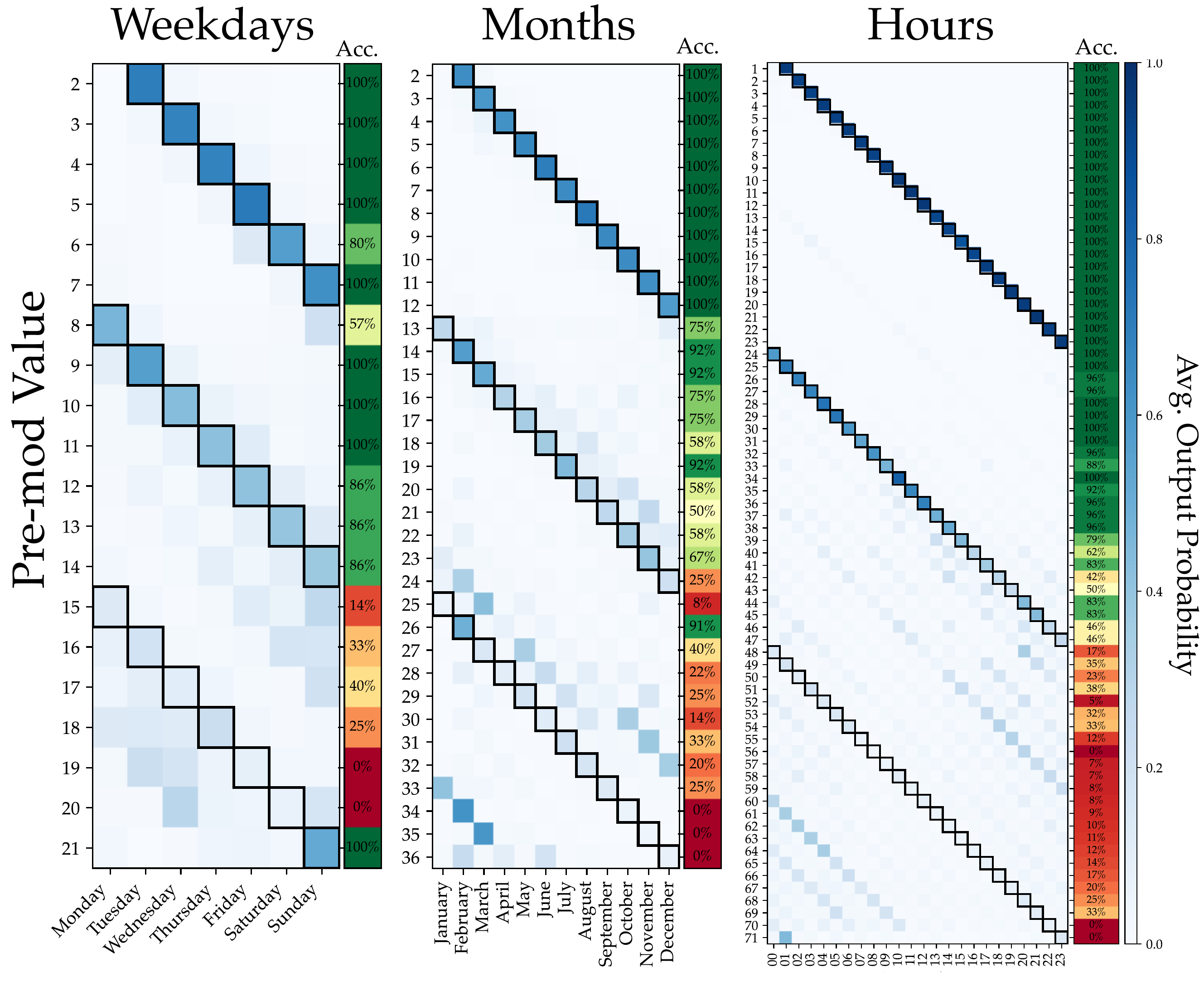}
    \caption{\textbf{Model performance for cyclic tasks.} \llama\ output probabilities for all prompts in each dataset are displayed, aggregated by pre-modulo sum. Cells for correct answers are outlined in black, and accuracy for each pre-modulo value is displayed beside each row. When the pre-modulo sum is less than or equal to the cycle length for a given concept (e.g., \textit{January + four}=\texttt{5}=\textit{May}), model accuracy is perfect, but performance begins to waver once a modulo must be calculated (e.g., \textit{January + fifteen}=\texttt{16}=\textit{April} has 75\% accuracy). As the pre-modulo sum increases, the model begins to make systematic errors (i.e., probability is placed on the wrong diagonal).}
    \label{fig:output-probs}
\end{figure}

For a more fine-grained understanding of Table~\ref{tab:task-performance}, we aggregate prompts by their pre-modulo sum and plot in Figure~\ref{fig:output-probs}. For example, if January=1, then the pre-modulo value of \textit{January + fourteen} is 15. We observe a clear pattern: as the pre-mod value increases, model accuracy decreases. Interestingly, the model's errors for higher pre-mod values are quite regular, with probability mass placed along diagonal offsets from the true answer. Table~\ref{tab:task-performance-range} summarize these heatmaps.

\paragraph{Discussion.} If weekday, month, and hour concepts are represented as circles \citep{engelscircles}, one might expect that Llama-3.1-8B performs calculations over these circles by rotating along them. In such a world, there would be no difference between \textit{four months after January} and \textit{four months after November}, as the circular geometry would treat both calculations in the same way. However, the fact that we see such a drastic drop in performance for prompts that must ``wrap around'' to the canonical start of the sequence tells us that there \textit{is} still a notion of ``start'' and ``end'' for these circular concepts.

\subsection{Can Llama Perform Standard Modular Addition?}
\label{app:explicit-mod}
\begin{figure}
    \centering
    \includegraphics[width=\linewidth]{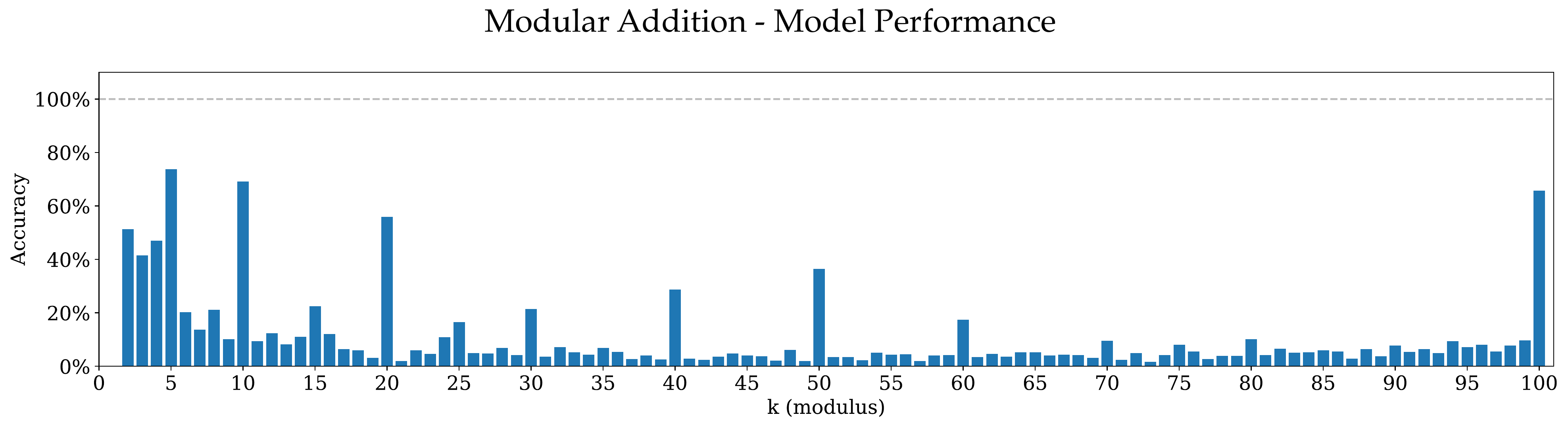}
    \caption{We test whether Llama-3.1-8B can solve prompts of the form
\texttt{Q: What is (\{a\} + \{b\}) mod \{k\}?\textbackslash nA:},
where \(a, b \in [1, 200]\) and \(k \in [2, 100]\).
We measure performance on 1{,}000 \textbf{randomly sampled} prompts of this form.
Performance is poor for most moduli.}
    \label{fig:mod_addition}
\end{figure}

We evaluate whether Llama can perform standard modular addition. To this end, we use prompts of the form
\texttt{Q: What is (\{a\} + \{b\}) mod \{k\}?\textbackslash nA:},
where \(a, b \in [1, 200]\) and \(k \in [2, 100]\), sampling 1{,}000 prompts in total. As shown in Figure~\ref{fig:mod_addition}, performance is low across most moduli, including 7, 12, and 24.

\begin{figure}
    \centering
    \includegraphics[width=0.5\linewidth]{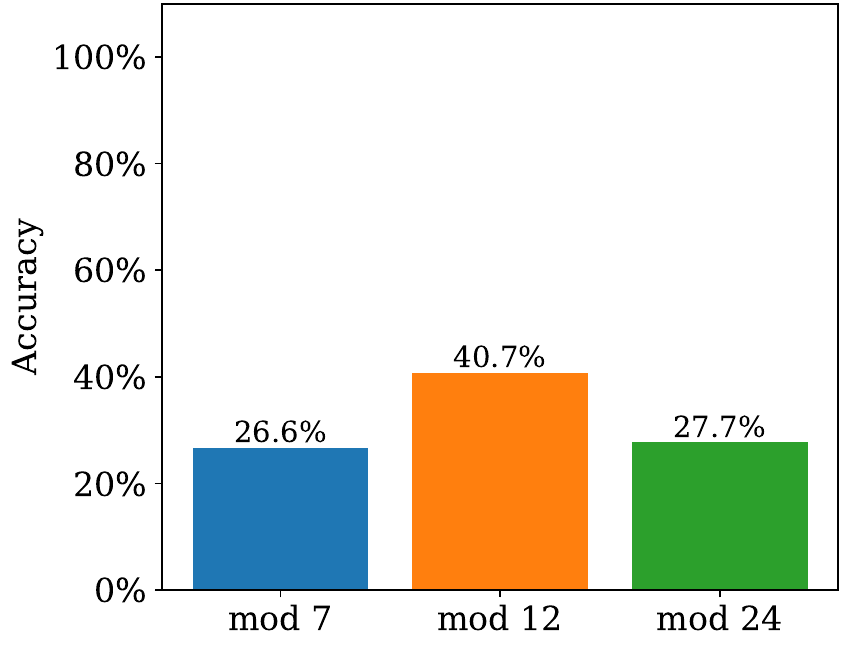}
    \caption{We test whether Llama-3.1-8B can solve prompts of the form
\texttt{Q: What is (\{a\} + \{b\}) mod \{k\}?\textbackslash nA:},
where \( k \in \{7, 12, 24\} \). For each modulus, we evaluate performance on prompts where \( a + b \leq 3k \), which is \textbf{equivalent to the range used in the cyclic tasks}. }
    \label{fig:mod_addition_comparison}
\end{figure}

To enable a more direct comparison with the cyclic tasks, for each modulus \( k \in \{7,12,24\} \) we also evaluate performance over the same range used in those tasks, namely for prompts satisfying \( a + b \leq 3k \). The results are shown in Figure~\ref{fig:mod_addition_comparison}. Comparing these results with the model’s performance on the cyclic tasks (Table~\ref{tab:task-performance}) shows that, although the cyclic tasks implicitly require modular addition, the model performs better on those tasks than on standard modular addition.

\section{Causal Abstraction and Causal Models}
\label{app:causal}

\paragraph{Input-output causal models.} 
Define an input-output causal model $\mathcal{A}$ to be a directed acyclic graph where each node is a \textit{variable} $V$ with a \textit{mechanism} $\mathcal{F}_{V}(\mathbf{p}) = v$ that determines the value $v$ of $V$ from the value $\mathbf{p}$ of its parents $\mathbf{P}$.\footnote{
We use case to distinguish a value $v$ from its variable $V$. We use bold to distinguish individual values and variables, i.e., $v$ and $V$, from sets of values and variables, i.e., $\mathbf{p}$ and $\mathbf{P}$.}
Input variables $\mathbf{X}$ lack parents and output variables $\mathbf{Y}$ lack children. 
Denote the input-output behavior of a causal model with $\mathcal{A}(\mathbf{x}) = \mathbf{y}$. We represent both algorithmic hypotheses about neural networks and neural networks themselves as causal models, with neural networks outputting logits. 

\paragraph{Interventions.} 
An \textit{intervention} $\mathbf{V} \leftarrow \mathbf{v}$ on $\mathcal{A}$ produces a new model $\mathcal{A}_{\mathbf{V} \leftarrow \mathbf{v}}$  where the mechanisms of $\mathbf{V}$ are fixed to constant functions outputing the values $\mathbf{v}$. 
Given a counterfactual input $\counter$, define an \textit{interchange intervention} $\mathbf{V} \leftarrow \mathcal{A}(\counter)$ to fix the value of variable $\mathbf{V}$ to the value they would have taken for input $\counter$. 

\paragraph{Neural network features.} 
An activation vector $\mathbf{h}$ will often not have interpretable dimensions in a standard basis. 
To solve this problem, we can \textit{featurize} the activation vector using an invertible transformation mapping into a feature space $\mathbb{F}$ where a feature $F$ is a dimension of $\mathbb{F}$.\footnote{See \cite{geiger2025causalabstractionunderpinscomputational} for a technical definition.} Principal component analysis, sparse autoencoders, and rotation-based distributed alignment search all produce linear features, where a feature value is computed by projecting a hidden vector onto a line.  
More complex representational schemes involve projecting hidden vectors onto circles or onions \citep{csordas-etal-2024-recurrent}, where features correspond to angle or magnitude rather than direction.

\paragraph{Causal abstraction.} 
Let $\mathcal{D}$ be a dataset of input-pairs. $\mathcal{A}$ is a causal abstraction of $\mathcal{N}$ if the following holds for all $(\orig, \counter) \in \mathcal{D}$:
\begin{equation}
\mathcal{N}_{\mathbf{F} \leftarrow \mathcal{N}(\counter)}(\orig) = \mathcal{A}_{\mathbf{V} \leftarrow \mathcal{M}(\counter)}(\orig)
\end{equation}

The \textit{interchange intervention accuracy} is proportion of pairs in $\mathcal{D}$ for which Eq.~\ref{eq:commute-appdx} holds.

\paragraph{Distributed Alignment Search.} 
The method \textit{distributed alignment search} localizes an abstract causal variable $V$ to a low-rank subspace of an activation vector. 
Concretely, it learns a low-rank orthogonal matrix $\mathbf{R}^{\theta} \in \mathbb{R}^{d \times k}$ such that interchange interventions on the subspace defined by the matrix produce the outputs of the algorithm under interchange intervention.
An interchange intervention on $\mathbf{R}^{\theta}$ will run the LM on $\orig$ and fix the $k$-dimensional subspace to the value it takes on for the input $\counter$. The change of basis is optimized (with $\mathcal{N}$ frozen) to maximize cross entropy with the label from $\mathcal{A}$ under an interchange intervention:
\begin{equation}
\label{eq:commute-appdx}
\mathsf{CrossEntropy}\bigl(\mathcal{N}_{\mathbf{R}^{\theta} \leftarrow \mathcal{N}(\counter)}(\orig),\mathcal{A}_{\mathbf{V} \leftarrow \mathcal{M}(\counter)}(\orig)\bigl)
\end{equation}

\subsection{Causal Model of the Addition Module}
\paragraph{Causal model.}
Let $\mathcal{M}$ be an input-output causal model with input variables
$\mathbf{X}=\{N, C^{\mathrm{in}}\}$ and output variables
$\mathbf{Y}=\{C^{\mathrm{out}}\}$.
Here, $N$ is the \inputnumber\ (shift amount), and
$C^{\mathrm{in}}$ is the \inputconcept\
(day, month, hour-word, or number-token). The intermediate variables are:
$B$ (base), $K$ (concept number), $S$ (sum), and $M$ (output number).

\paragraph{Lookup tables.}
Fix two lookup tables:
$\mathsf{con\_to\_num}$ maps concepts to number-space, and
$\mathsf{num\_to\_con}$ is its inverse on each concept-domain.
For days and months it is 1-indexed
(e.g.\ $\texttt{Mon}\mapsto 1,\ldots,\texttt{Sun}\mapsto 7$ and
$\texttt{Jan}\mapsto 1,\ldots,\texttt{Dec}\mapsto 12$);
for hour-words it is zero-indexed
(e.g.\ $\texttt{00:00}\mapsto 0,\ldots,\texttt{23:00}\mapsto 23$);
and for number-tokens it is the identity
(e.g.\ $\texttt{10}\mapsto 10$). The inverse table maps back accordingly.

\paragraph{Mechanisms.}
\[
\renewcommand{\arraystretch}{2.0}
\begin{array}{@{} l l @{}}
\mathcal{F}_B(c)=
\begin{cases}
7  & c\in\{\texttt{Mon},\ldots,\texttt{Sun}\}\\
12 & c\in\{\texttt{Jan},\ldots,\texttt{Dec}\}\\
24 & c\in\{\texttt{00:00},\ldots,\texttt{23:00}\}\\
\infty & \text{otherwise (addition)}
\end{cases}
&
\mathcal{F}_K(c)=\mathsf{con\_to\_num}(c),
\\[0.8em]
\mathcal{F}_S(n,k)=n+k,
&
\mathcal{F}_M(s)=s,
\\
\multicolumn{2}{@{}l@{}}{%
\mathcal{F}_{C^{\mathrm{out}}}(m,b)=
\mathsf{num\_to\_con}(m \bmod b),
\quad
\text{where } s \bmod \infty \equiv s.}
\end{array}
\]

\section{Coarse Localization with Residual Stream Patching}\label{app:interchanges}
\begin{figure}[h]
    \centering
    \includegraphics[width=0.8\linewidth]{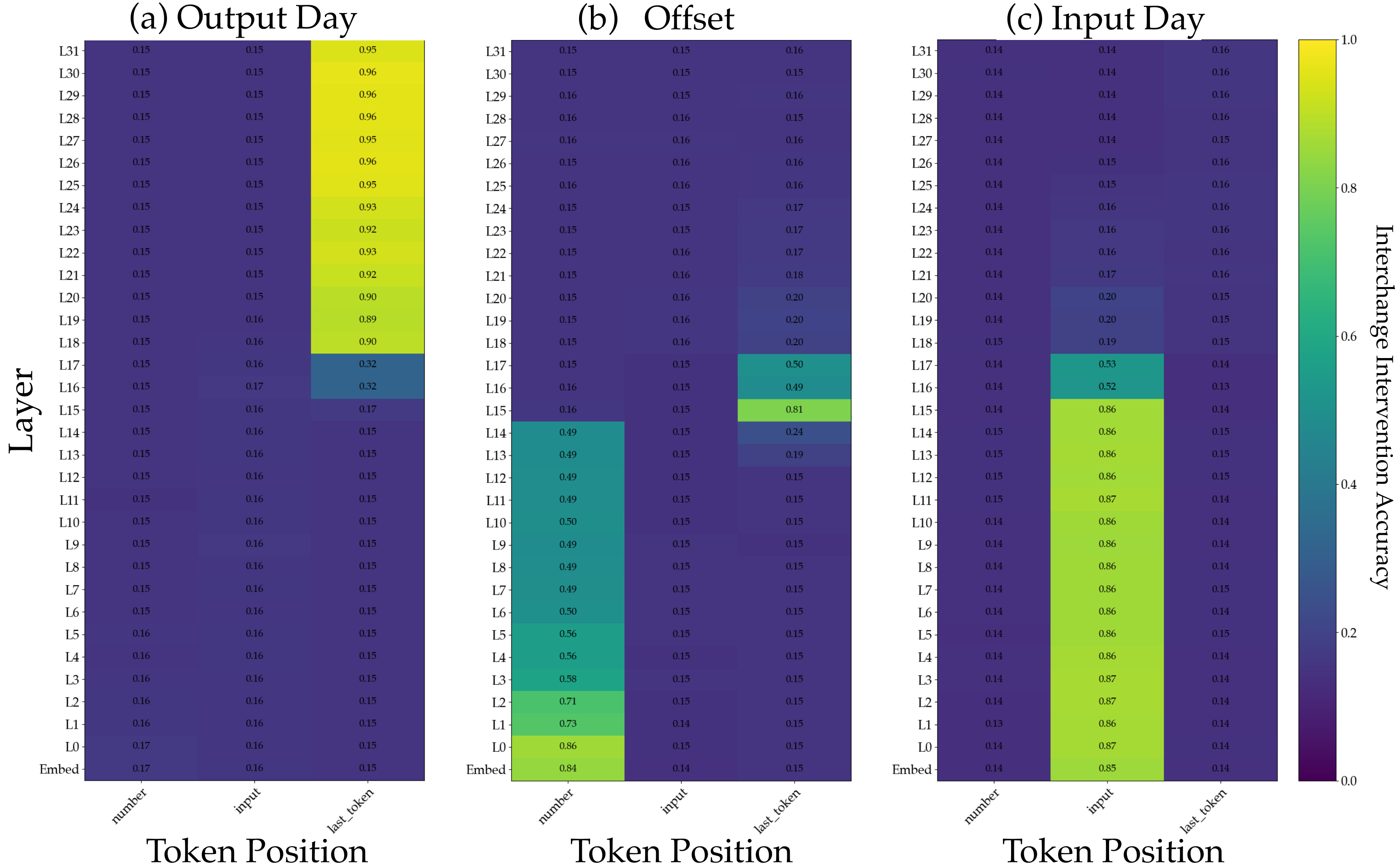}
    \caption{Residual stream patching results for the \texttt{weekdays} task for each causal variable. Patching is done for $n=4096$ counterfactual pairs (Section~\ref{app:das-setup}). (a) The \outputconcept\ appears at the last token position after approximately layer 18. (b) The \inputnumber\ is moved to the last token position at layer 15. (c) Residual stream patching can only isolate the \inputconcept\ variable at its own token position.  Patching has no effect after layer 15-17, which suggests that the input day is copied to the last token position in layer 18. The template for this task is \textit{Q: What day is \{number\} days after \{input\}?\textbackslash{}nA:} }
    \label{fig:weekdays-interchange}
\end{figure}

To determine which token positions are important, we begin with simple residual stream interchange interventions \citep{geiger2021}. Using the counterfactual dataset from Section~\ref{app:das-setup}, we patch the entire residual stream from a source prompt $\mathbf{h}_{src}$ at layer $l$ into a base prompt and measure how the model's output changes. Figure~\ref{fig:weekdays-interchange}a shows results for the \texttt{output} variable in the \texttt{weekdays} task: the model's eventual output appears at the last token position at layer 18. 

But is this answer actually \textit{computed} at the last token position, or is it copied? The answer appears to be that computation occurs at the last token, because we can intervene on the \inputnumber\ at this position without changing the \inputconcept. Figure~\ref{fig:weekdays-interchange}b shows that the \texttt{number} variable can also be localized at the last token position, by patching entire hidden states at layer 15: 81\% of the time, patching from e.g. \textit{two days after Monday} $\rightarrow$ \textit{five days after Thursday} at this position causes the model to output \textit{Saturday} (\textit{Thursday} + 2). 

The \texttt{input} variable is not as cleanly separable at the last token position (Figure~\ref{fig:weekdays-interchange}c), but this makes sense: if the \texttt{number} variable is always copied to the last token position first, then there is no point in the residual stream where \texttt{input} can be disentangled via residual stream patching. Patching at the input token position loses its effectiveness after layer 17, suggesting that a head in layer 18 copies the input day to the last token position. 

Taken together, these results suggest that for all three tasks, answers are calculated at the last token position. The model first copies \texttt{number} information to the last token at layer 15, followed by \texttt{input} day/month/hour information, yielding a result around layer 18.

\begin{figure}[h]
    \centering
    \includegraphics[width=0.8\linewidth]{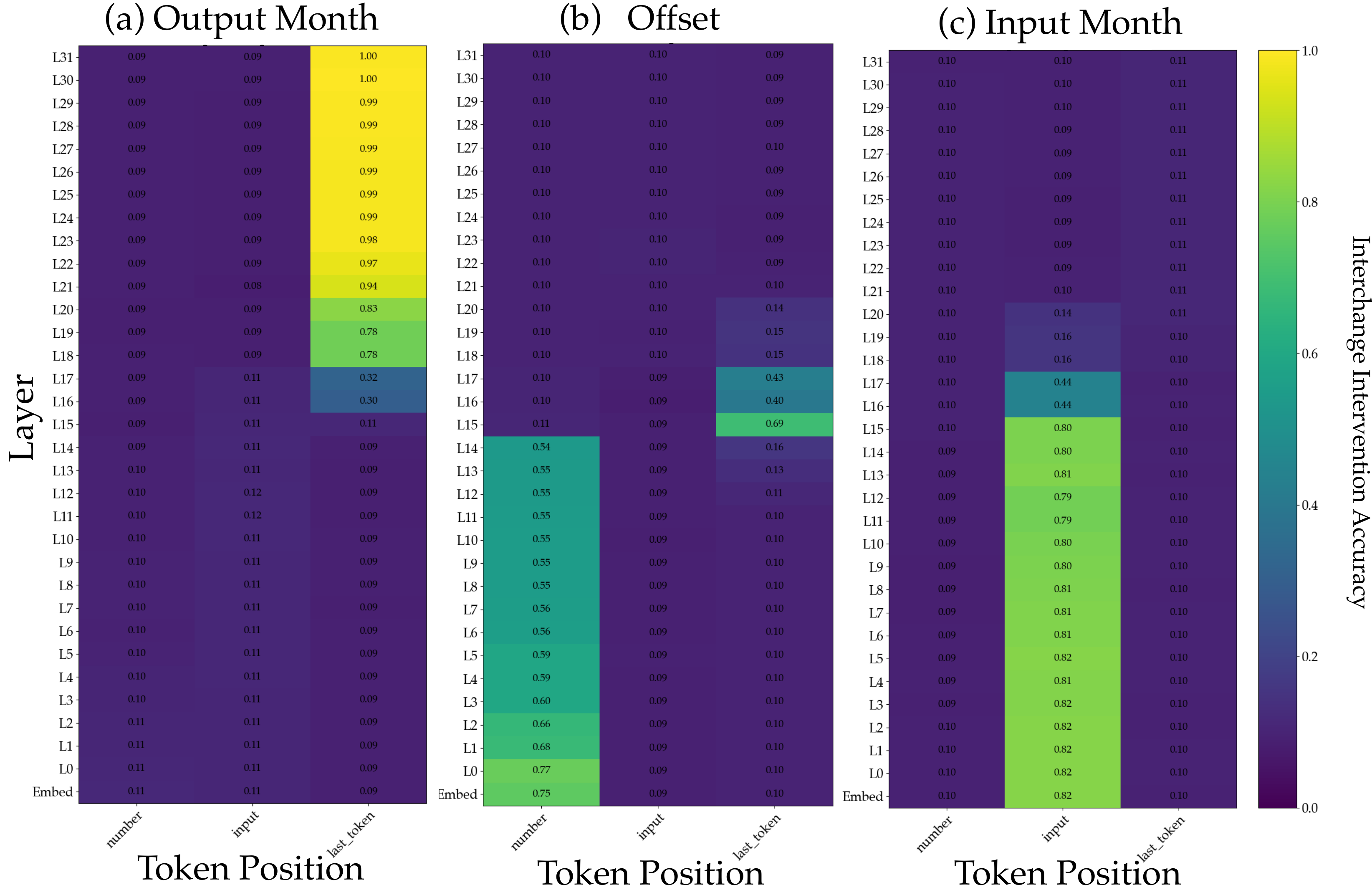}
    \caption{Residual stream patching results for the \texttt{months} task for each causal variable. Patching is done for $n=4096$ counterfactual pairs (Section~\ref{app:das-setup}). See Figure~\ref{fig:weekdays-interchange} for interpretation of results. The template for this task is \textit{Q: What month is \{number\} months after \{input\}?\textbackslash{}nA:} }
    \label{fig:months-interchange}
\end{figure}

\begin{figure}[h]
    \centering
    \includegraphics[width=0.8\linewidth]{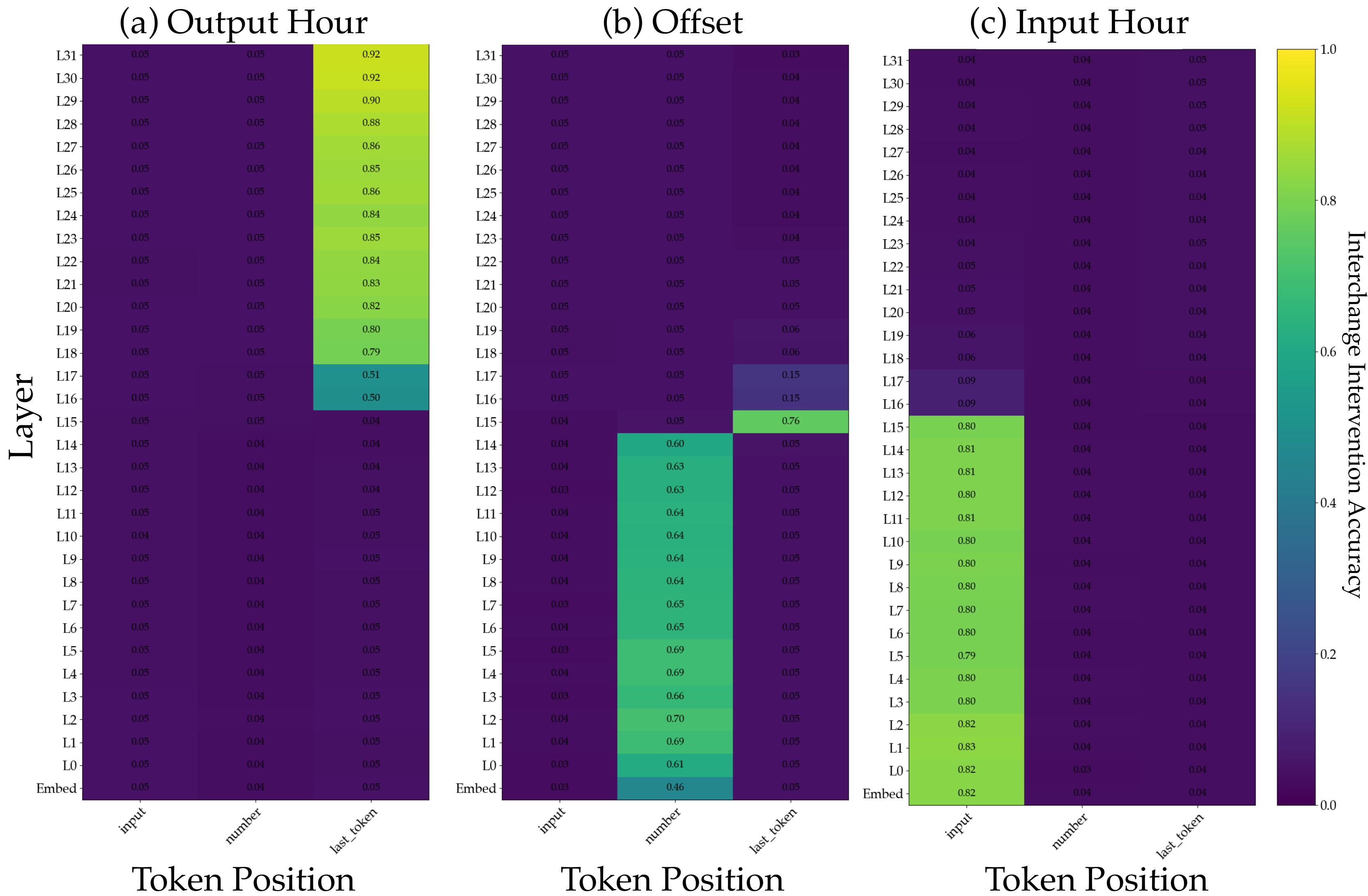}
    \caption{Residual stream patching results for the \texttt{hours} task for each causal variable. Patching is done for $n=4096$ counterfactual pairs (Section~\ref{app:das-setup}). See Figure~\ref{fig:weekdays-interchange} for interpretation of results. The template for this task is \textit{Q: In 24-hour time, it is now 10:00. What time will it be in twenty-four hours?\textbackslash{}nA: In 24-hour time, it will be } }
    \label{fig:hours-interchange}
\end{figure}

\clearpage
\section{Distributed Alignment Search}\label{app:das}

\subsection{Training Details}\label{app:das-setup}
To train distributed alignment search (DAS) on the \texttt{months}, \texttt{weekdays}, \texttt{hours}, and \texttt{addition} tasks, we create datasets of 4096 randomly-sampled pairs of prompts per task, sampled from prompts that \llama\ answers correctly (Table~\ref{tab:task-templates}). Setting aside $n_{\text{test}}$ random pairs for testing, we train on $n_{\text{train}}$ counterfactual pairs, with $n_{\text{test}}=512$, $n_{\text{train}}=3584$. 

With \texttt{weekdays} as a representative task, we run a hyperparameter sweep with $k=8$ over epochs $\in$ \{4, 8\}, learning rate $\in$ \{0.0001, 0.001, 0.01\}, and batch size $\in$ \{16, 32, 64\} for the \inputconcept\ variable at layer 18 after attention and before the MLP. This is a position we know to be important from initial exploratory runs. From this sweep, for \texttt{weekdays}, we choose epochs=8, lr=0.0001, and bsz=16. We then run DAS for every task with those particular hyperparameters. 

To stabilize optimization, subspace basis vectors are initialized within the space spanned by the top principal components explaining $\geq90\%$ of the variance in the training set.

\subsection{Intervention Details}

The simplest causal model we can construct for this task consists of the \inputconcept\ (e.g., \textit{January}), \inputnumber\ (e.g., \textit{three}), and \outputconcept\ (e.g., \textit{three months after January is \textbf{April}}). Our goal is to isolate subspaces of the last token residual stream that ``contain'' each of these three causal variables: 

\begin{itemize}[topsep=2pt, itemsep=4pt, parsep=0pt]
    \item Patching the \inputconcept\ should change the input month: e.g. \textit{three months after \textbf{January}}$\rightarrow$\textit{one month after \textbf{July}} should yield \textit{one month after \textbf{January}=February}.
    \item Patching the \inputnumber\ should change the input offset: e.g. \textit{\textbf{three} months after January}$\rightarrow$\textit{\textbf{one} month after July} should yield \textit{\textbf{three} months after July=October}.
    \item Patching the \outputconcept\ should change the entire output: e.g. \textit{three months after January=\textbf{April}}$\rightarrow$\textit{one month after July=\textbf{August}} should yield \textit{one month after January=\textbf{April}}.
\end{itemize}

\begin{figure}
    \centering
    \includegraphics[width=\linewidth]{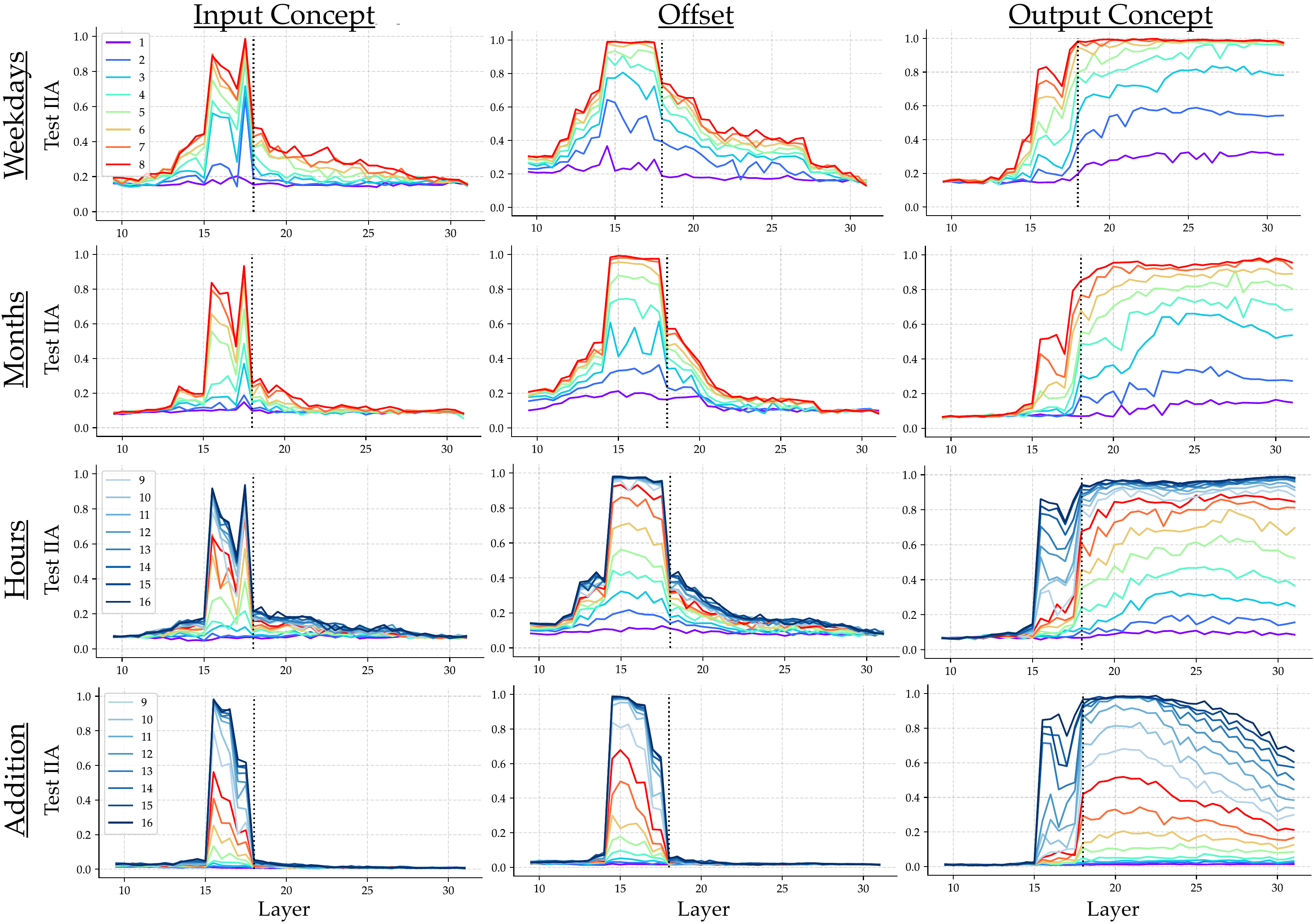}
    \caption{DAS results for every dimension across layers, alternating between sublayer (post attention) and layer output (post MLP). We include $8<k\leq16$ for \texttt{hours} and \texttt{addition} in blue, as $k\leq8$ does not achieve maximum IIA.}
    \label{fig:coarse-das-all}
\end{figure}

Focusing on the last token position, we train DAS for each causal variable at every layer $0\leq l\leq31$ and subspace dimension $1\leq k\leq8$. If performance does not break 90\% at any point for a particular task, we add eight more dimensions, expanding to $1\leq k\leq 16$ for the \texttt{hours} and \texttt{addition} tasks. Results are plotted in Figure~\ref{fig:coarse-das-all}. Immediately before the layer 18 MLP (dotted black line), input variables are cleanly separable in low-dimensional subspaces; immediately after, the \outputconcept\ becomes patchable. This implies that across all tasks, the layer 18 MLP implements a major piece of computation. 

\subsection{Best DAS Subspaces}
Pre-MLP DAS subspaces are trained on the residual stream immediately after attention has been added, but before LayerNorm. Post-MLP DAS subspaces are trained on the residual stream at the output of the layer, immediately after the MLP output has been added.

\begin{table}[H]
\centering
\caption{Best DAS runs for every task $\times$ causal variable at layer 18. All subspaces are matrices $\mathbf{R}\in\mathbb{R}^{d\times k}$. We use these subspaces for all experiments throughout the paper.}
\label{tab:das-results}
\begin{tabular}{llllcc}
\toprule
Task & Variable & Pre/Post-MLP & Best Dim. $k$ & Test IIA \\
\midrule
\multirow{3}{*}{Months}
  & \inputconcept  & Pre-MLP & $k=8$ & 93.4\% \\
  & \inputnumber & Pre-MLP & $k=8$ & 97.5\% \\
  & \outputconcept & Post-MLP & $k=8$ & 85.2\% \\
\midrule
\multirow{3}{*}{Weekdays}
  & \inputconcept  & Pre-MLP & $k=8$ & 98.7\% \\
  & \inputnumber & Pre-MLP & $k=8$ & 98.5\% \\
  & \outputconcept & Post-MLP & $k=8$ & 98.3\% \\
\midrule
\multirow{3}{*}{Hours}
  & \inputconcept  & Pre-MLP & $k=16$ & 94.0\%  \\
  & \inputnumber & Pre-MLP & $k=14$ & 95.6\% \\
  & \outputconcept & Post-MLP & $k=16$ & 94.3\% \\
\midrule
\multirow{3}{*}{Addition}
  & \inputconcept  & Pre-MLP & $k=16$ & 61.7\% \\
  & \inputnumber & Pre-MLP & $k=15$ & 63.8\% \\
  & \outputconcept & Post-MLP & $k=16$ & 96.6\% \\
\bottomrule
\end{tabular}
\end{table}

\subsection{Overlap Between DAS Subspaces}\label{app:principal-angles}

To get a measure of how much the best subspaces for each task/causal variable overlap, we use principal angles. Let $\mathbf{A}\in\mathbb{R}^{d\times m}$ be a matrix with orthonormal columns spanning an $m$-dimensional subspace, and $\mathbf{B}\in\mathbb{R}^{d\times n}$ be a matrix with orthonormal columns spanning an $n$-dimensional subspace. We can calculate the principal angles between these subspaces by taking the singular values $\sigma_1...\sigma_{\min(m,n)}$ of $\mathbf{A}^T\mathbf{B}$. The first singular value $\sigma_1$ corresponds to the cosine of the smallest angle between the closest vectors in each subspace; $\sigma_2$ is the next smallest angle orthogonal to the first, and so on. 

\begin{figure}[t]
    \centering
    \begin{subfigure}[t]{\textwidth}
        \centering
        \includegraphics[width=0.8\textwidth]{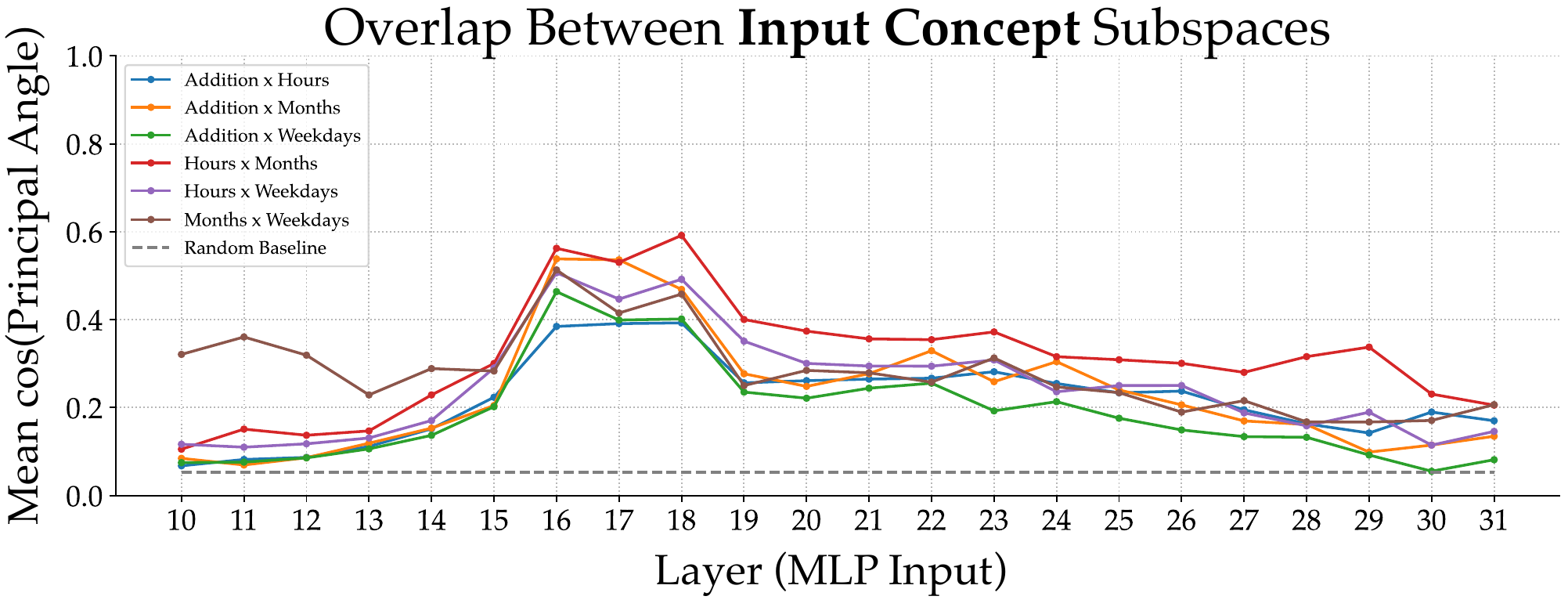}
    \end{subfigure}
    
    \begin{subfigure}[t]{\textwidth}
        \centering
        \includegraphics[width=0.8\textwidth]{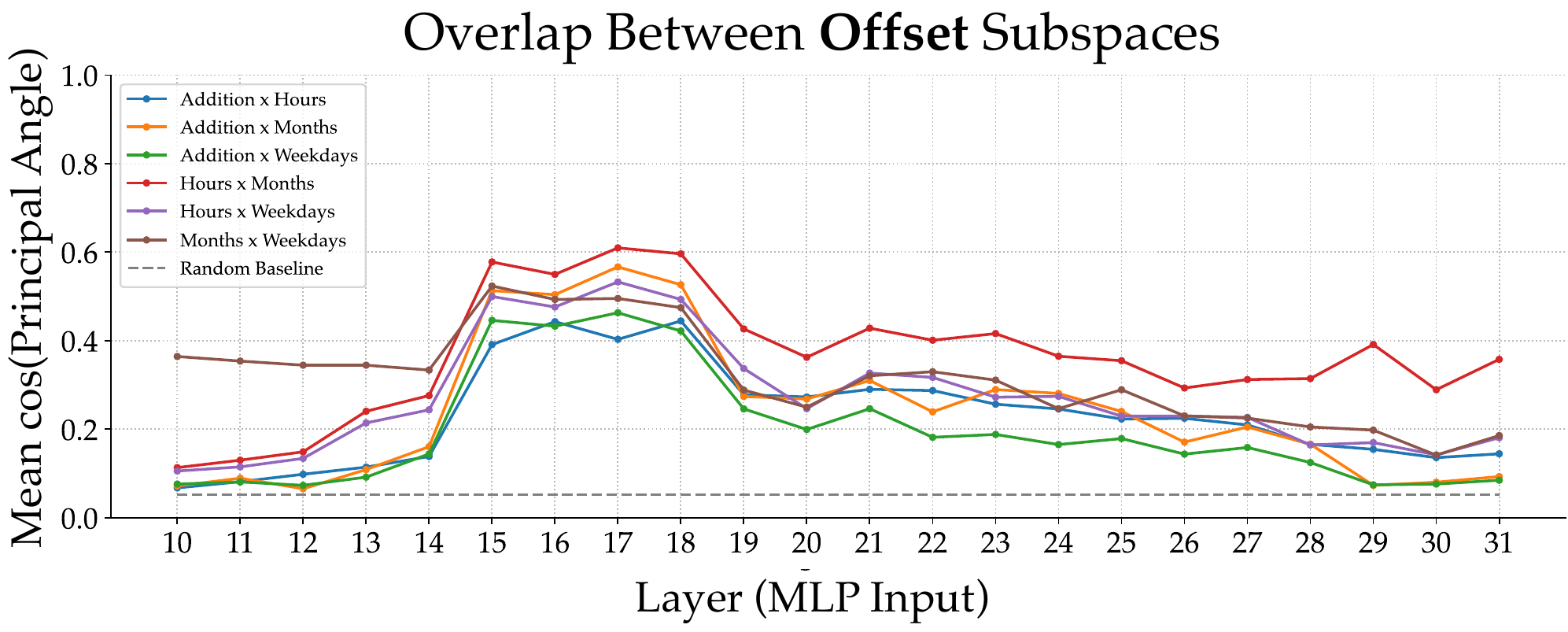}
    \end{subfigure}
    
    \begin{subfigure}[t]{\textwidth}
        \centering
        \includegraphics[width=0.8\textwidth]{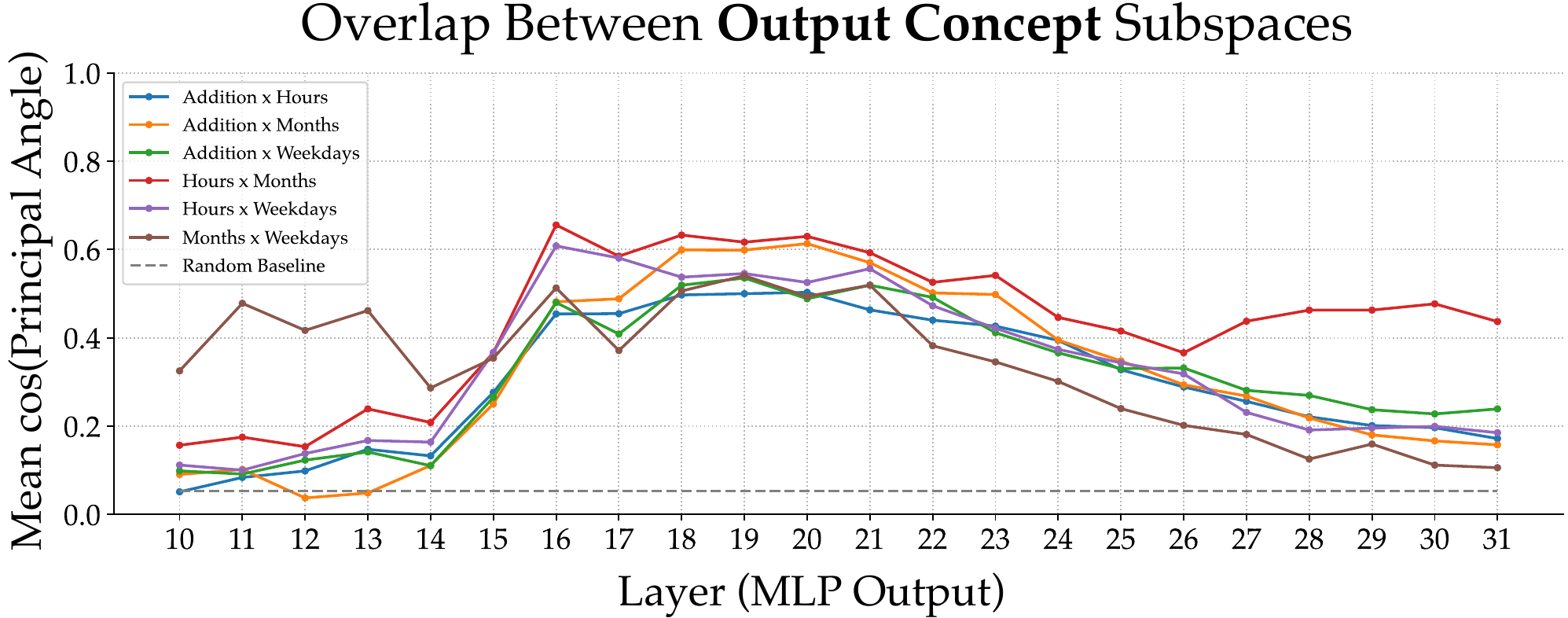}
    \end{subfigure}
    \caption{Average cosines of principal angles between subspaces for each causal variable in different tasks. Right before the important computation at layer 18, \inputconcept\ and \inputnumber\ overlap significantly above change. Subspaces for \outputconcept\ also increase in overlap in these layers.}
    \label{fig:principal-angles}
\end{figure}

We take the average of all of these singular values and report them in Figure~\ref{fig:principal-angles}. If this metric is 0, the subspaces are orthogonal; if it is 1, they are identical (or the smaller one is fully contained in the larger one). We observe that similarity for \outputconcept\ subspaces increases around layers 16-20, much above a baseline between 1000 pairs of random subspaces with $m,n=16$. Subspaces also overlap for \inputconcept\ and \inputnumber\ variables. 

\section{Weekday Alignment with Numbers}\label{app:weekdays}

\begin{figure}
    \centering
    \includegraphics[width=\linewidth]{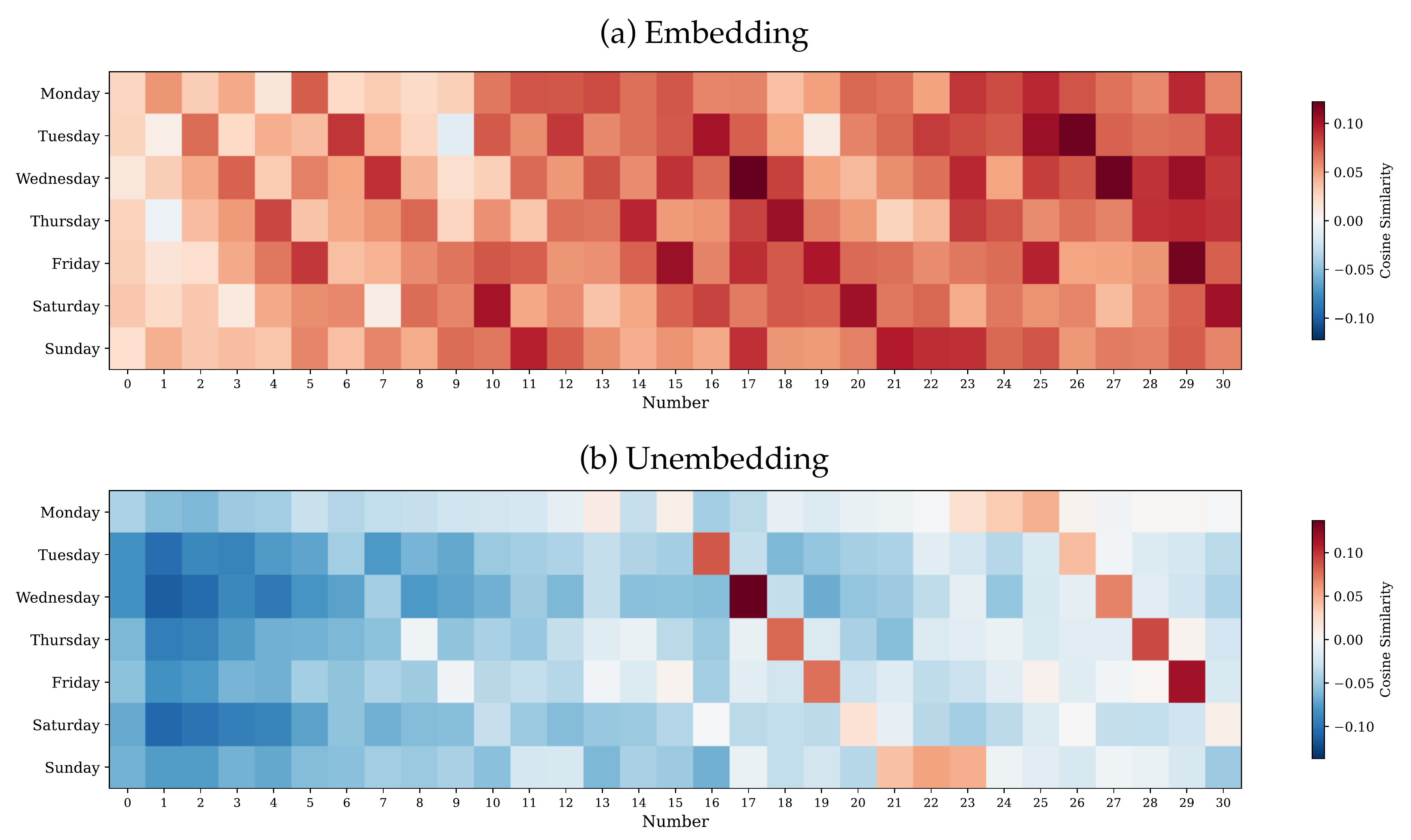}
    \caption{Alignment between weekday/number tokens for \llama\ with (a) token embeddings and (b) token unembeddings. }
    \label{fig:weekdays_aligment}
\end{figure}

Although we find strong evidence that addition is used in computation for the \texttt{weekdays} task, the exact alignment between weekdays and numbers is unclear. We hypothesize that this is due to a lack of a ``canonical'' enumeration of weekdays, which causes noise/misalignment in the embedding and unembedding spaces. Figure~\ref{fig:weekdays_aligment} shows cosine similarities between weekday tokens and number tokens up to 30: we observe that alignment in the embedding space is very noisy. Although we do observe diagonal patterns in weekday alignment with numbers, there are odd offsets: for example, Wednesday is most aligned with the number 17 in the unembedding space. In our experiments, we account for this misalignment in a few ways. 

\begin{figure}[H]
    \centering
    \includegraphics[width=\linewidth]{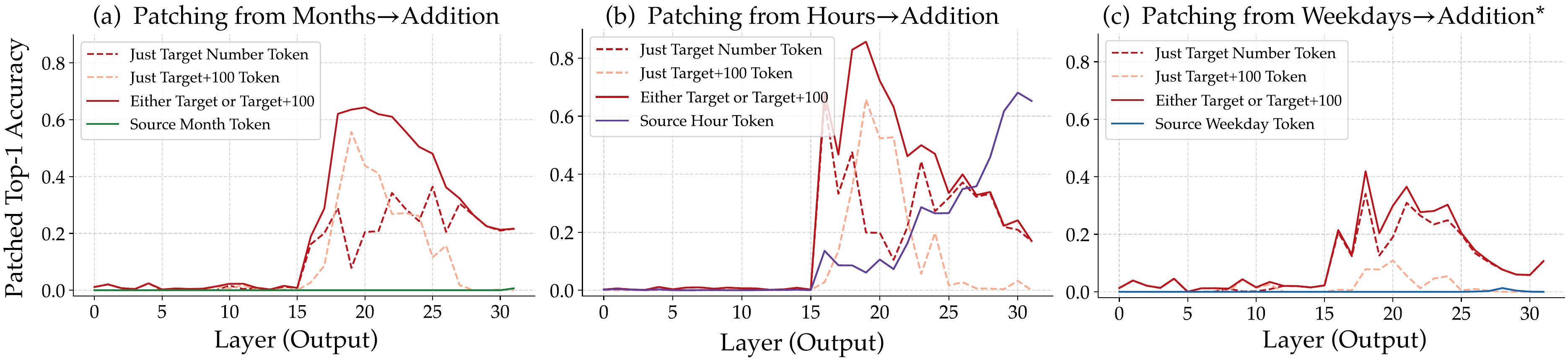}
    \caption{Accuracy when patching \textbf{from cyclic tasks into \texttt{addition}}: this intervention ``exposes'' a sum computed in the forward pass of a cyclic prompt (e.g., \textit{three months after November=3+11=14}). (a) Patching from \texttt{months}$\rightarrow$\texttt{addition}. More than 60\% of the time, this intervention exposes the pre-modulo sum, but sometimes the highest probability is that sum + 100. See Figure~\ref{fig:months-to-addition-appx} for heatmaps. (b) Patching from \texttt{hours}$\rightarrow$\texttt{addition}. In intermediate layers, this exposes the pre-modulo sum, but patching too late brings over the modded hour token. See Figure~\ref{fig:hours-to-addition-appx} for heatmaps. (c) Patching from \texttt{weekdays}$\rightarrow$\texttt{addition}. Weekdays are misaligned---these accuracies are based on an offset of +4, where Monday=5.}
    \label{fig:all-everything-to-addition}
\end{figure}
\begin{figure}
    \centering
    \includegraphics[width=0.8\linewidth]{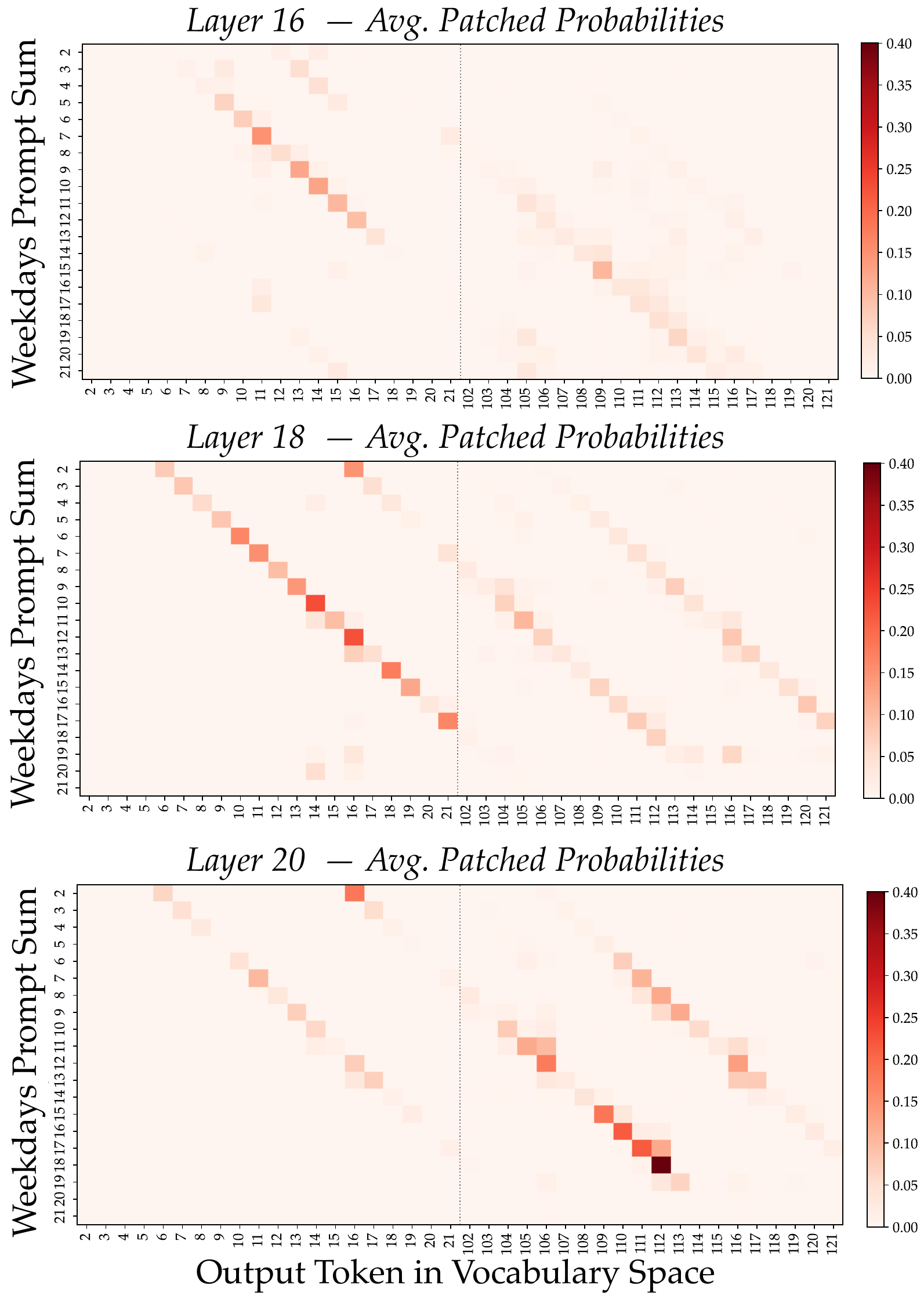}
    \caption{Patching from \texttt{weekdays}$\rightarrow$\texttt{addition} at layers 16, 18, and 20. Note that patching is most consistent at layer 18, and that we observe a strange +4 offset for \texttt{weekdays}:  see App.~\ref{app:weekdays} for discussion. See Figure~\ref{fig:all-everything-to-addition}(b) for performance across all layers.}
    \label{fig:weekdays-to-addition-appx}
\end{figure}

When patching from \texttt{weekdays} into \texttt{addition} (Figure~\ref{fig:all-everything-to-addition}c, Figure~\ref{fig:weekdays-to-addition-appx}), we find that probability heatmaps have a consistent diagonal pattern, but shifted with an offset of +4 (i.e., when the \texttt{addition} sum is 6, the model outputs 10, and so on). This means that when we patch the other way, from \texttt{addition} to \texttt{weekdays}, we must shift our expectations: instead of assuming that 1 will map to Monday, we assume that the model computes sums for the weekdays task with an offset of four, meaning that 1+4=5 will actually map to Monday. Although this assumption is imperfect, it appears to predict model behavior for this experiment reasonably well (Figure~\ref{fig:addition-to-weekdays-appx}). 

This +4 shift also appears consistent with the fact that the unembedding vector for, e.g., Tuesday is most similar to the number token 16 (Figure~\ref{fig:weekdays_aligment}). If Tuesday was the number 2 under our initial incorrect assumptions, then a shift of +4 to match the model would give us 6. Due to the fact that \llama\ mainly uses periodicities $T\in\{2,5,10\}$ for the \texttt{weekdays} task (Tab.~\ref{tab:steering_periods}, Figure~\ref{fig:unclipped-ribbons}), the model would not be able to distinguish between 6, 16, 26, etc.

For Fourier steering, we show results when defining Saturday as 0, which we empirically find to have the most consistent results. We do not have a good explanation, other than the fact that this would possibly imply that Sunday is 1, which appears to be a plausible enumeration. If the model is ignoring the tens place in these calculations, then this would also be consistent with Saturday being similar to 20 and 30 in the unembedding space (Figure~\ref{fig:weekdays_aligment}). 

We acknowledge that this is messy, but reiterate that the addition neurons we find \textit{are} causally important for the \texttt{weekdays} task (Tab.~\ref{tab:l18-neuron-ablation})---we are quite sure that these neurons (which perform addition cleanly for all other tasks) are being used to compute, e.g., \textit{three days after Friday}, but do not claim to understand \llama's internal logic for how it enumerates weekdays.

\clearpage
\section{Cross-Task Patching}\label{app:cross-task}

When we ``patch within the union of subspaces,'' this simply means that we concatenate the columns spanning the source and target subspaces to obtain a new, larger subspace. For example, for a source subspace $\mathbf{A}\in\mathbb{R}^{d_{\text{model}}\times m}$ and a target subspace $\mathbf{B}\in\mathbb{R}^{d_{\text{model}}\times n}$, we build a new matrix $\mathbf{C}=[\mathbf{A}\ \mathbf{B}]\in\mathbb{R}^{d_{\text{model}}\times (m+n)}$. We then orthogonalize the columns of $\mathbf{C}$ and use it to patch from source to target prompts following Eq.~\ref{eq:das}.

\begin{figure}[H]
    \centering
    \includegraphics[width=\linewidth]{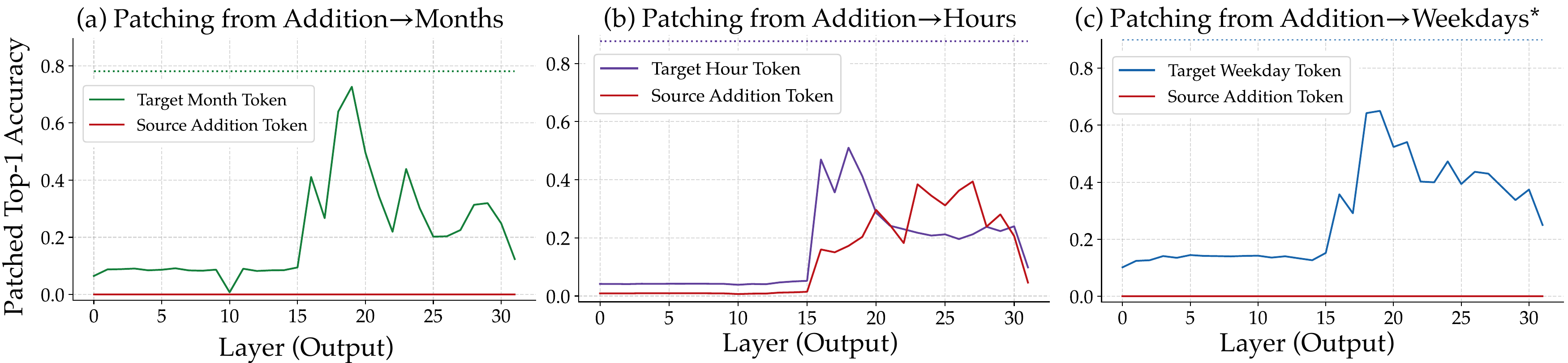}
    \caption{\textit{[Duplicate of Figure~\ref{fig:addition-to-everything} for easy cross-reference.]} Accuracy when patching \textbf{from \texttt{addition} into cyclic tasks}: this intervention forces the model to ``take the modulo'' of a sum calculated in an addition setting, mapping that value to a particular output concept (e.g., \textit{5+9=14$\rightarrow$February}). We patch within the union of subspaces for both tasks. Because the clean model's accuracy begins to break down for larger numbers (Figure~\ref{fig:output-probs}), we limit \texttt{addition} prompts to those with sum $\leq2p$. (a) See heatmaps in Figure~\ref{fig:addition-to-months-appx}. (b) See heatmaps in Figure~\ref{fig:addition-to-hours-appx}. (c) See heatmaps in Figure~\ref{fig:addition-to-weekdays-appx}. Again, as weekdays are misaligned, we use an offset of +4.}
    \label{fig:all-addition-to-everything}
\end{figure}

\begin{figure}
    \centering
    \includegraphics[width=\linewidth]{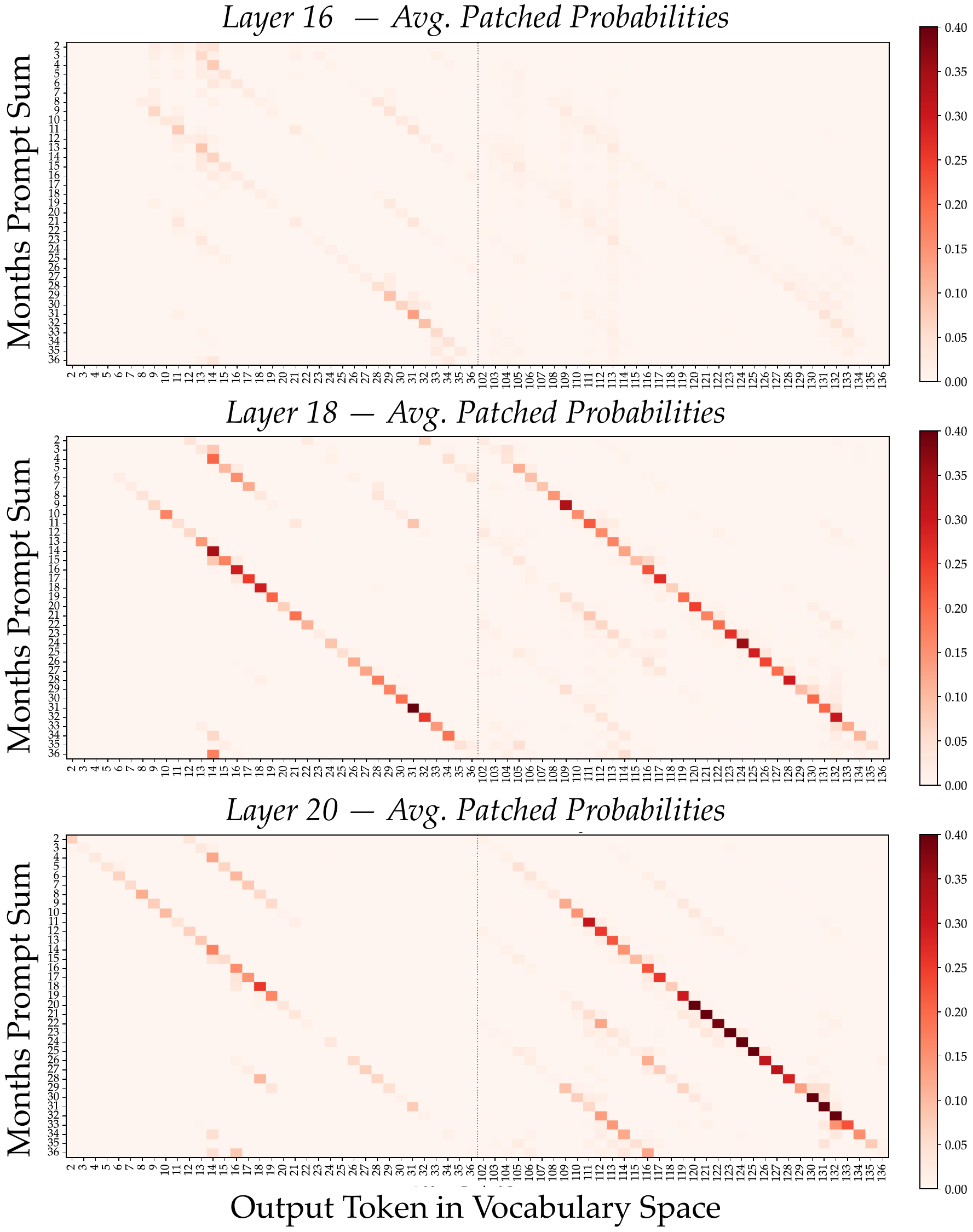}
    \caption{Patching from \texttt{months}$\rightarrow$\texttt{addition} at layers 16, 18, and 20. Note that patching is most consistent at layer 18; this plot is the same as Figure~\ref{fig:months-to-addition}. See Figure~\ref{fig:all-everything-to-addition}(a) for performance across all layers.}
    \label{fig:months-to-addition-appx}
\end{figure}

\begin{figure}
    \centering
    \includegraphics[width=0.9\linewidth]{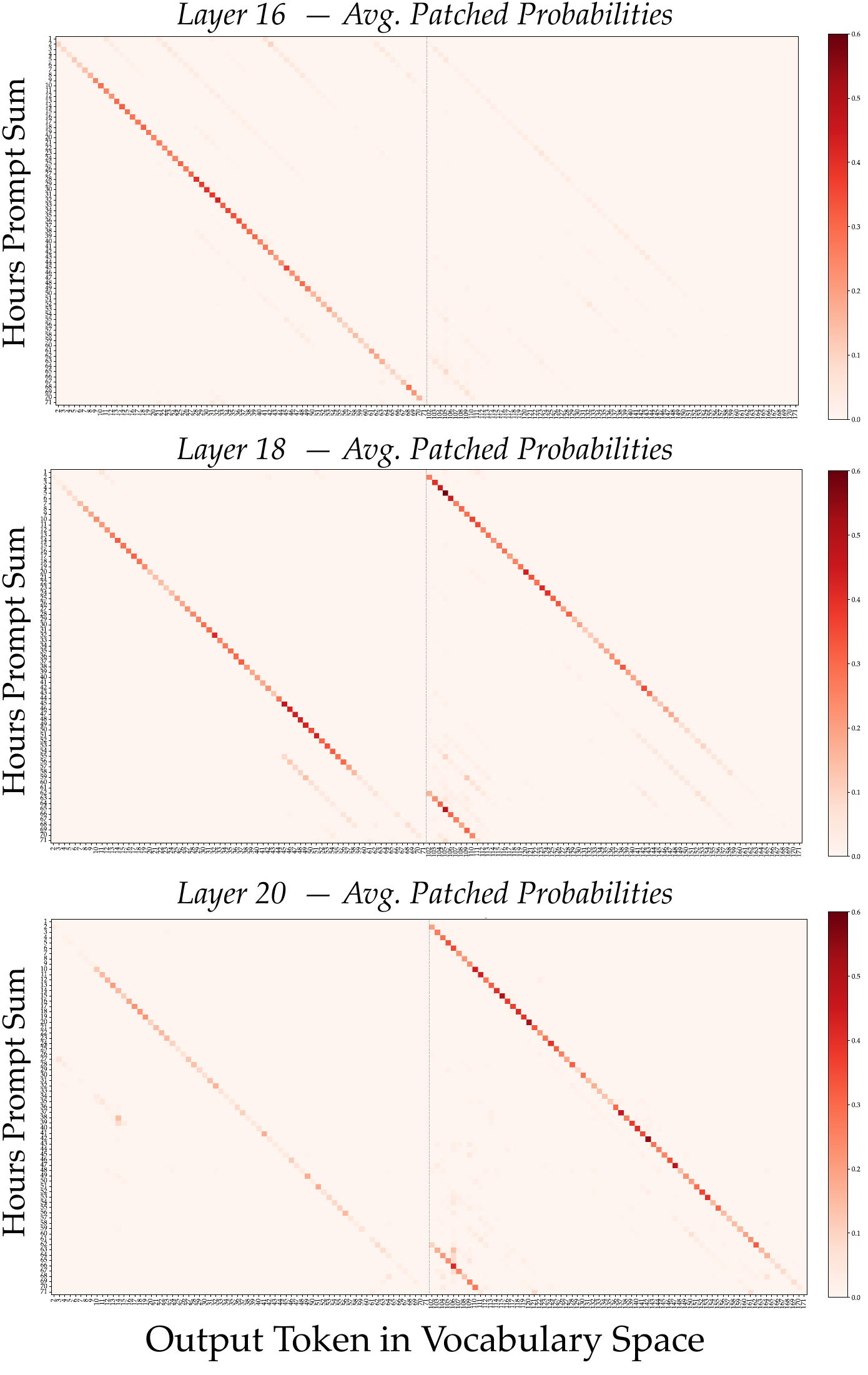}
    \caption{Patching from \texttt{hours}$\rightarrow$\texttt{addition} at layers 16, 18, and 20. Surprisingly, patching is also quite effective at layer 16 for this task, implying that the input representations for \texttt{hours} transfer well to \texttt{addition}; this may be because \texttt{hours} uses literal number tokens. See Figure~\ref{fig:all-everything-to-addition}(b) for performance across all layers.}
    \label{fig:hours-to-addition-appx}
\end{figure}

\begin{figure}
    \centering
    \includegraphics[width=0.8\linewidth]{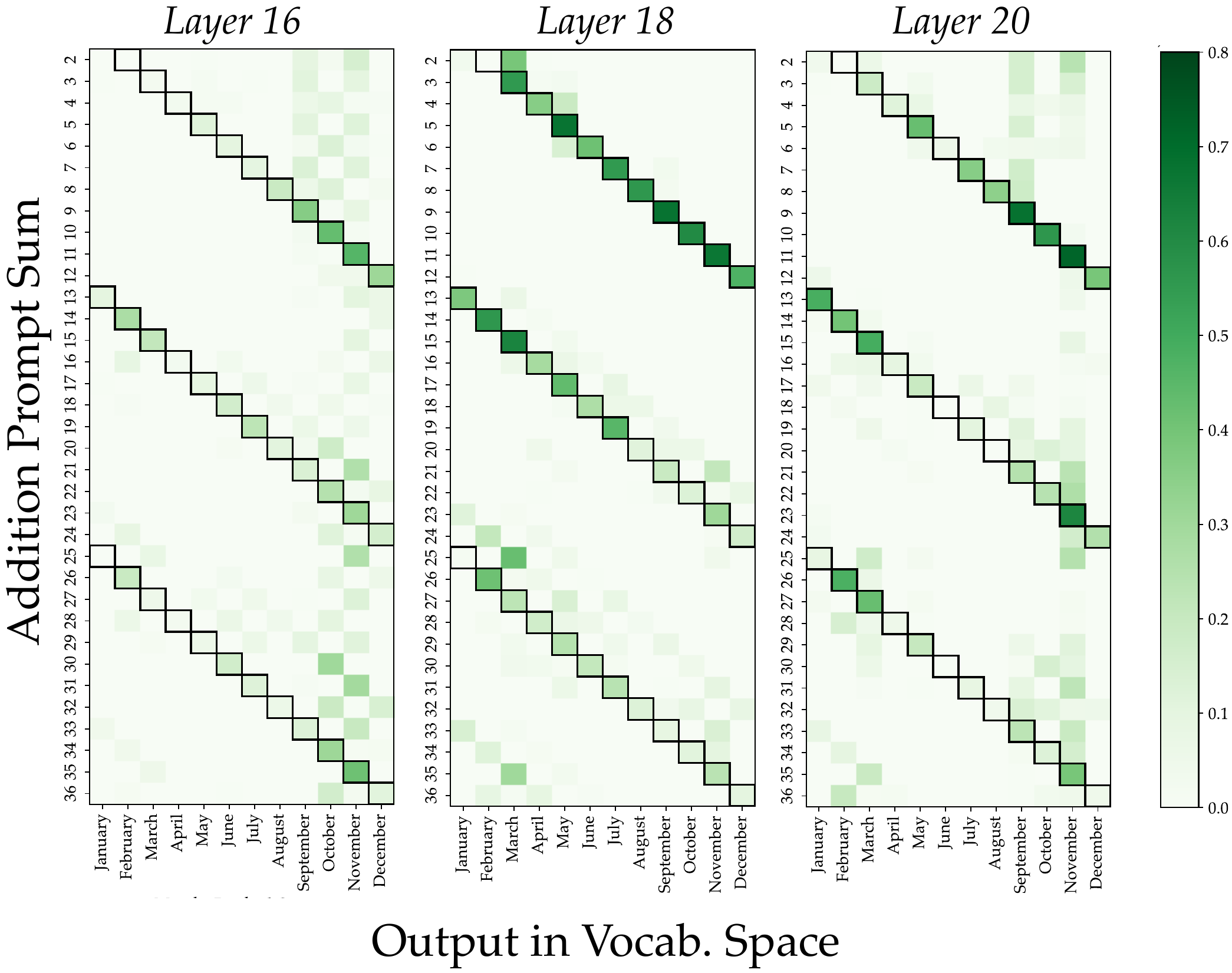}
    \caption{Patching from \texttt{addition}$\rightarrow$\texttt{months} at layers 16, 18, and 20. Note that patching works best at layer 18, and that performance begins to break down as the sum increases, closely matching errors in a clean forward pass in Figure~\ref{fig:output-probs}. See Figure~\ref{fig:all-addition-to-everything}(a) for performance across all layers.}
    \label{fig:addition-to-months-appx}
\end{figure}

\begin{figure}
    \centering
    \includegraphics[width=0.8\linewidth]{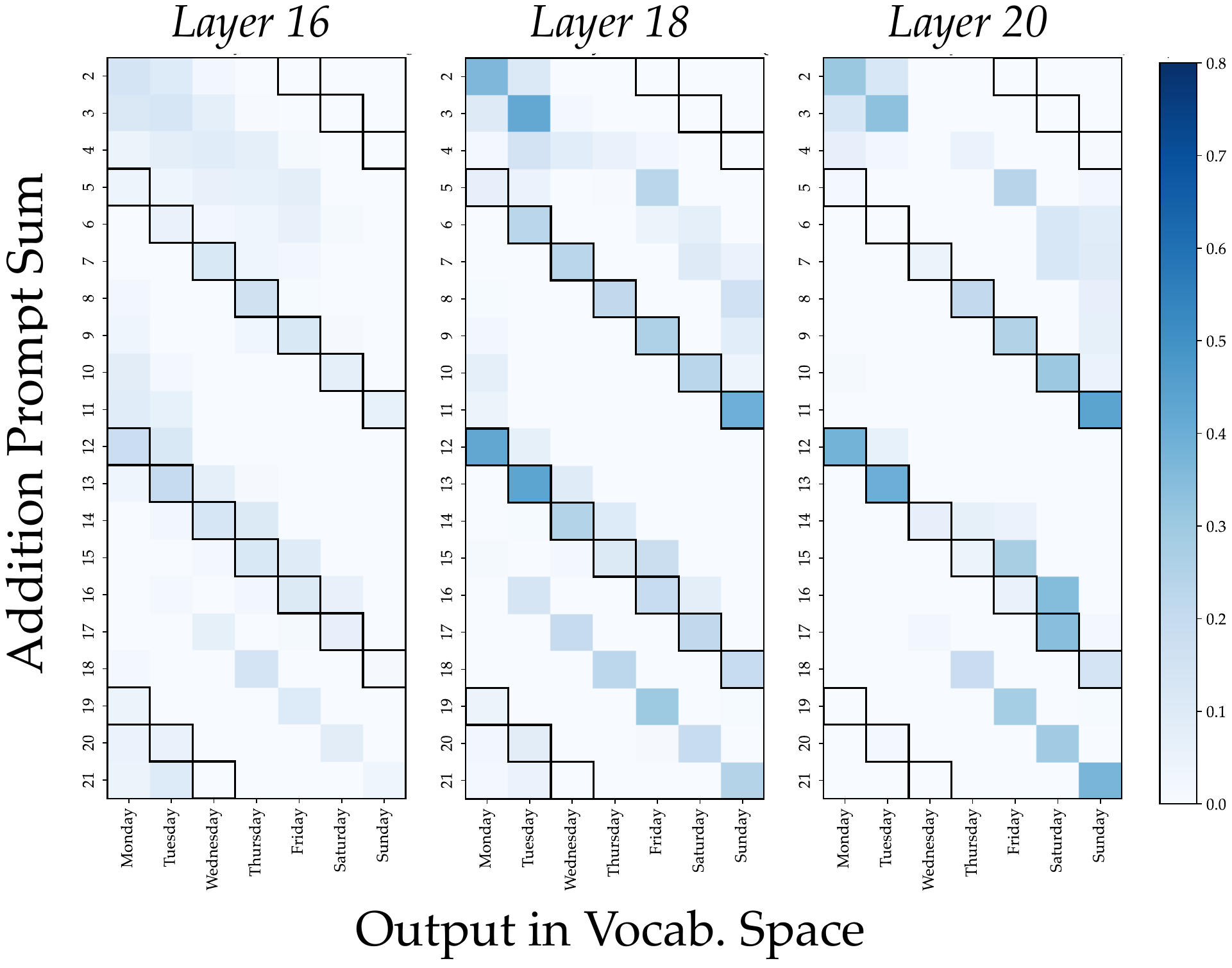}
    \caption{Patching from \texttt{addition}$\rightarrow$\texttt{weekdays} at layers 16, 18, and 20. Note that patching works best at layer 18, and that performance begins to break down as the sum increases, closely matching errors in a clean forward pass in Figure~\ref{fig:output-probs}. See Figure~\ref{fig:all-addition-to-everything}(c) for performance across all layers. We define Monday=5; see App.~\ref{app:weekdays} for discussion.}
    \label{fig:addition-to-weekdays-appx}
\end{figure}

\begin{figure}
    \centering
    \includegraphics[width=0.9\linewidth]{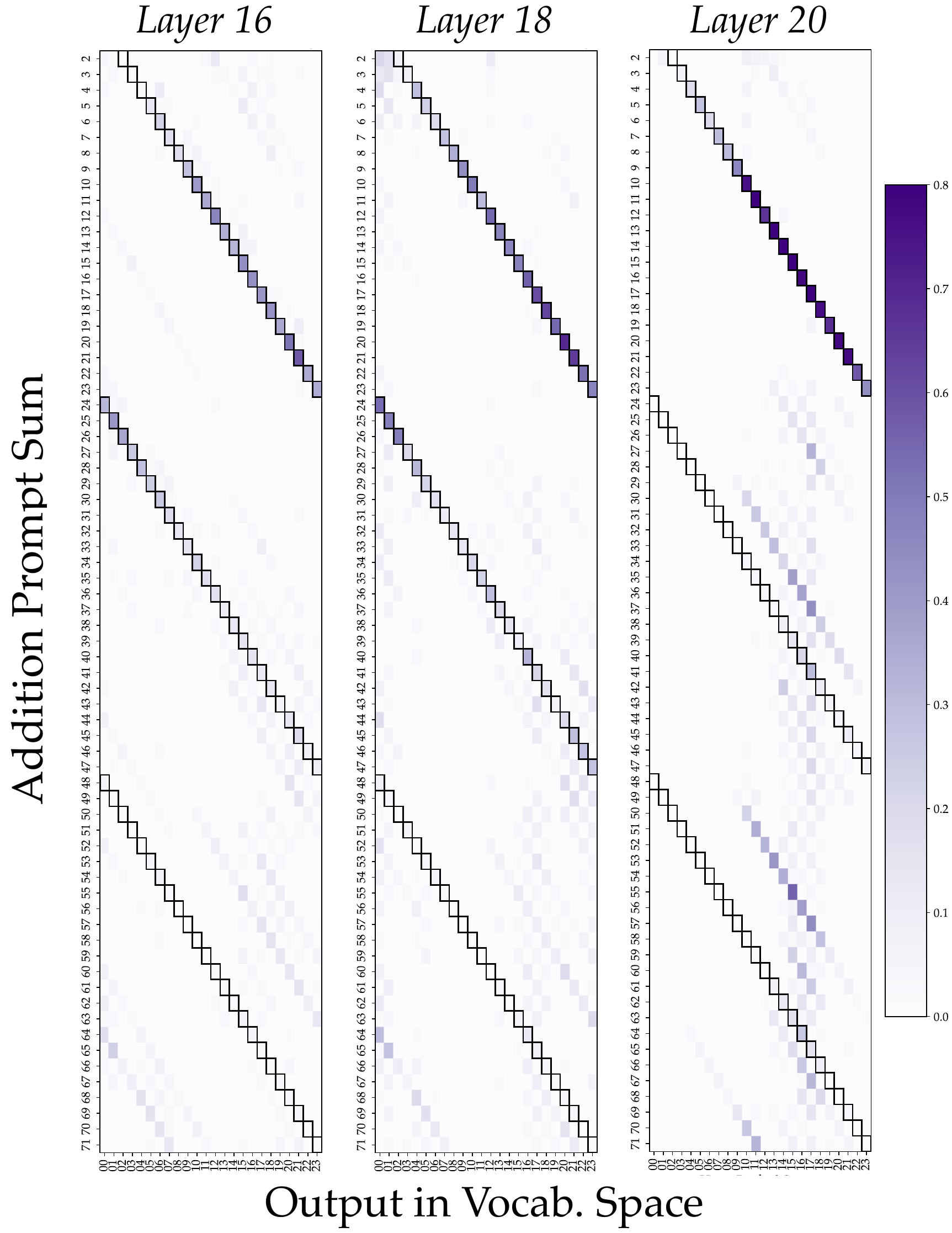}
    \caption{Patching from \texttt{addition}$\rightarrow$\texttt{hours} at layers 16, 18, and 20. See Figure~\ref{fig:all-addition-to-everything}(b) for performance across all layers. }
    \label{fig:addition-to-hours-appx}
\end{figure}

\clearpage
\section{Fourier Probes}
\label{app:fourier_probes}

\subsection{Fourier Probes Training}
\label{app:fourier_training}

The probes are trained on the addition task using the template \texttt{"\{a\}+\{b\}="}, where \(a, b \in \{1, \ldots, 199\}\). 

Let \(\mathbf{h}_{a + b}^{(l)} \in \mathbb{R}^{4096}\) denote the hidden state at layer \(l\) at the final tokenn position. For each period \(1 \leq T \leq 150\), we train two affine probes to follow sine and cosine harmonics by minimizing the following losses:

\begin{equation}
    \mathsf{MSE}\Bigg( \langle\mathbf{w}_{\sin}^{(l,T)}, \mathbf{h}^{(l)}_{a + b}\rangle + b_{\sin}^{(l,T)}, \sin\!\left(\tfrac{2\pi(a+b)}{T}\right)\Bigg),  \quad
    \mathsf{MSE}\Bigg(\langle\mathbf{w}_{\cos}^{(l,T)}, \mathbf{h}^{(l)}_{a+b}\rangle + b_{\cos}^{(l,T)},\cos\!\left(\tfrac{2\pi(a+b)}{T}\right)\Bigg).
\end{equation}

Each probe was trained independently for 500 epochs using Adam with learning rate \(10^{-3}\).

\begin{figure}[t!]
    \centering
    \includegraphics[width=\linewidth]{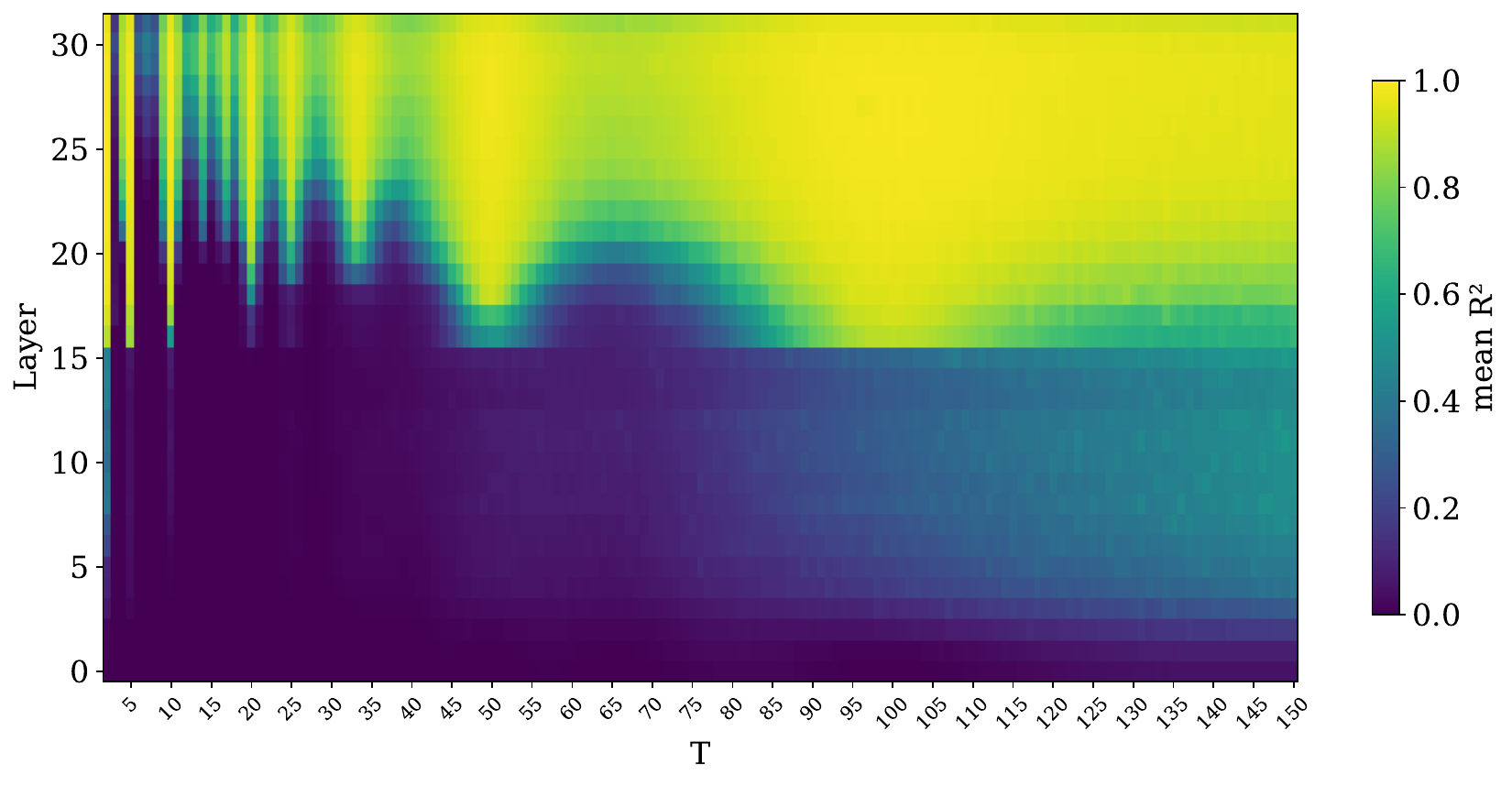}
    \caption{$R^2$ scores for Fourier probes across layers and for each \(T \in \{2, \dots, 150\}\). For each \(T\), we train sine and cosine probes and report the average \(R^2\).}
\label{fig:R2_probes}
\end{figure}

Figure~\ref{fig:R2_probes} shows the \(R^2\) scores across layers and periods \(T\). Fourier features begin to appear around layer 15, which, according to the DAS experiments in Section~\ref{sec:das}, is where the model passes operand \(a\) and operand \(b\) to the final token position and begins computing the output sum. We observe significant $R^2$ values for $T \in \{2, 5, 10, 20\}$. For larger periods, the signal remains strong but becomes more diffuse. Following \citet{kantamnenihelix}, we also focus on $T \in \{50, 100\}$, as these periods exhibit high $R^2$ and align with the inductive bias of a base-10 number system.

\begin{figure}[h]
    \centering
    \includegraphics[width=0.8\linewidth]{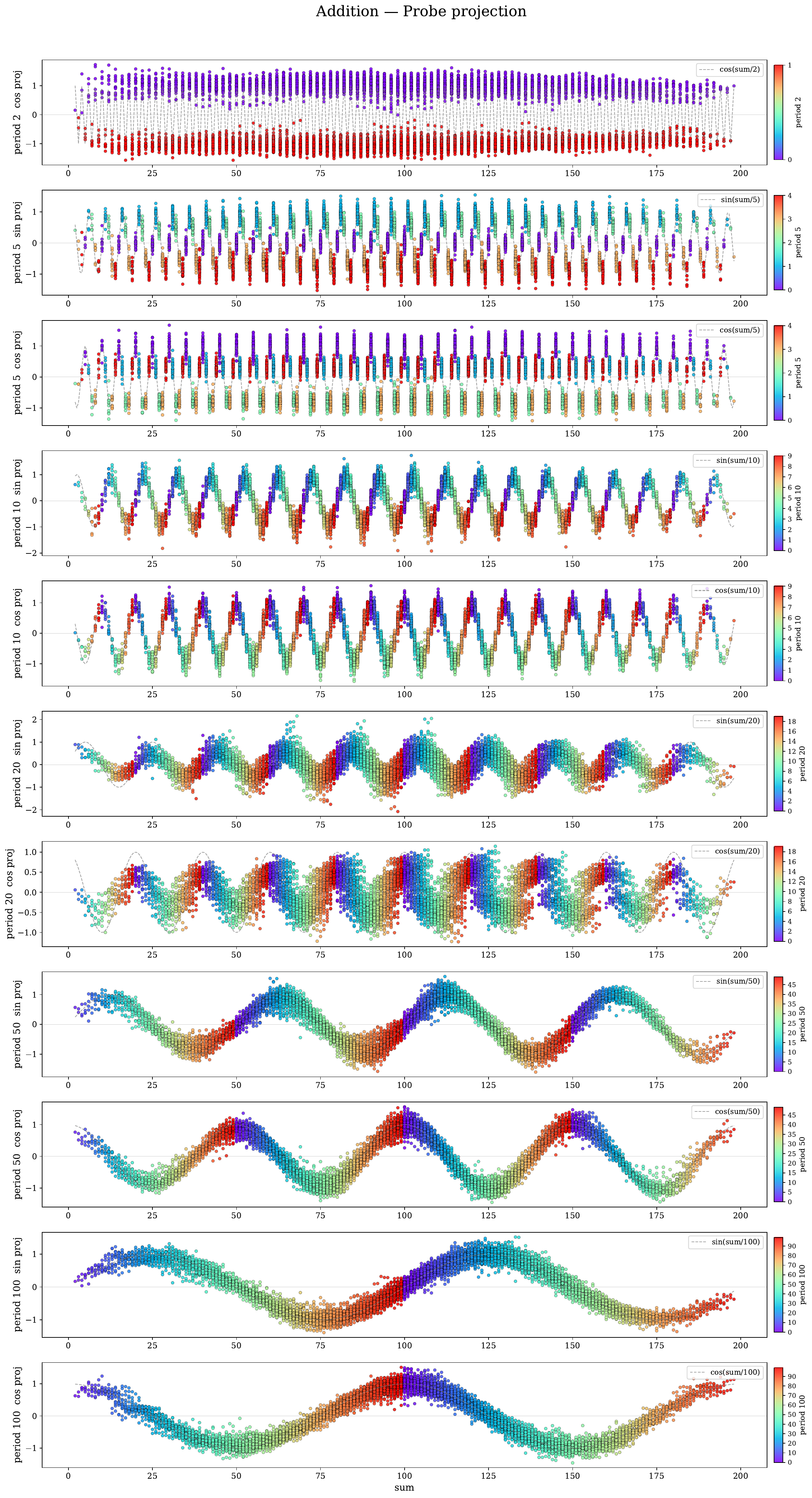}
   \caption{Projection of the layer 18 residual activations at the final token position onto the Fourier probe directions.}
    \label{fig:summation_direction}
\end{figure}

\begin{figure}[h]
    \centering
    \includegraphics[width=0.9\linewidth]{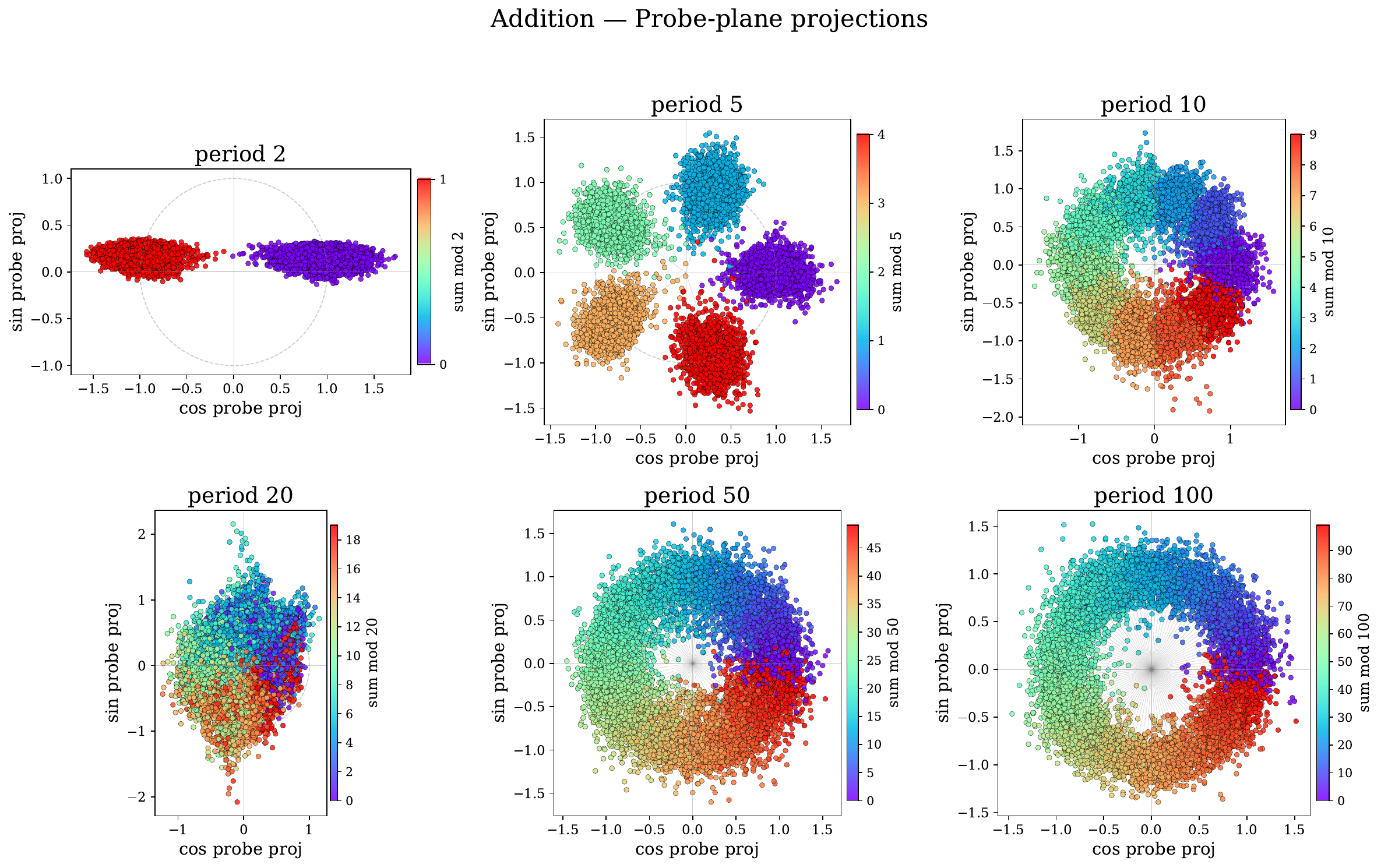}
    \caption{Projection of the layer 18 residual activations at the final token position onto the Fourier probe planes.
}
    \label{fig:summation_plane}
\end{figure}

\begin{figure}[h]
    \centering
    \includegraphics[width=0.9\linewidth]{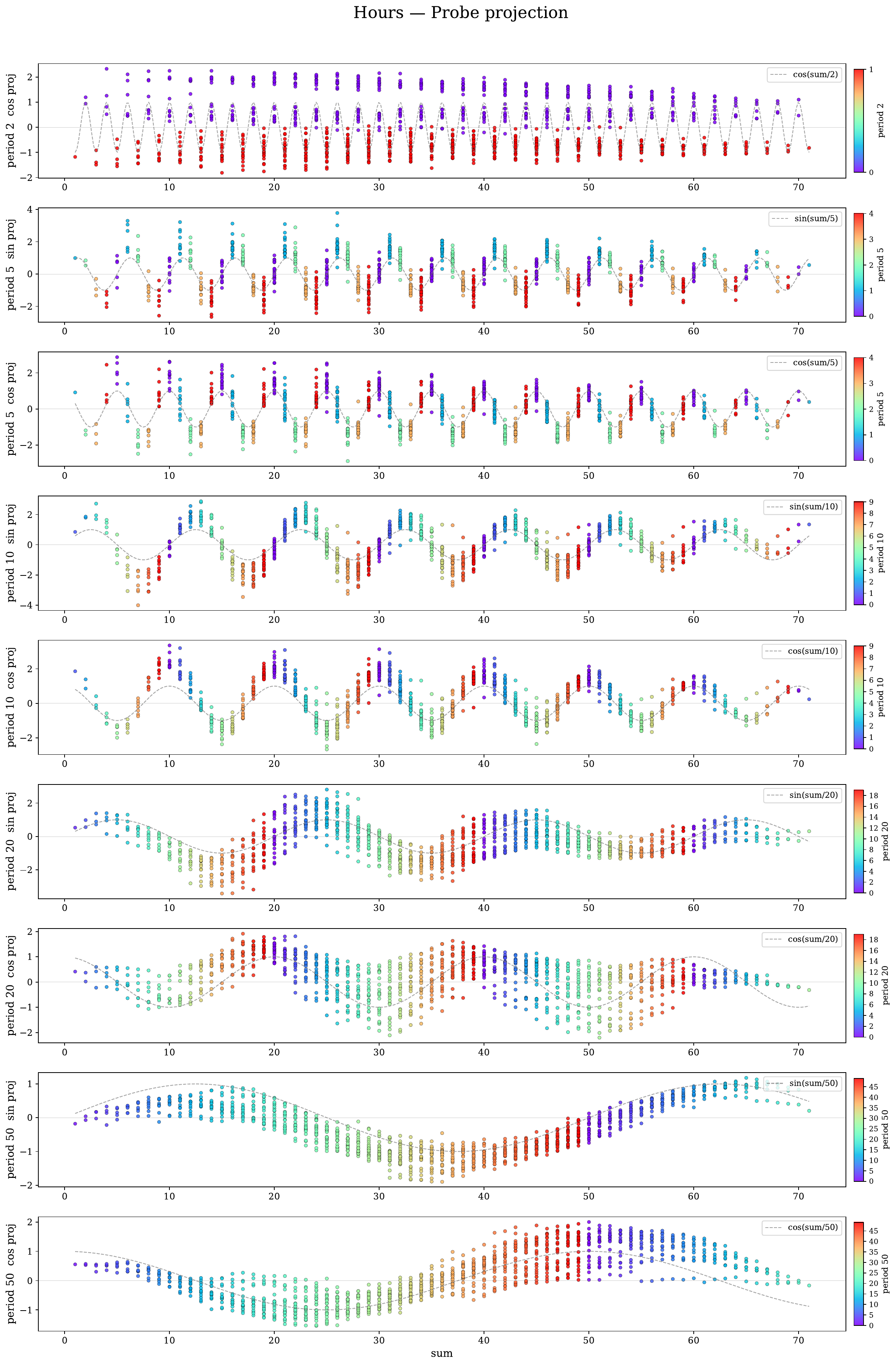}

    \caption{Projection of the layer 18 residual activations at the final token position for the hours task onto the Fourier probe directions learned from the addition task.}
    \label{fig:hours_direction}
\end{figure}

\begin{figure}[h]
    \centering
    \includegraphics[width=0.9\linewidth]{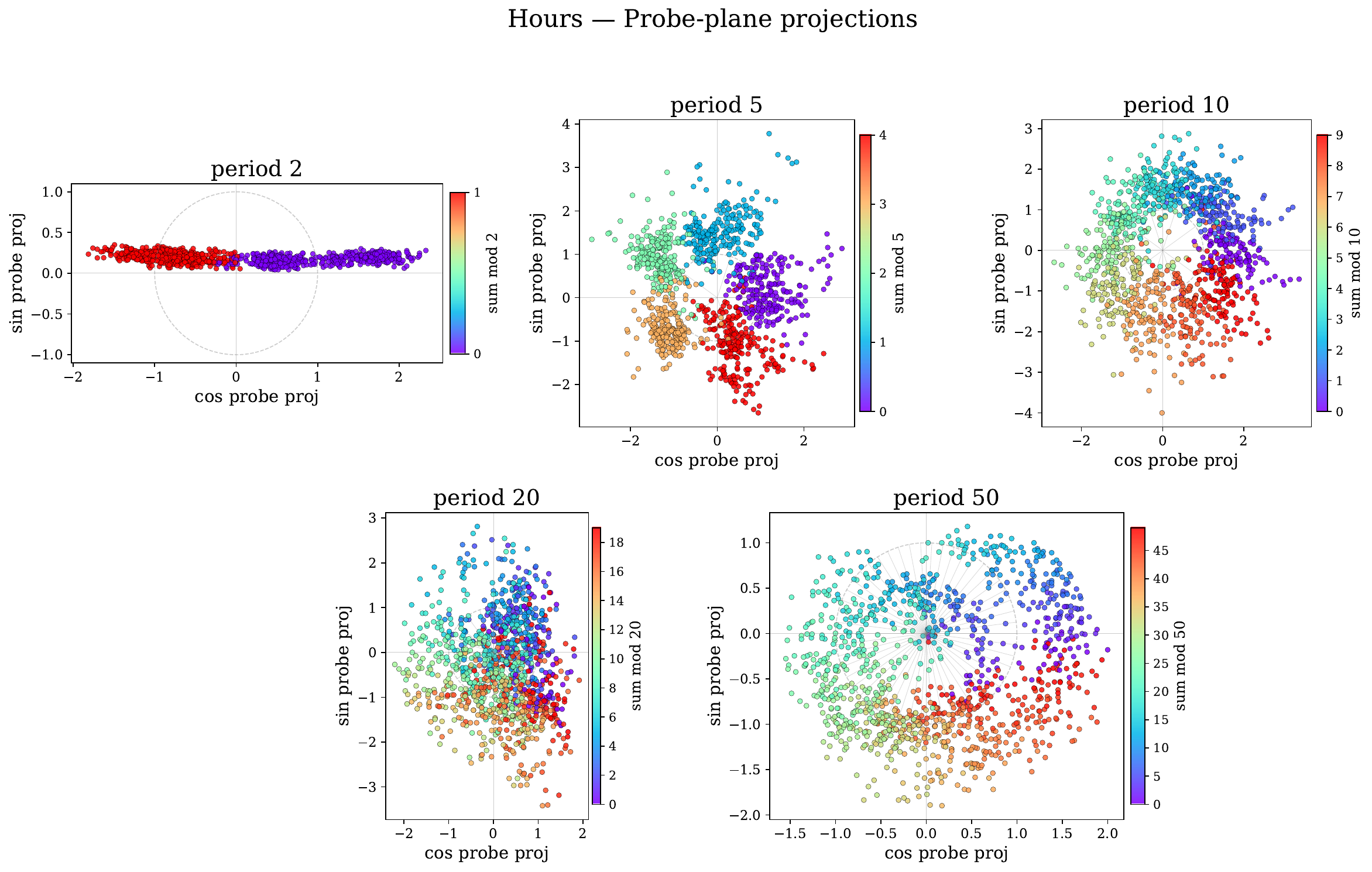}

        \caption{Projection of the layer 18 residual activations at the final token position for the hours task onto the Fourier planes learned from the addition task.}
    \label{fig:hours_plane}
\end{figure}

\begin{figure}[h]
    \centering
    \includegraphics[width=0.9\linewidth]{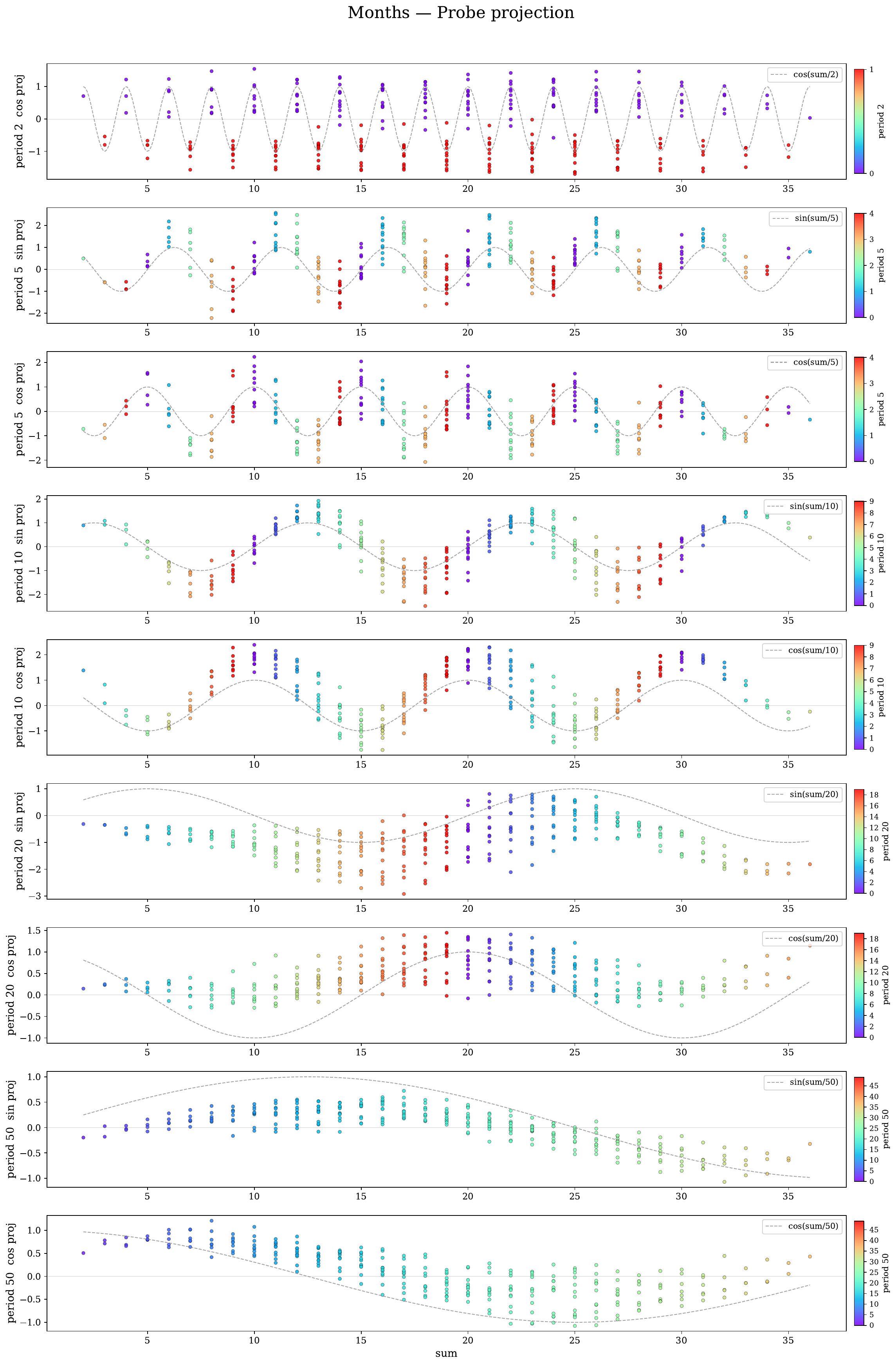}

    \caption{Projection of the layer 18 residual activations at the final token position for the months task onto the Fourier probe directions learned from the addition task.}
    \label{fig:months_direction}
\end{figure}

\begin{figure}[h]
    \centering
    \includegraphics[width=0.9\linewidth]{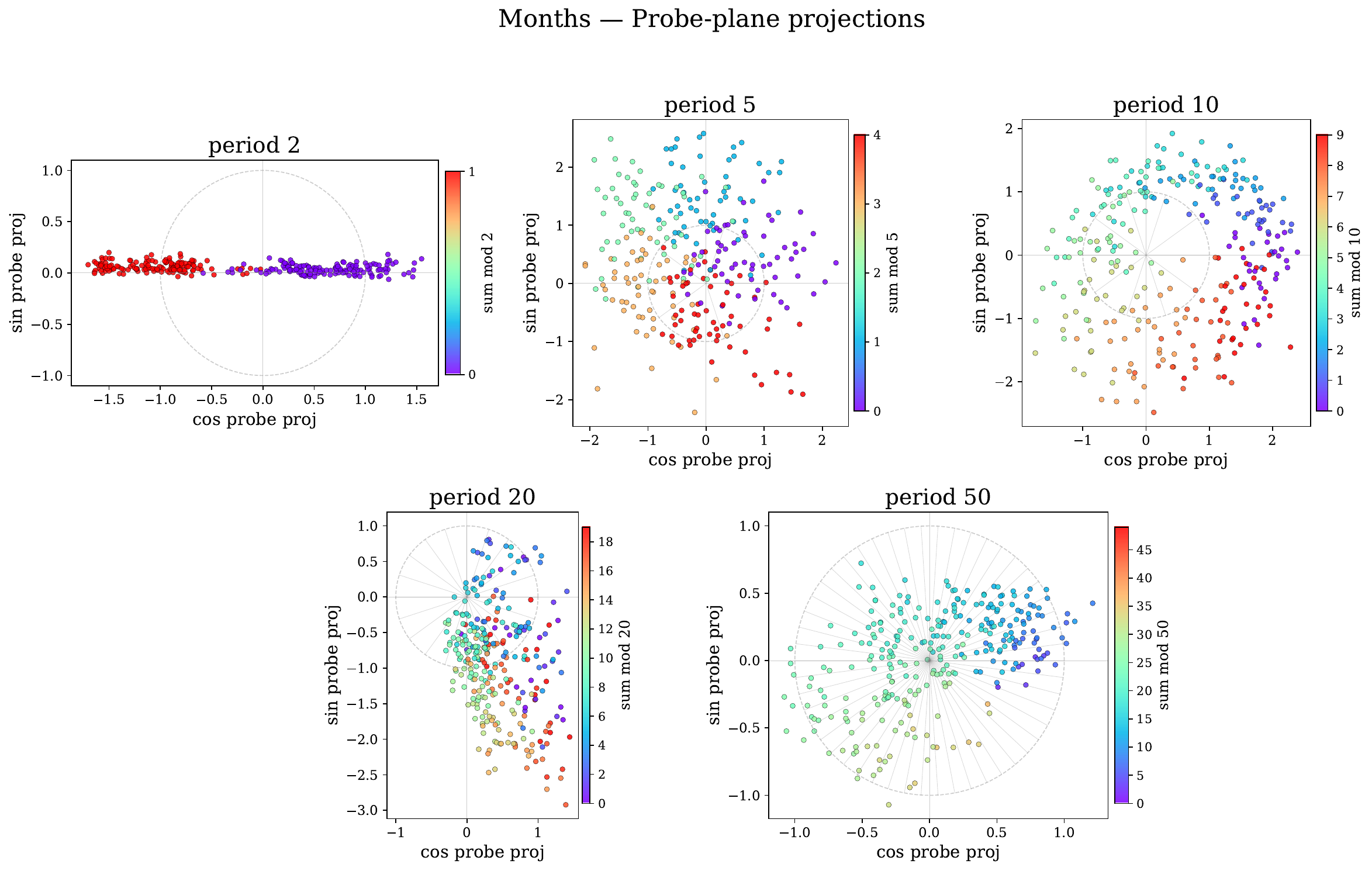}

    \caption{Projection of the layer 18 residual activations at the final token position for the months task onto the Fourier planes learned from the addition task.}
    \label{fig:months_plane}
\end{figure}

\begin{figure}[h]
    \centering
    \includegraphics[width=0.9\linewidth]{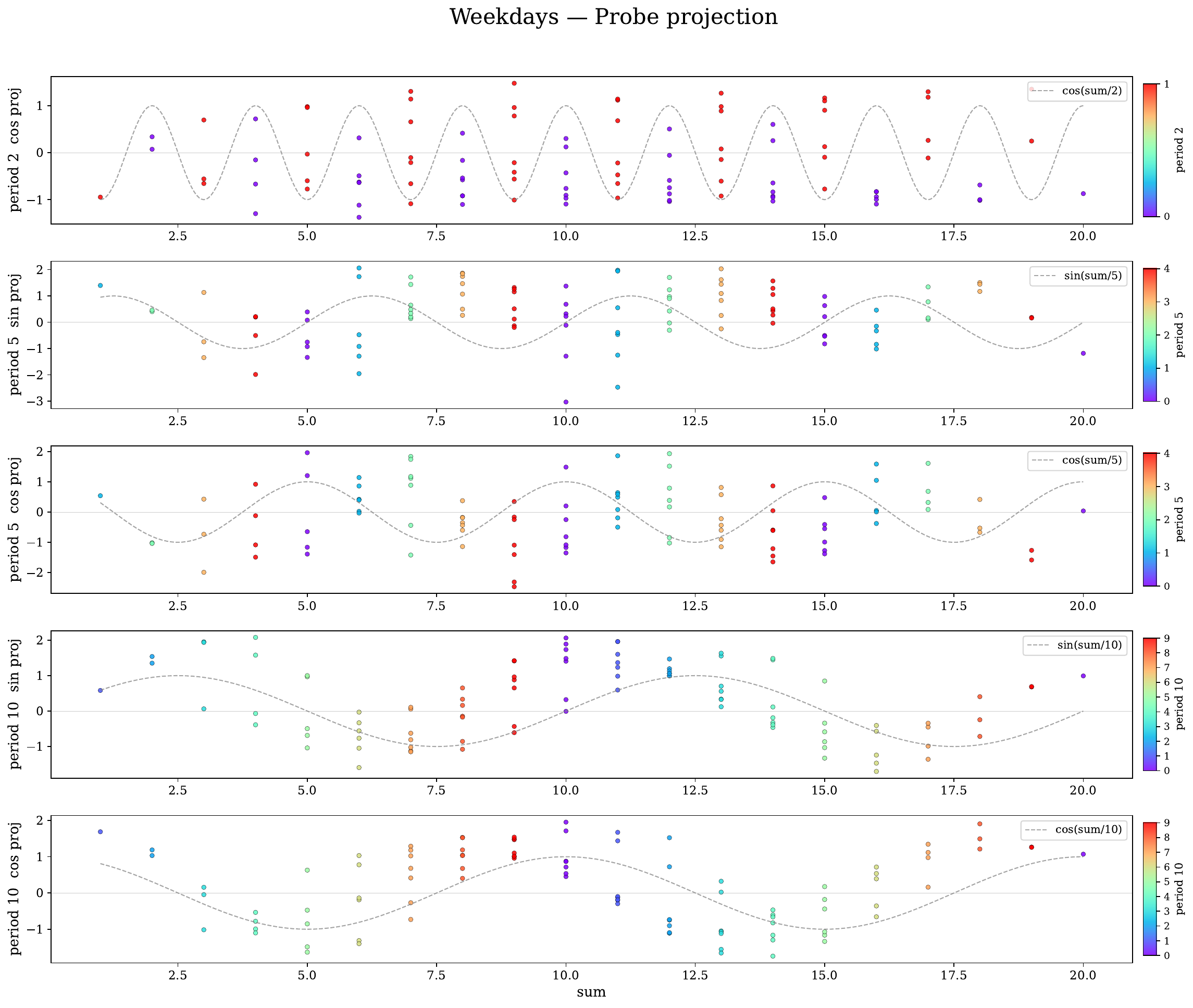}

    \caption{Projection of the layer 18 residual activations at the final token position for the weekdays task onto the Fourier probe directions learned from the addition task.}
    \label{fig:weekdays_direction}
\end{figure}

\begin{figure}[h]
    \centering
    \includegraphics[width=0.9\linewidth]{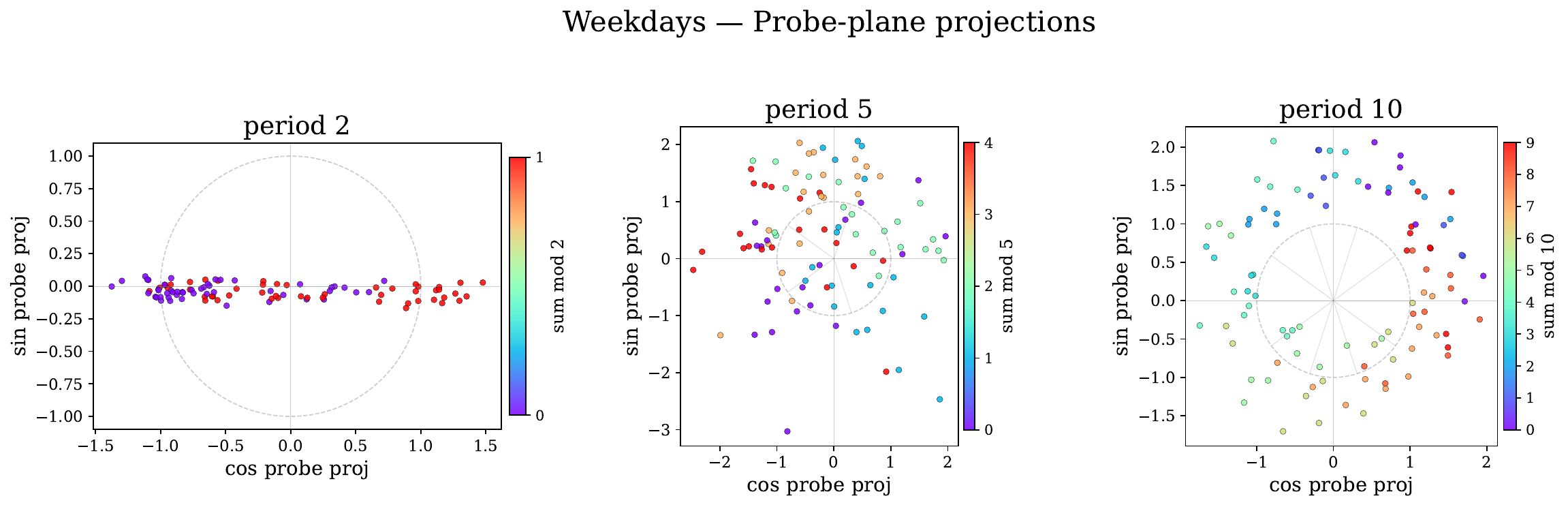}

    \caption{Projection of the layer 18 residual activations at the final token position for the weekdays task onto the Fourier planes learned from the addition task.}
    \label{fig:weekdays_plane}
\end{figure}

When applying the trained probes to the layer 18 residual activations at the final token position, the outputs follow a clear sinusoidal pattern, as expected (Figure~\ref{fig:summation_direction}). Taking the two probe outputs corresponding to the sine and cosine components for a given period T yields a circular pattern with that period (Figure~\ref{fig:summation_plane}). Although the probes are trained only on the addition task, we observe similar patterns when applying them to activations from the hours task (Figures~\ref{fig:hours_direction} and \ref{fig:hours_plane}), the months task (Figures~\ref{fig:months_direction} and \ref{fig:months_plane}), and the weekdays task (Figures~\ref{fig:weekdays_direction} and \ref{fig:weekdays_plane}).

\subsection{Fourier Probe Analysis}

\begin{figure}[t!]
    \centering
    \includegraphics[width=0.9\linewidth]{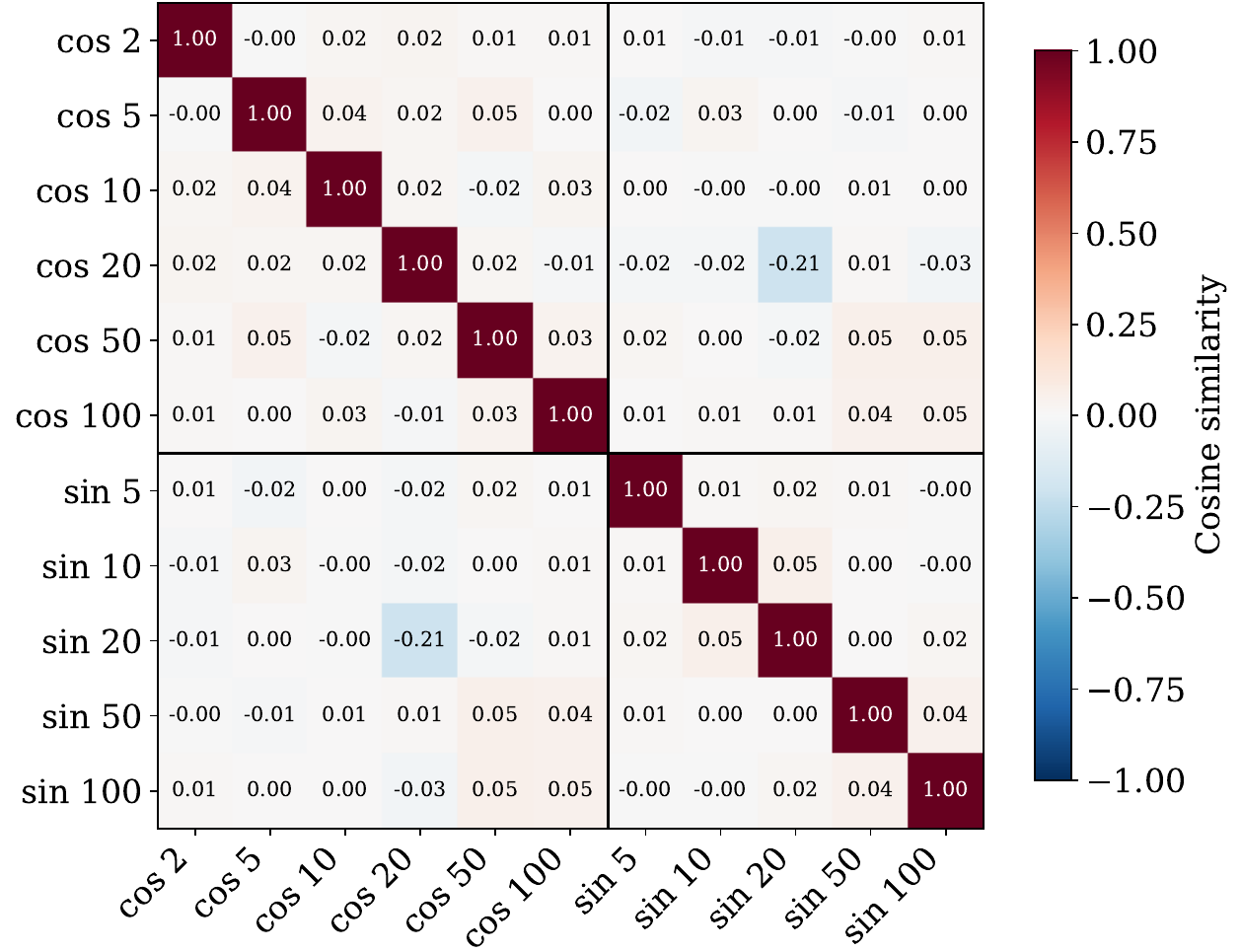}
    \caption{Cosine similarity scores for Fourier probes at layer 18. All probe directions are orthogonal, except for \(\cos(20)\) and \(\sin(20)\), which have cosine similarity \(-0.21\).}
    \label{fig:cosine}
\end{figure}

\paragraph{Fourier Probes are Orthogonal to Each Other}
Figure~\ref{fig:cosine} shows the cosine similarity between all probes for \(T \in \{2,5,10,20,50,100\}\) at layer 18. As can be seen, almost all probes are orthogonal to one another, with the exception of the \(\cos(20)\) and \(\sin(20)\) probes, whose cosine similarity is \(-0.21\).

\paragraph{Fourier Probes Overlap with DAS Subspaces}

\begin{figure}[t!]
    \centering
    \includegraphics[width=0.9\linewidth]{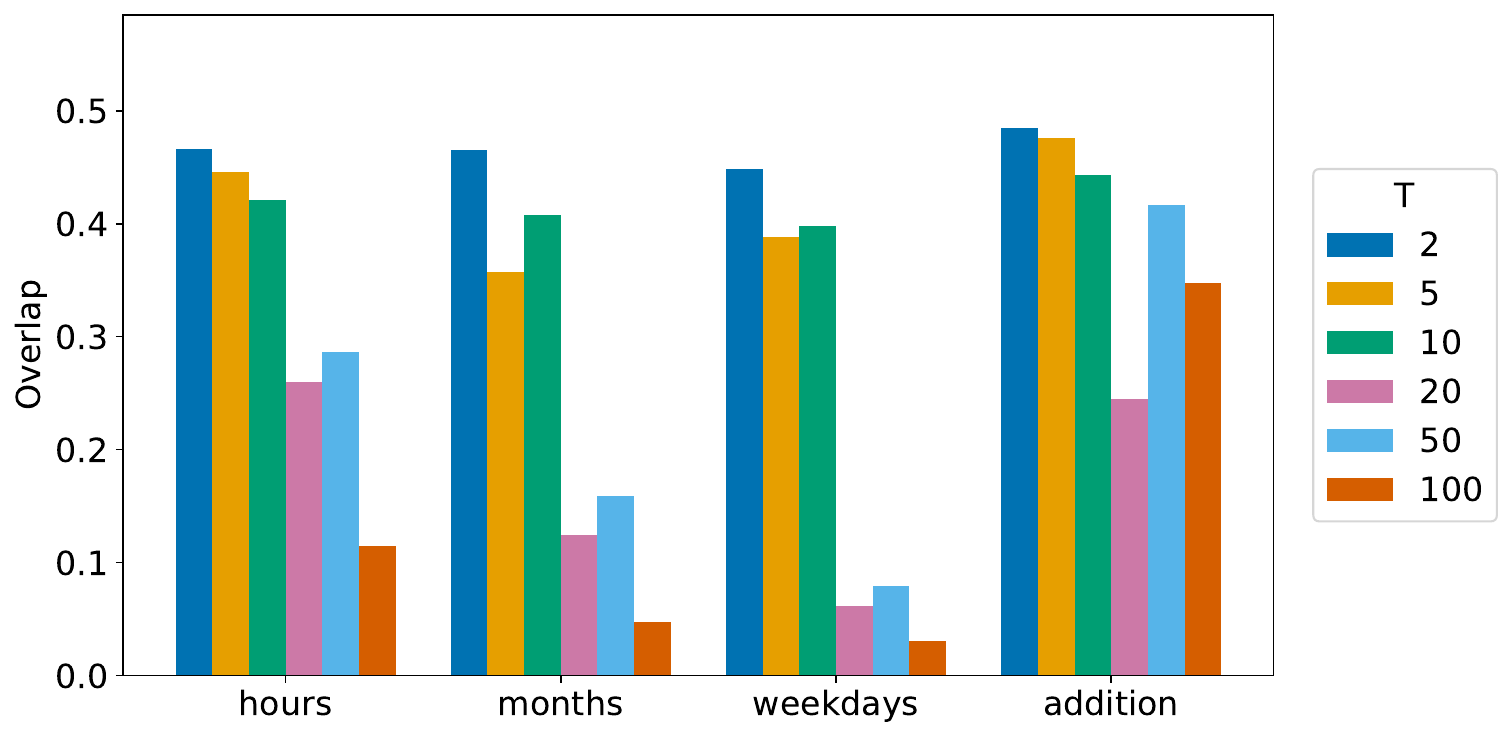}
    \caption{Overlap between addition Fourier probes and DAS output concept subspaces at layer 18. For each period \(T\), we report the average overlap of \(\mathbf{w}_{\cos}^{(T)}\) and \(\mathbf{w}_{\sin}^{(T)}\) with each DAS subspace, similar to Eq.~\ref{eq:write-score} (where the subspace is defined by the span of these two probes). For \texttt{months} and \texttt{weekdays}, only Fourier probes for $T\leq10$ overlap significantly with DAS output subspaces. This is likely because the maximum sums we train on for these tasks are 36 and 21 respectively. Even for \texttt{hours}, which has a larger output range (up to 72), overlap with $T=100$ is much lower than it is for \texttt{addition}, for which the largest sum is 200.}
  \label{fig:foruier_das_overlap}
\end{figure}

We measure the overlap between the Fourier probes and the \outputconcept\ DAS subspaces identified in Section~\ref{sec:das} for both the addition and cyclic tasks. We define the overlap score $\omega$ between a probe \(\mathbf{w}^{(T)}\) and a DAS subspace \(\mathbf{S}\) as:
\begin{equation}
\omega(\mathbf{w}^{(T)}, \mathbf{S}) = \frac{\|\mathbf{S}\mathbf{S}^\top \mathbf{w}^{(T)}\|}{\|\mathbf{w}^{(T)}\|}
\end{equation}
For each period \(T\), we report the average overlap across the sine and cosine probes. The results are shown in Figure~\ref{fig:foruier_das_overlap}. Notably, the Fourier probes exhibit substantial overlap with the DAS subspaces, despite being trained to find sinusoidal directions in the full residual stream and were trained only on the addition task. Interestingly, each task exhibits overlap with probes of different periods. While most tasks show overlap with probes for \(T \in \{2,5,10,20,50\}\), only the addition task also overlaps with probes at period \(T = 100\). This can be explained by the fact that the addition task is the only task with prompts where the output can exceed 100.

\subsection{Steering With Fourier Probes}
\label{app:steering}
To test whether the model uses the Fourier features, we evaluate whether the directions learned by the probes can be used to steer its predictions on the addition, month, hour, and weekday tasks. More concretely, we use the probes to steer the model's prediction from a value \(n\) to a counterfactual value \(n'\) (pre-modulo). For example, if we steer the model on the months task to \(n' = 6\), we expect it to predict ``June'', regardless of the prompt that we are steering on.

For the full steering algorithm, see Algorithm~\ref{alg:fourier-steering}. For the cyclic tasks, we use the same dataset described in Table~\ref{tab:task-templates}. For the addition task, we use a subset in which \(a,b \in [0,10]\). Table~\ref{tab:steering_periods} provides details on the steering targets and selected periods for each task. 

For each task, we steer every prompt in the dataset to each target value in turn. The results are shown in Figure~\ref{fig:steering}. Each row corresponds to the average output probabilities after steering all prompts toward a given target. A strong diagonal pattern indicates that the target token receives high probability after steering, while other tokens receive low probability. 

When projecting the layer 18 residual activations at the final token position onto the probe directions, the outputs follow a clear sinusoidal pattern, as expected (Figure~\ref{fig:summation_direction}). Furthermore, projecting onto the pair of directions corresponding to the sine and cosine components for a given period T yields a circular pattern with that period (Figure~\ref{fig:summation_plane}). Although the probes are trained only on the addition task, we observe similar patterns when projecting activations from the hours task (Figures~\ref{fig:hours_direction} and \ref{fig:hours_plane}), the months task (Figures~\ref{fig:months_direction} and \ref{fig:months_plane}), and the weekdays task (Figures~\ref{fig:weekdays_direction} and \ref{fig:weekdays_plane}).

\begin{figure}[t!]
    \centering
    \includegraphics[width=0.9\linewidth]{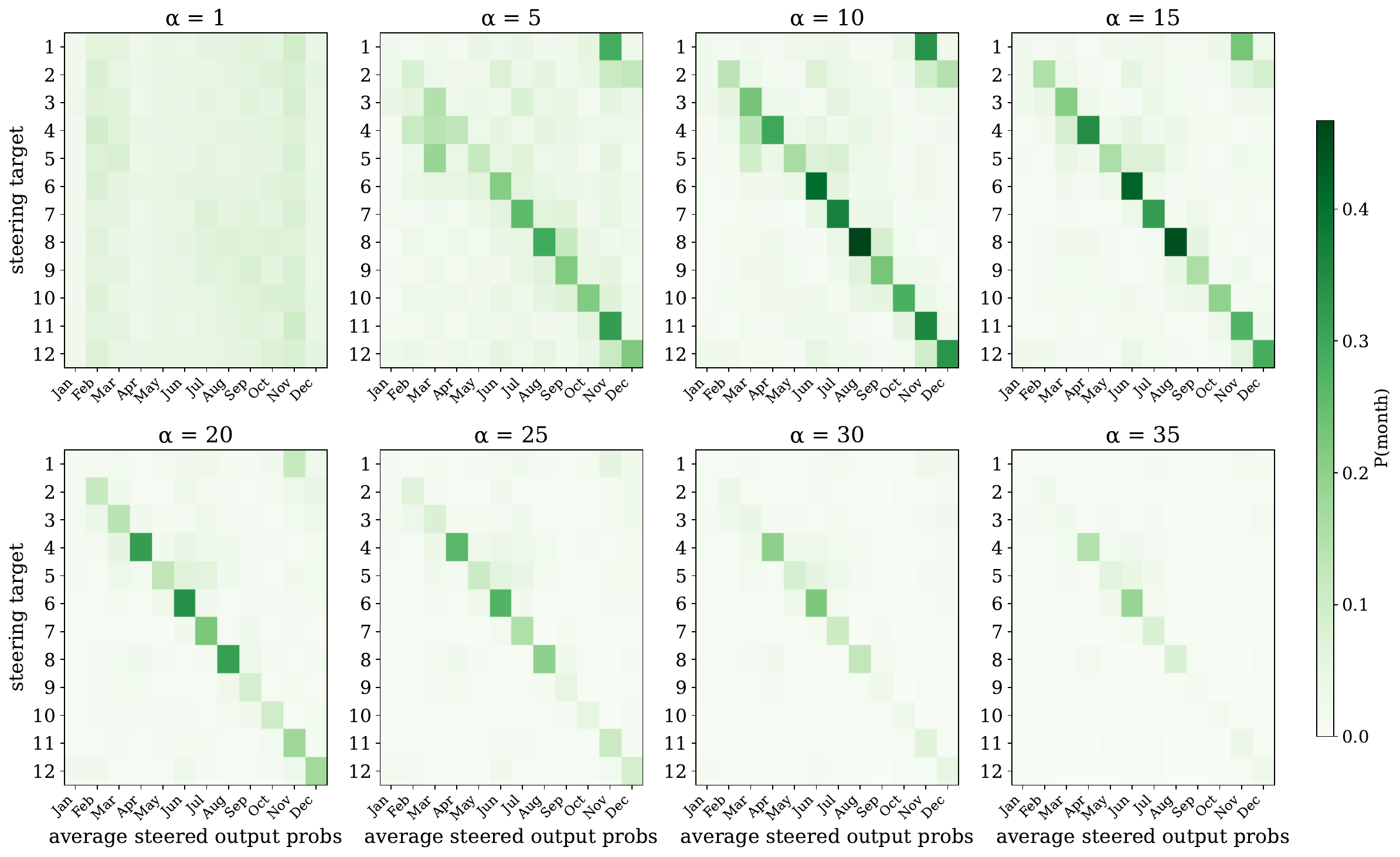}
    \caption{ Average output probabilities after steering with Fourier probes at the output of layer 18 for different steering factors \(\alpha\). For each original prompt, we steer toward a numeric target \(n'\) by modifying its Fourier features to encode that value, and average the resulting output distributions across targets. A strong diagonal indicates a successful intervention.}
  \label{fig:foruier_steering_per_alpha}
\end{figure}

\begin{figure}[t!]
    \centering
    \includegraphics[width=0.9\linewidth]{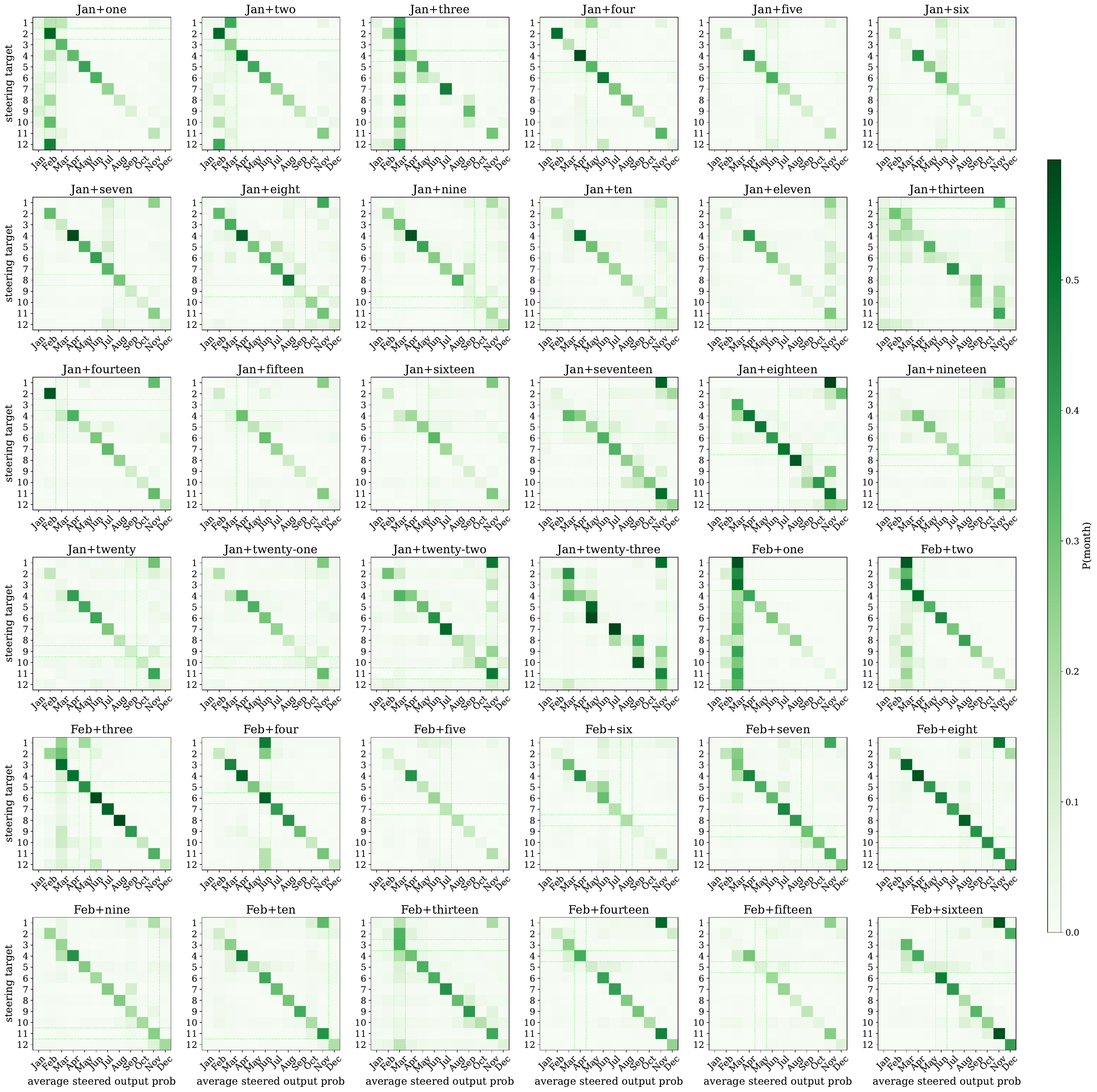}
    \caption{Output probabilities after steering each prompt with Fourier probes at the output of layer 18, using a steering factor of $\alpha = 10$. Each heatmap corresponds to a specific prompt and shows the output probabilities after applying the steering intervention to that prompt, with each row showing the probabilities after steering toward a particular target. A strong diagonal indicates a successful intervention. As shown, the effectiveness of steering varies across prompts. For example, steering the prompt “Three months after January” (top row, third column) largely fails, whereas steering “Seven months after January” (second row, left) performs well across most targets.}
  \label{fig:foruier_steering_per_prompt}
\end{figure}

\paragraph{\textbf{Steering Performance Across Individual Prompts.}}Figure~\ref{fig:foruier_steering_per_prompt} presents the steering results when applying the intervention to individual prompts in the months task. While steering performs well on average across targets, the per-prompt results are noticeably noisier. For example, steering the prompt “Three months after January” (top row, third column) largely fails, whereas steering “Seven months after January” (second row, left) performs well across most targets. Together with the need for a scaling factor, this suggests that the Fourier features at layer 18 alone do not fully override downstream computation.

\begin{table}[t]
\centering
\caption{Steering targets and selected Fourier periods for each task. Periods are chosen based on overlap with DAS output subspaces (Section~\ref{sec:das}, Figure~\ref{fig:foruier_das_overlap}).}
\begin{tabular}{l l l}
\toprule
\textbf{Task} & \textbf{Steering Targets} & \textbf{Steered Periods \(T\)} \\
\midrule
Weekdays & \(\{0,\dots,6\}\)   & \(\{2,5,10\}\) \\
Months   & \(\{1,\dots,12\}\)  & \(\{2,5,10,20,50\}\) \\
Hours    & \(\{0,\dots,23\}\)  & \(\{2,5,10,20,50\}\) \\
Addition & \(\{0,\dots,23\}\)  & \(\{2,5,10,20,50,100\}\) \\
\bottomrule
\end{tabular}
\label{tab:steering_periods}
\end{table}

\subsection{Circular Probes}
\label{app:circular_probes}

\citet{engelscircles} suggested training circular probes for cyclic tasks by first reducing activations with PCA ($d_{\text{PCA}} = 5$) and then fitting the probes via least squares, an approach that is particularly effective for small datasets.

The target in this approach is the same as in \ref{app:fourier_training}:
\begin{equation}
    \mathsf{MSE}\Bigg( \langle\mathbf{w}_{\sin}^{(l,T)}, \mathbf{h}^{(l)}_{a + b}\rangle, \sin\!\left(\tfrac{2\pi(a+b)}{T}\right)\Bigg),  \quad
    \mathsf{MSE}\Bigg(\langle\mathbf{w}_{\cos}^{(l,T)}, \mathbf{h}^{(l)}_{a+b}\rangle, \cos\!\left(\tfrac{2\pi(a+b)}{T}\right)\Bigg).
\end{equation}

Here, \(\mathbf{h}_{a+b}^{(l)} \in \mathbb{R}^{d_{\text{PCA}}}\) denotes the PCA-reduced hidden state at layer \(l\) corresponding to the final token, and the probe parameters are given by \(\mathbf{w}_{\sin}^{(l,T)}, \mathbf{w}_{\cos}^{(l,T)} \in \mathbb{R}^{1 \times d_{\text{PCA}}}\).

\begin{algorithm}[t]
\caption{Activation Steering via Fourier Probes}
\label{alg:fourier-steering}
\begin{algorithmic}[1]
 
\REQUIRE Base residual stream activation $\mathbf{h} \in \mathbb{R}^d$; steering target $k$; set of Periods $\mathcal{T}$; steering factor $\alpha$; learned probe parameters $\{(\mathbf{w}_{\sin}^{(t)}, b_{\sin}^{(t)}, \mathbf{w}_{\cos}^{(t)}, b_{\cos}^{(t)})\}_{t \in \mathcal{T}}$

\medskip
\STATE $\tilde{\mathbf{h}} \gets \mathbf{h}$ \COMMENT{Copy of base activation}
 
\medskip
\FOR{each period $t \in T$}
    \STATE $\triangleright$ \textit{Compute original Fourier radius}
    \STATE $\hat{s}_t \gets \mathbf{w}_{\sin}^{(t)} \cdot \mathbf{h} + b_{\sin}^{(t)}$; \quad $\hat{c}_t \gets \mathbf{w}_{\cos}^{(t)} \cdot \mathbf{h} + b_{\cos}^{(t)}$
    \STATE $r_t \gets \sqrt{\hat{s}_t^2 + \hat{c}_t^2}$
 
    \medskip
    \STATE $\triangleright$ \textit{Compute target Fourier coefficients}
    \STATE $\theta_t^{*} \gets \frac{2\pi\,k}{t}$ \COMMENT{Target angle on Fourier circle}
    \STATE $s_t^{*} \gets \alpha \, r_t \sin(\theta_t^{*})$; \quad $c_t^{*} \gets \alpha \, r_t \cos(\theta_t^{*})$
 
    \medskip
    \STATE $\triangleright$ \textit{Patch sine direction}
    \STATE $\hat{s}_t \gets \mathbf{w}_{\sin}^{(t)} \cdot \tilde{\mathbf{h}} + b_{\sin}^{(t)}$
    \STATE $\tilde{\mathbf{h}} \gets \tilde{\mathbf{h}} + \frac{s_t^{*} - \hat{s}_t}{\|\mathbf{w}_{\sin}^{(t)}\|^2}\;\mathbf{w}_{\sin}^{(t)}$
 
    \medskip
    \STATE $\triangleright$ \textit{Patch cosine direction}
    \STATE $\hat{c}_t \gets \mathbf{w}_{\cos}^{(t)} \cdot \tilde{\mathbf{h}} + b_{\cos}^{(t)}$
    \STATE $\tilde{\mathbf{h}} \gets \tilde{\mathbf{h}} + \frac{c_t^{*} - \hat{c}_t}{\|\mathbf{w}_{\cos}^{(t)}\|^2}\;\mathbf{w}_{\cos}^{(t)}$
\ENDFOR
 
\medskip
\STATE Replace $\mathbf{h}$ with $\tilde{\mathbf{h}}$ at the hook layer's last-token position
\RETURN $\mathrm{softmax}\bigl(\text{Model}(\text{input};\; \mathbf{h} \to \tilde{\mathbf{h}})\bigr)$
\end{algorithmic}
\end{algorithm}

\clearpage
\section{The Shared MLP Addition Module}
\subsection{Identifying Addition Neurons}\label{app:neuron-selection}

\begin{figure}[h!]
    \centering
    \includegraphics[width=\linewidth]{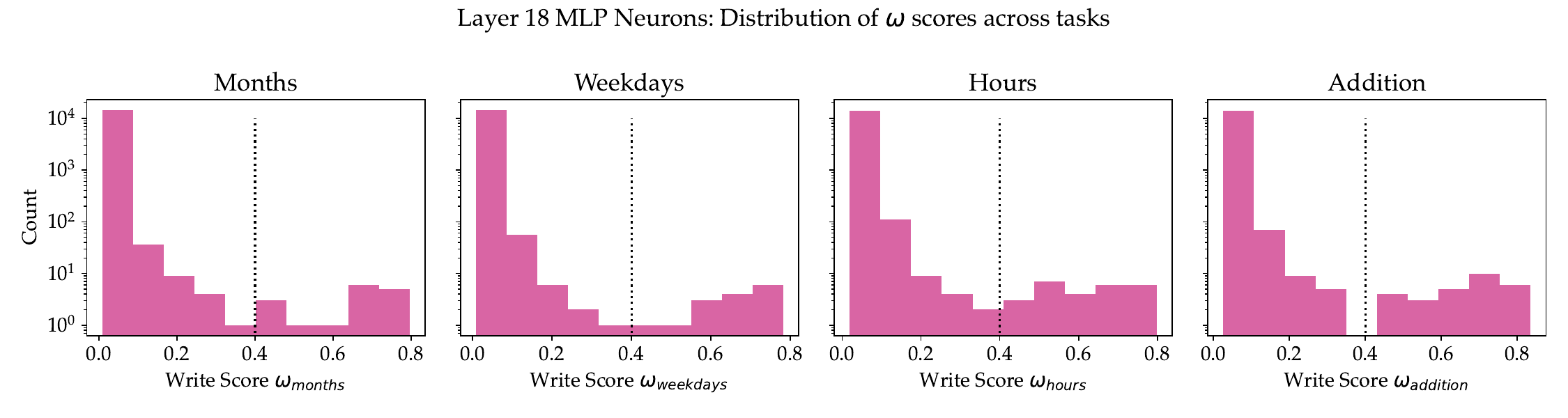}
    \caption{Distribution of write scores $\omega$ for layer 18 MLP neurons. A higher score means that this neuron's down projection row $\mathbf{d}_i$ is within the best output DAS subspace for this task. We pick the threshold $\tau=0.4$ by eye based on the \texttt{addition} task. There are 28 neurons with $\omega>0.4$ for the \texttt{addition} task, 16 for \texttt{months}, 15 for \texttt{weekdays}, and 26 for \texttt{hours}. All neurons for cyclic tasks are contained within the set of \texttt{addition} neurons, except for a single \texttt{hours} neuron.}
    \label{fig:neuron-hist}
\end{figure}

Figure~\ref{fig:neuron-hist} shows the distribution of write scores for all layer 18 MLP neurons (Section~\ref{sec:identify-neurons}): this score measures the proportion of a neuron's down projection row $\mathbf{d}_i$ that is within the best DAS output subspace at layer 18. We choose a threshold $\tau=0.4$ by eye based on the \texttt{addition} task, and find that neurons in all other tasks are a subset of these 28 \texttt{addition} neurons. Table~\ref{tab:l18-neuron-ablation} gives change in performance when this set of 28 neurons is ablated, as well as when all other neurons at layer 18 are ablated. These addition neurons can explain most of the important computation at the layer 18 MLP.

\begin{figure}[h!]
    \centering
    \includegraphics[width=0.9\linewidth]{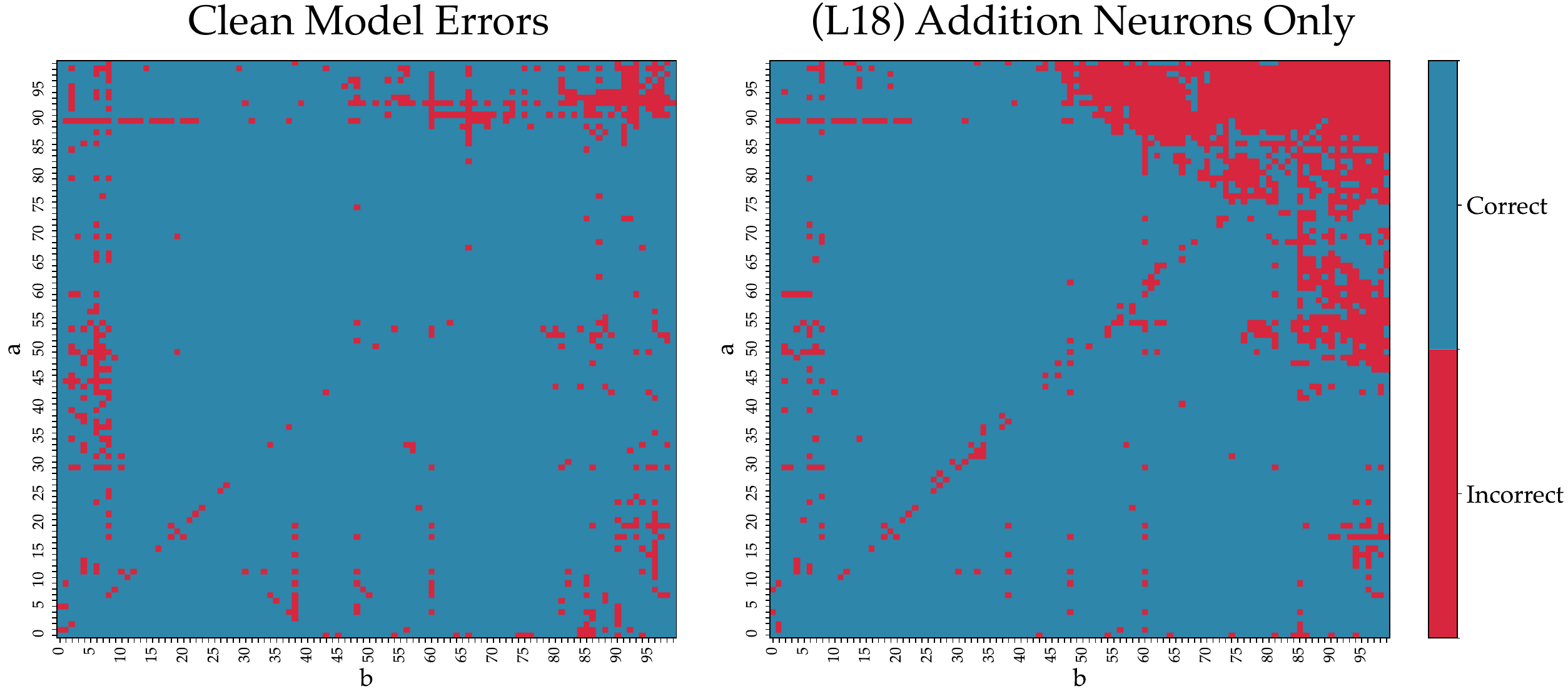}
    \caption{\llama\ errors on the \texttt{addition} task. We show errors for a clean model run (95\% accuracy), as well as errors when all neurons at the L18 MLP are zero-ablated except for our 28 addition neurons $\mathcal{N}_{\text{add}}$ (86\% accuracy). Most errors come from higher number ranges, suggesting that $\mathcal{N}_{\text{add}}$ excludes some neurons with larger periods.}
    \label{fig:only-addition-errors}
\end{figure}

\paragraph{Neuron ablations.} We show ablation results for our canonical tasks in Table~\ref{tab:l18-neuron-ablation}. We also test ablation on unseen templates in Table~\ref{tab:unseen-templates}, finding that these neurons are also important for rephrasings of the same task. 
Figure~\ref{fig:only-addition-errors} shows errors for \llama\ on the \texttt{addition} task, compared to errors when all MLP neurons at layer 18 are zero-ablated except for $\mathcal{N}_{\text{add}}$; failure for larger numbers suggests that we may be missing larger period neurons.

\begin{table}[h!]
\centering
\caption{Accuracy (\%) intervening on the set of 28 \texttt{addition} neurons $\mathcal{N}_{\text{add}}$. \textbf{Clean}: \llama\ accuracy on this task. \textbf{Only $\mathcal{N}_{\text{add}}$}: accuracy when all other neuron activations at the layer 18 MLP are set to 0, except for neurons in $\mathcal{N}_{\text{add}}$. Despite these neurons making up 0.2\% of layer 18 neurons, accuracy remains high. \textbf{Zeroed}: accuracy drops significantly when these neurons' activations are set to zero. \textbf{Flipped}: reversing the sign of these neurons' activations is destructive, likely because they represent periodic functions that oscillate between negative and positive values across inputs.}
\begin{tabular}{llccccc}
\toprule
\textbf{Task} & \textbf{Number Range} & \textbf{Clean} & \textbf{Only $\mathcal{N}_{\text{add}}$} & \textbf{Zeroed} & \textbf{Flipped} & \textit{Chance} \\
\midrule
\multirow{2}{*}{Weekdays} & 1$\rightarrow$7   & 91.84 & 89.80 & 40.82 & 20.41 & \textit{14.29} \\
                          & 8$\rightarrow$14  & 68.37 & 58.16 & 28.57 & 17.35 & \textit{14.29} \\
\midrule
\multirow{2}{*}{Months}   & 1$\rightarrow$12  & 81.25 & 77.78 & 36.81 & 14.58 &  \textit{8.33} \\
                          & 13$\rightarrow$24 & 64.93 & 70.83 & 23.61 & 16.32 &  \textit{8.33} \\
\midrule
\multirow{2}{*}{Hours}    & 1$\rightarrow$24  & 93.06 & 80.56 & 50.69 & 16.32 &  \textit{4.17} \\
                          & 24$\rightarrow$48 & 69.18 & 51.65 & 29.77 & 11.63 &  \textit{4.17} \\
\midrule
Addition                  & $a,b\in[1..100]$  & 94.60 & 86.36 & 24.11 &  2.97 & \textit{0.50} \\
\bottomrule
\end{tabular}
\label{tab:l18-neuron-ablation}
\end{table}

\begin{table}[h!]
\caption{Neuron intervention results across alternative prompt templates, for one cycle each. \textbf{Clean}: unmodified accuracy for \llama. \textbf{Only}: accuracy when all neurons at layer 18 except for $\mathcal{N}_{\text{add}}$ are zero-ablated. \textbf{Zeroed}: accuracy when all 28 $\mathcal{N}_{\text{add}}$ addition neurons are zero-ablated. \textbf{Flipped}: accuracy when 28 $\mathcal{N}_{\text{add}}$ neurons have their signs flipped.}
\begin{tabular}{@{}l>{\raggedright}p{5.5cm}>{\raggedright}p{1.3cm}>{\raggedright}p{1.3cm}>{\raggedright}p{1.3cm}l@{}}
\toprule
\bf Task & \bf Template & \bf Clean & \bf Only & \bf Zeroed & \bf Flipped \\ \midrule
\tt weekdays & \it (default prompt) & 91.84\% & 89.80\% & 40.82\% & 20.41\% \\
\tt weekdays & Question: Today is \{input\}. In \{number\} days, what day will it be? Answer: It will be & 95.24\% & 97.62\% & 50.00\% & 26.19\% \\
\tt weekdays & If today is \{input\}, and I have a deadline in \{number\} days, what day is my deadline? My deadline is on & 40.82\% & 40.82\% & 22.45\% & 12.24\% \\
\tt months & \it (default prompt) & 81.25\% & 77.78\% & 36.81\% & 14.58\% \\
\tt months & Question: The current month is \{input\}. In \{number\} months, it will be my birthday. What month is my birthday? Answer: My birthday is in & 50.76\% & 51.52\% & 34.09\% & 20.45\% \\
\tt months & Q: A pregnant woman will give birth in \{number\} months. If it is currently \{input\}, what month will she give birth? Answer: She will give birth in & 18.75\% & 27.08\% & 13.19\% & 6.25\% \\
\tt hours & \it (default prompt) & 93.06\% & 80.56\% & 50.69\% & 16.32\% \\
\tt hours & Question: It's \{input\}:00 right now. My flight is in \{number\} hours. What time is my flight? Answer: My flight is at & 39.49\% & 20.29\% & 23.01\% & 10.87\% \\
\tt hours & When I was in the military, I learned 24-hour time. So if right now it is \{input\}:00, in \{number\} hours it will be & 86.28\% & 78.82\% & 48.44\% & 16.67\% \\
\bottomrule
\end{tabular}
\label{tab:unseen-templates}
\end{table}

\subsection{Addition Neurons Group by Fourier Frequency}

\begin{figure}
    \centering
    \includegraphics[width=\linewidth]{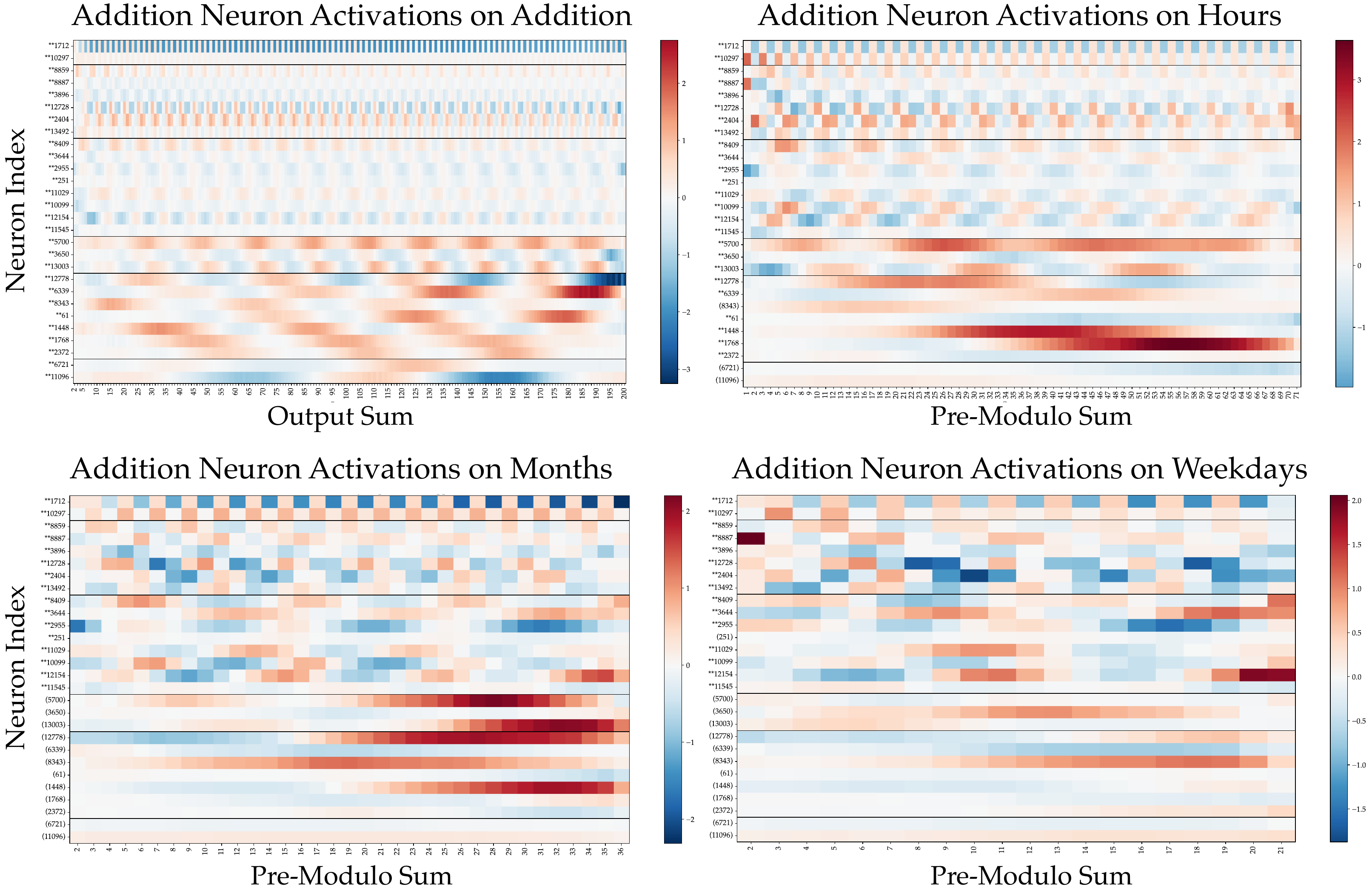}
    \caption{Addition neuron activations $\mathcal{N}_{\text{add}}$ across all examples for all tasks: \texttt{addition}, \texttt{hours}, \texttt{months}, and \texttt{weekdays}. We observe the same periodic structure across sums for all four tasks, although it is more difficult to see for tasks with smaller output ranges. Neurons that are also within the set for a respective task are **starred (e.g., $n_{1712}$ is starred in all plots, so is used for all tasks). Addition neurons that are irrelevant for this task are marked in parentheses. Notably, \texttt{addition} neurons that are irrelevant for \texttt{months} and \texttt{weekdays} all correspond to larger $m$. In fact, the largest $m=100$ neurons, $n_{6721}$ and $n_{11096}$, are \textit{only} relevant for the \texttt{addition} task.}
    \label{fig:unclipped-ribbons}
\end{figure}

Figure~\ref{fig:unclipped-ribbons} shows neuron activations averaged across output sums for all addition neurons. Averages are taken across output sums for the \texttt{addition} task and pre-modulo sums for \texttt{hours}, \texttt{months}, and \texttt{weekdays}. 

\begin{figure}
    \centering
    \includegraphics[width=0.9\linewidth]{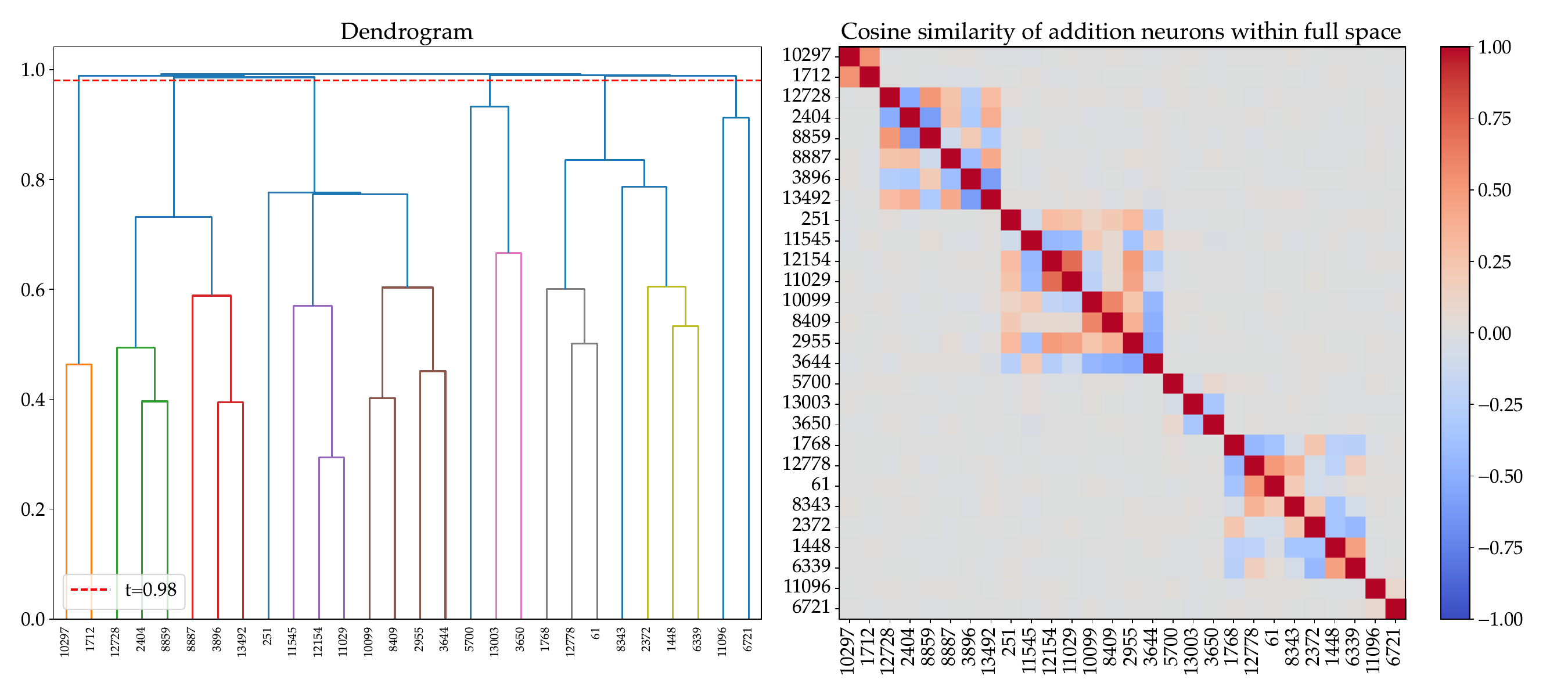}
    \caption{Simple hierarchical clustering of addition neurons with $1 - |\text{cosine\_sim}|$ as a distance metric recovers the same neuron clusters we observe in Figure~\ref{fig:unclipped-ribbons}.}
    \label{fig:cluster-neurons}
\end{figure}

\subsection{Neuron Activations Across Tasks}\label{app:neurons-all-tasks}

\begin{figure}[H]
    \centering
    \includegraphics[width=\linewidth]{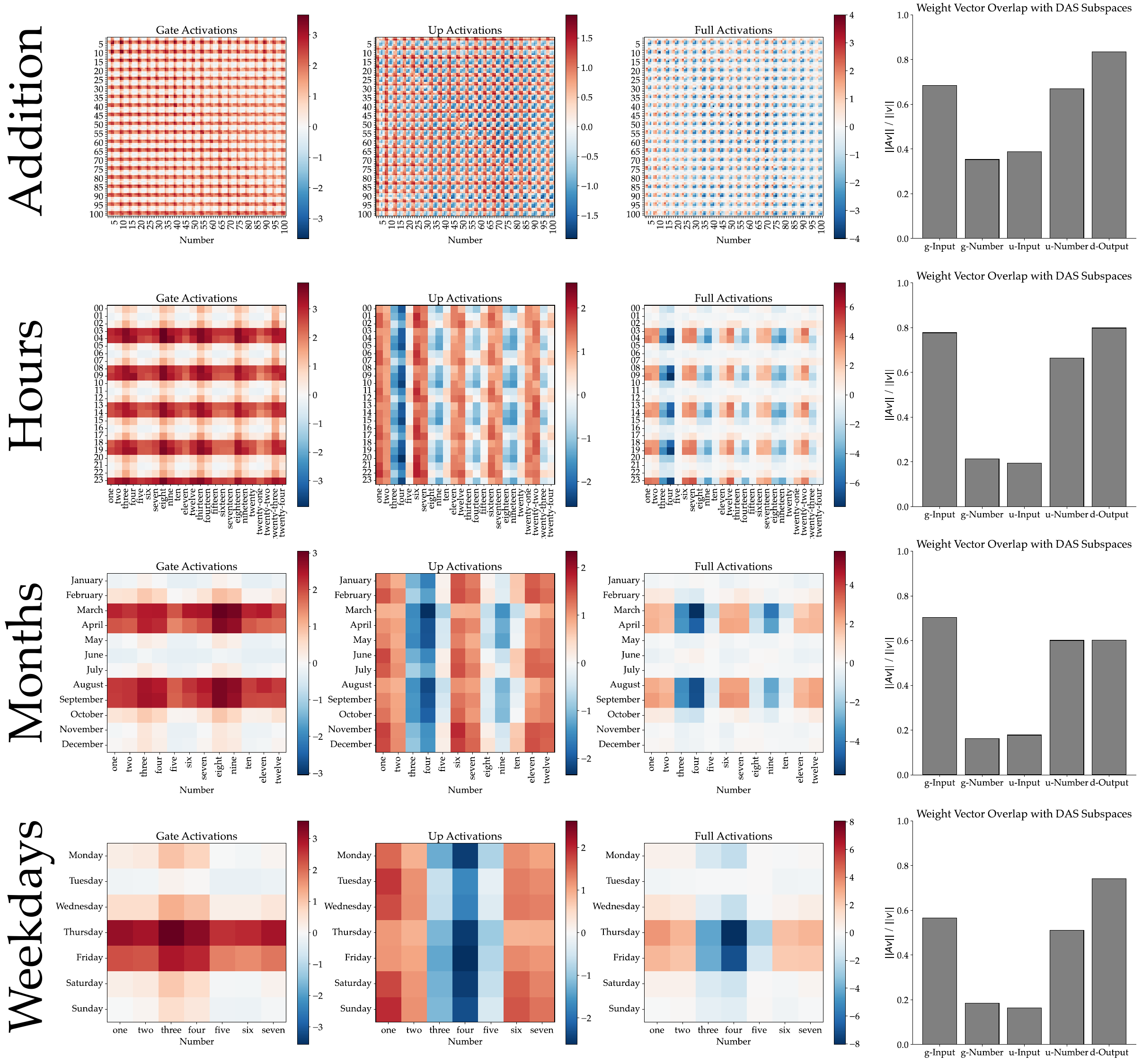}
    \caption{\textbf{N12728} activations across prompts for all four tasks, organized by input variables. This is a period 5 neuron. We also show read/write scores for $\mathbf{g}_i,\mathbf{u}_i,\mathbf{d}_i$ with input and output spaces. For all cyclic tasks, we can see that this neuron's gate vector $\mathbf{g}_{12728}$ has a much higher read scores from the input subspace (high first bar, horizontal stripes), whereas its up vector $\mathbf{u}_{12728}$ reads more heavily from the number subspace (high fourth bar, vertical stripes). Interestingly, activations are not as ``split'' for the \texttt{addition} task; this may be related to the fact that variables for \textit{a} and \textit{b} in \textit{a+b=} are not causally separable by DAS, as results from Figure~\ref{fig:coarse-das-all} show.}
    \label{fig:12728-alltasks}
\end{figure}

\begin{figure}[H]
    \centering
    \includegraphics[width=\linewidth]{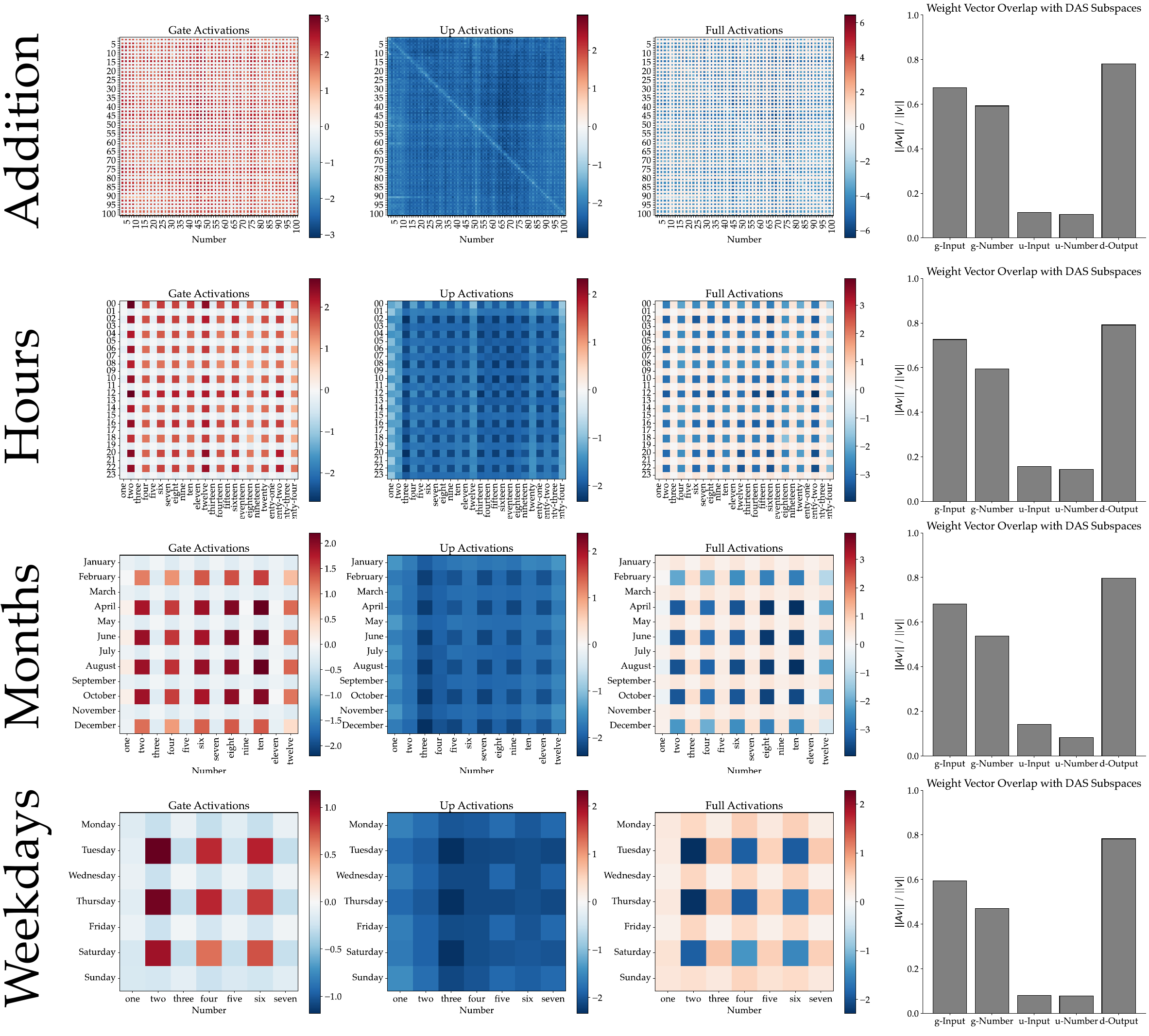}
    \caption{\textbf{N1712} activations across prompts for all four tasks, organized by input variables. This is a parity neuron. We also show read/write scores for $\mathbf{g}_i,\mathbf{u}_i,\mathbf{d}_i$ with input and output spaces. We can see that this neuron's gate vector $\mathbf{g}_{1712}$ reads equally from both input subspaces (checkered pattern across examples, and the first two bars are similar heights), whereas its up vector $\mathbf{u}_{1712}$ is mostly negative across examples, only slightly reading from the input subspaces.}
    \label{fig:1712-alltasks}
\end{figure}

\begin{figure}[H]
    \centering
    \includegraphics[width=\linewidth]{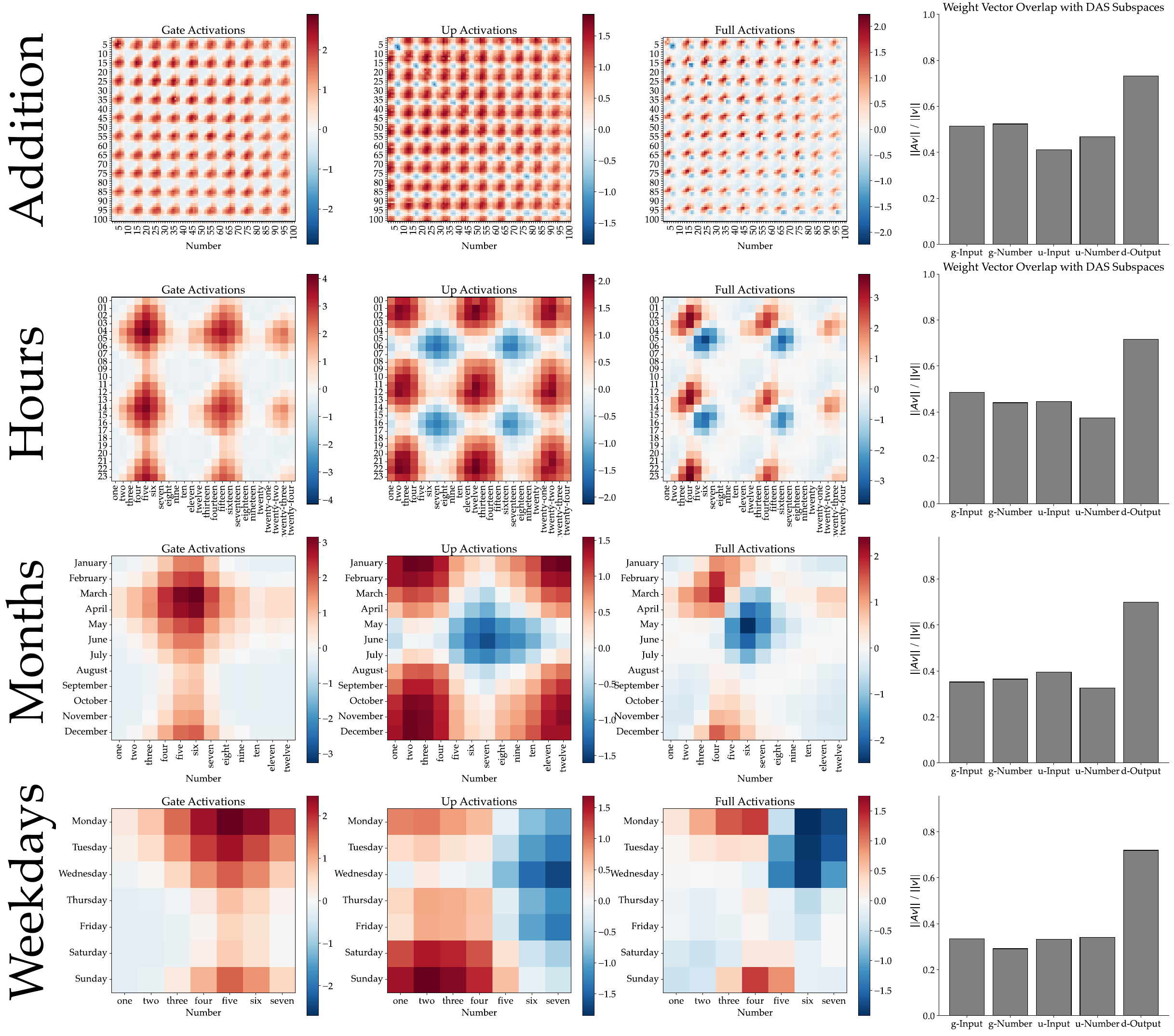}
    \caption{\textbf{N8409} activations across prompts for all four tasks, organized by input variables. This is a period 10 neuron. We also show read/write scores for $\mathbf{g}_i,\mathbf{u}_i,\mathbf{d}_i$ with input and output spaces. We can see that this neuron's gate vector $\mathbf{g}_{8409}$ reads equally from both input subspaces (checkered pattern across examples, and the first two bars are similar heights), as well as its up vector $\mathbf{u}_{8409}$.}
    \label{fig:8409-alltasks}
\end{figure}

We show activations across all tasks for three neurons: \textbf{N12728}, which is a period 5 neuron (Figure~\ref{fig:12728-alltasks}), \textbf{N1712}, the even parity neuron (Figure~\ref{fig:1712-alltasks}), and \textbf{N8409}, which is a period 10 neuron (Figure~\ref{fig:8409-alltasks}). The first neuron is a ``split'' neuron that dedicates its gate activations to the \inputconcept\ and its up activations to the \inputnumber, whereas the latter two neurons read equally from both inputs. Activations are consistent across tasks, with e.g. \texttt{weekdays} being a more ``zoomed in'' version of tasks with larger number ranges.

\clearpage
\subsection{All Addition Neurons at Layer 18 $\mathcal{N}_{\text{add}}$}\label{app:all-neurons}

\paragraph{Identifying split neurons.} For each addition neuron, we compute ``read scores'' (overlap with input subspaces) following Eq.~\ref{eq:write-score} with \inputconcept\ and \inputnumber\ subspaces for $\mathbf{g}_i,\mathbf{u}_i$. We say an activation pattern is \textit{split} for this task if, for both the gate and up projections, one of the \inputconcept\ or \inputnumber\ overlap scores is at least 50\% greater than the other score. That is, both of the below conditions should be true:
\begin{align}\label{eq:split-score}
    \max(\omega_{\inputconcept}(\mathbf{g}_i), \omega_{\inputnumber}(\mathbf{g}_i)) &> 1.5 \cdot \min(\omega_{\inputconcept}(\mathbf{g}_i), \omega_{\inputnumber}(\mathbf{g}_i)) \\
    \max(\omega_{\inputconcept}(\mathbf{u}_i), \omega_{\inputnumber}(\mathbf{u}_i)) &> 1.5 \cdot \min(\omega_{\inputconcept}(\mathbf{u}_i), \omega_{\inputnumber}(\mathbf{u}_i)).
\end{align}
We find that under this threshold, 17/28 neurons in $\mathcal{N}_{\text{add}}$ have \textit{split} activations for the \texttt{hours} task.

\begin{figure}
    \centering
    \includegraphics[width=\linewidth]{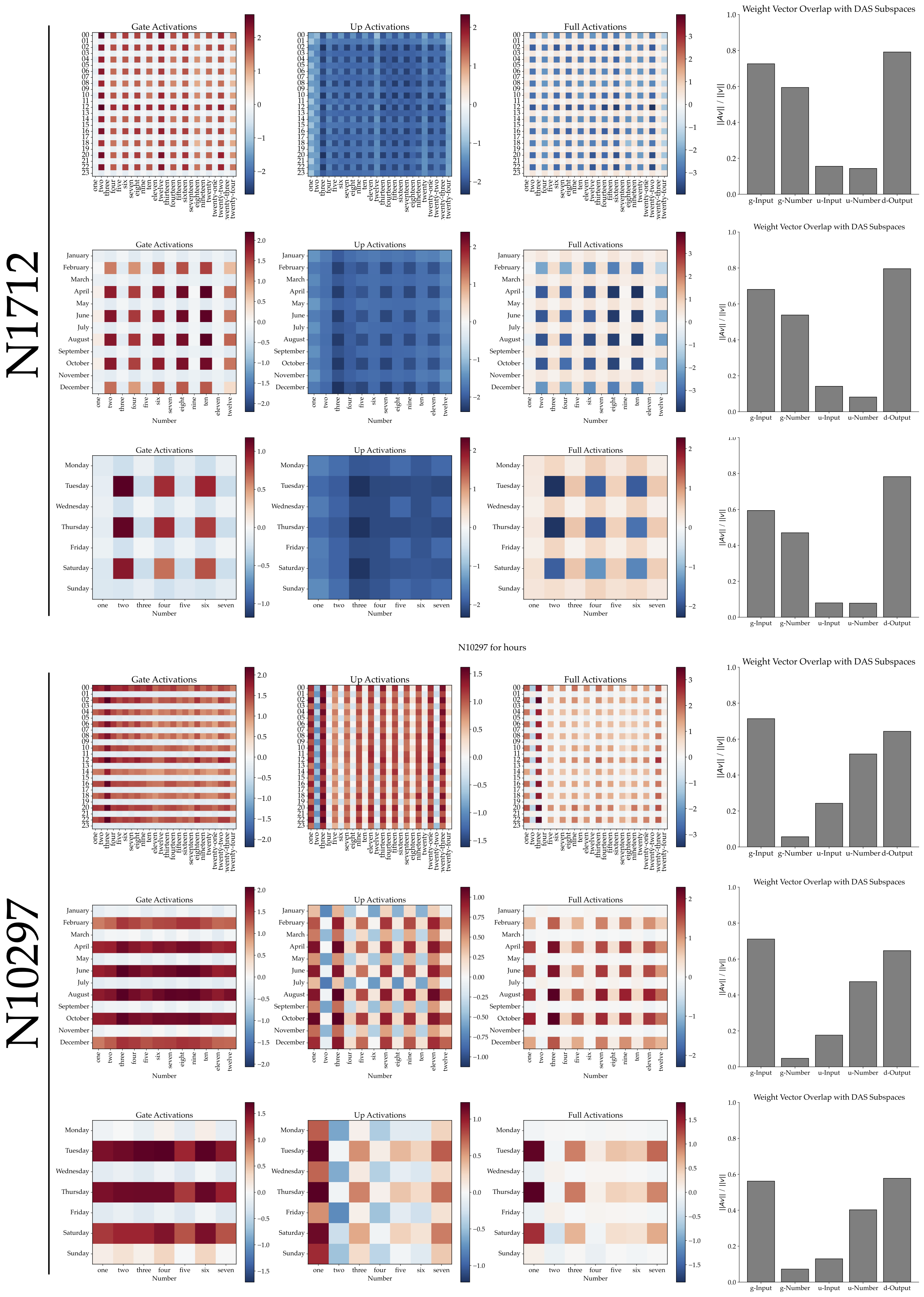}
    \caption{\textbf{All period 2 neurons}, activations for the \texttt{hours} and \texttt{months} tasks.}
    \label{fig:allmod2-hours}
\end{figure}

\begin{figure} 
    \centering
    \includegraphics[width=\linewidth]{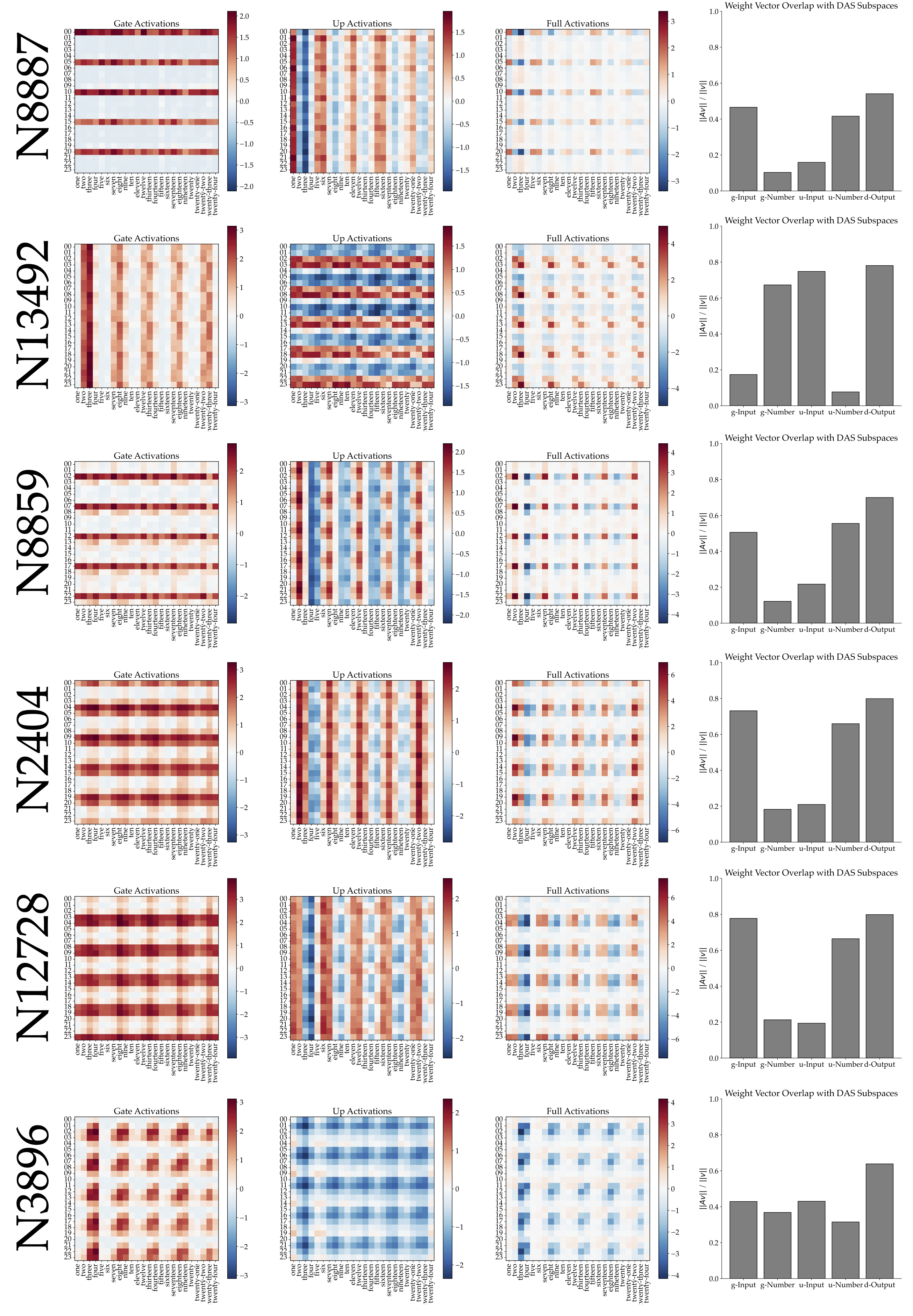}
    \caption{\textbf{All period 5 neurons}, activations for the \texttt{hours} task.}
    \label{fig:allmod5-hours}
\end{figure}

\begin{figure}
    \centering
    \includegraphics[width=0.8\linewidth]{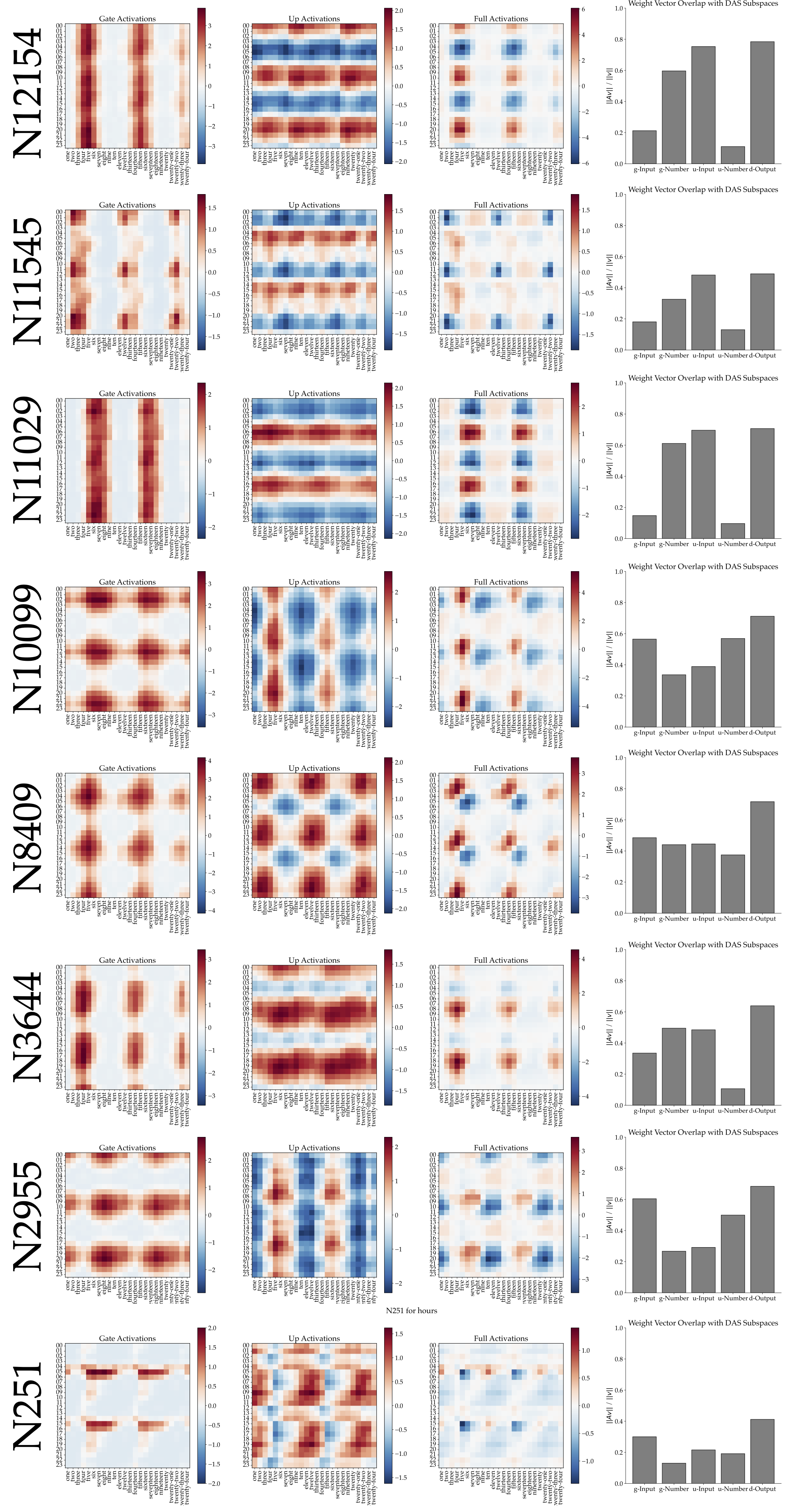}
    \caption{\textbf{All period 10 neurons}, activations for the \texttt{hours} task.}
    \label{fig:allmod10-hours}
\end{figure}

\begin{figure}
    \centering
    \includegraphics[width=\linewidth]{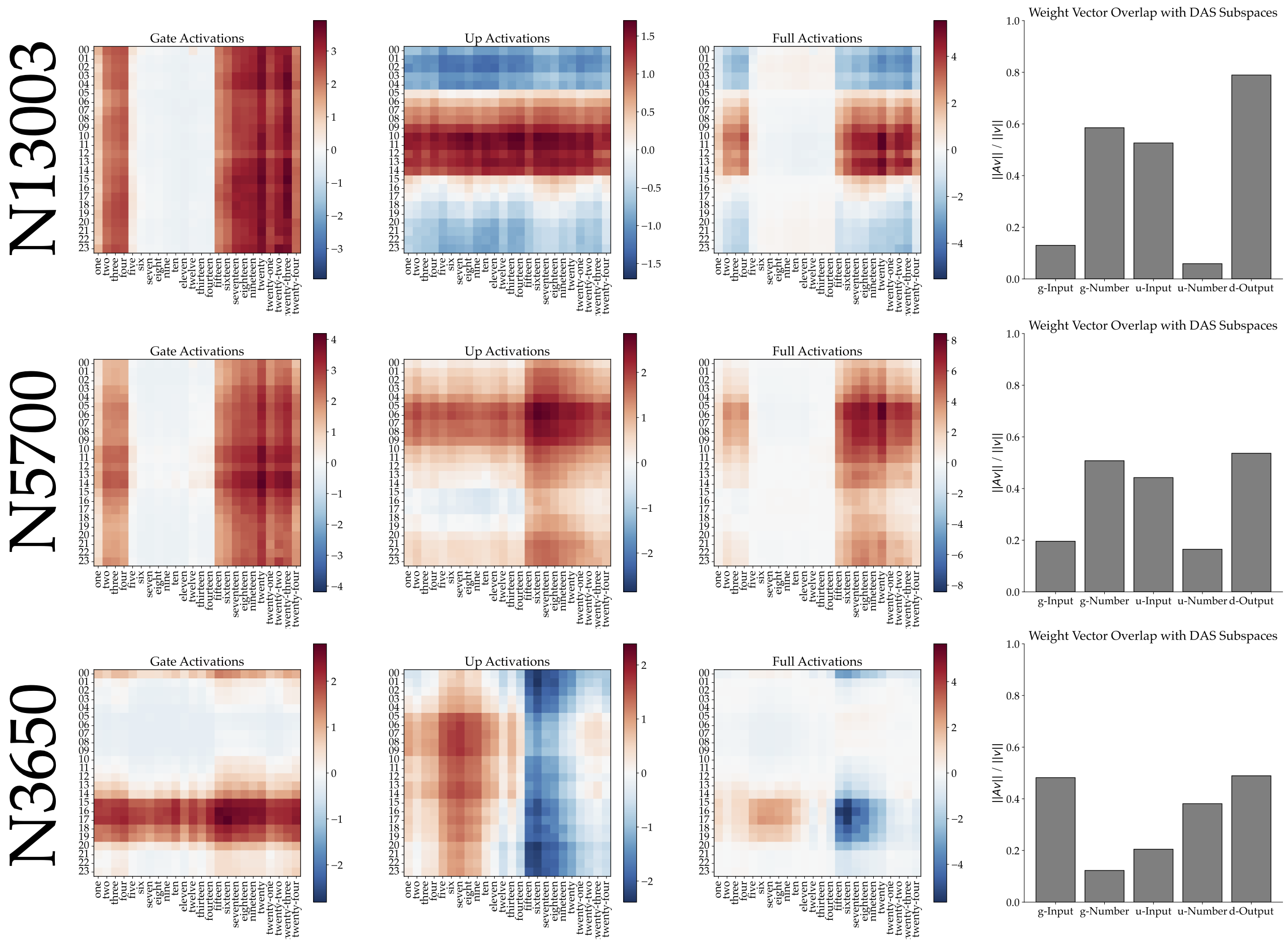}
    \caption{\textbf{All period 20 neurons}, activations for the \texttt{hours} task.}
    \label{fig:allmod20-hours}
\end{figure}

\begin{figure}
    \centering
    \includegraphics[width=0.92\linewidth]{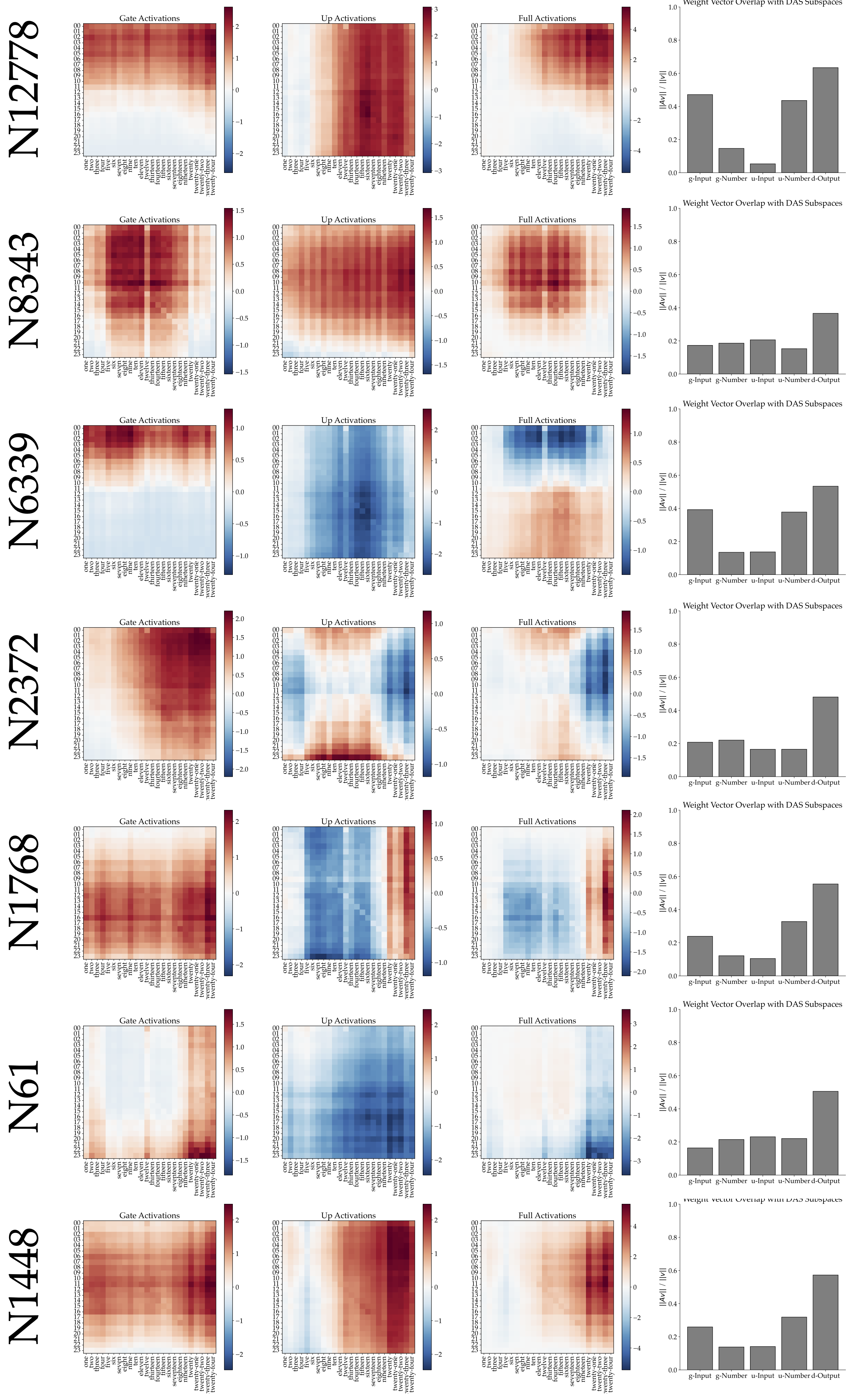}
    \caption{\textbf{All period 50 neurons}, activations for the \texttt{hours} task.}
    \label{fig:allmod50-hours}
\end{figure}

\begin{figure}
    \centering
    \includegraphics[width=\linewidth]{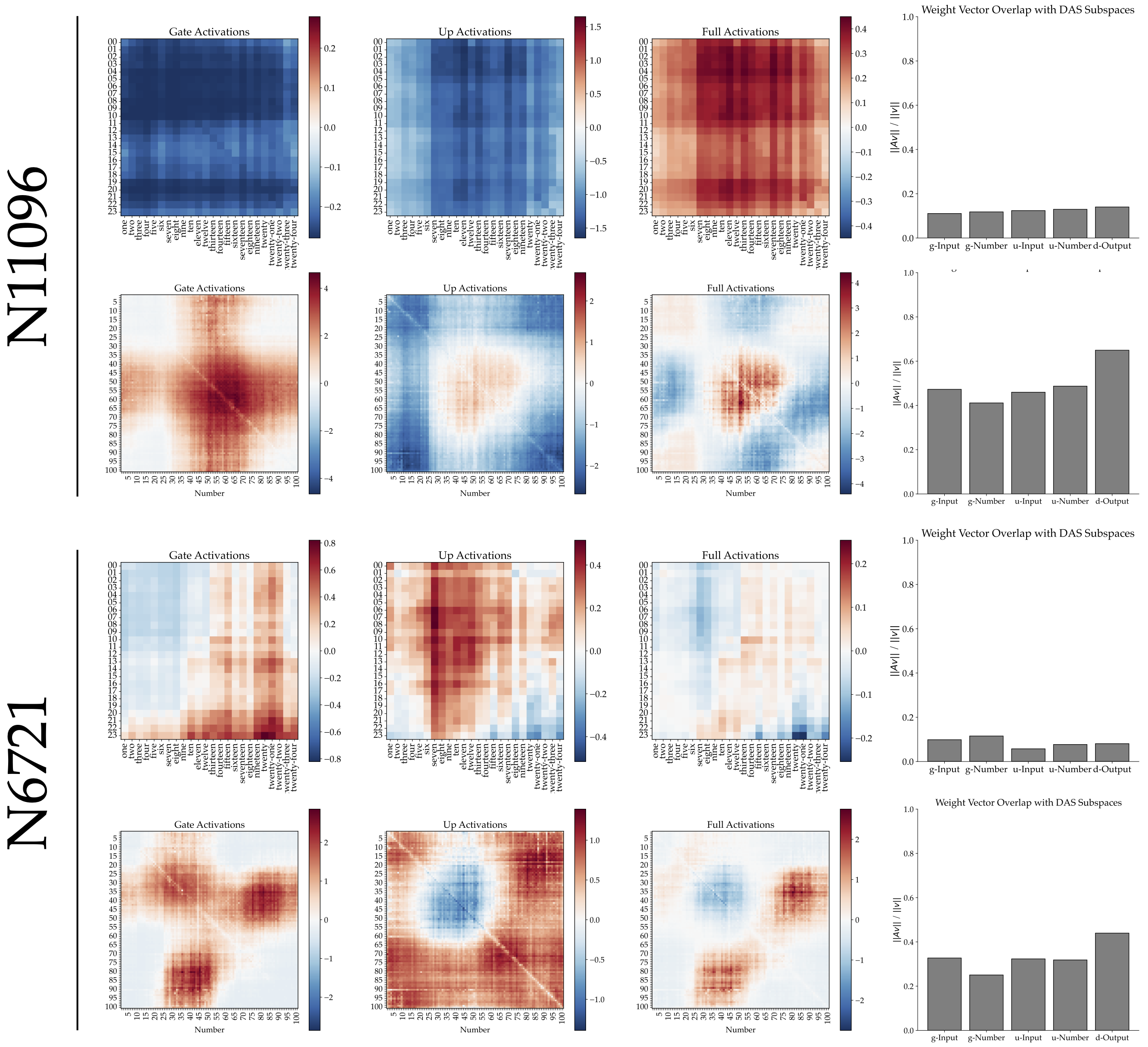}
    \caption{\textbf{All period 100 neurons}, activations for the \texttt{hours} task.}
    \label{fig:allmod100-hours}
\end{figure}

\clearpage
\subsection{Addition Neuron Down Projection Analysis}\label{app:all-neurons-downproj}

\begin{figure}[h]
    \centering
    \includegraphics[width=0.9\linewidth]{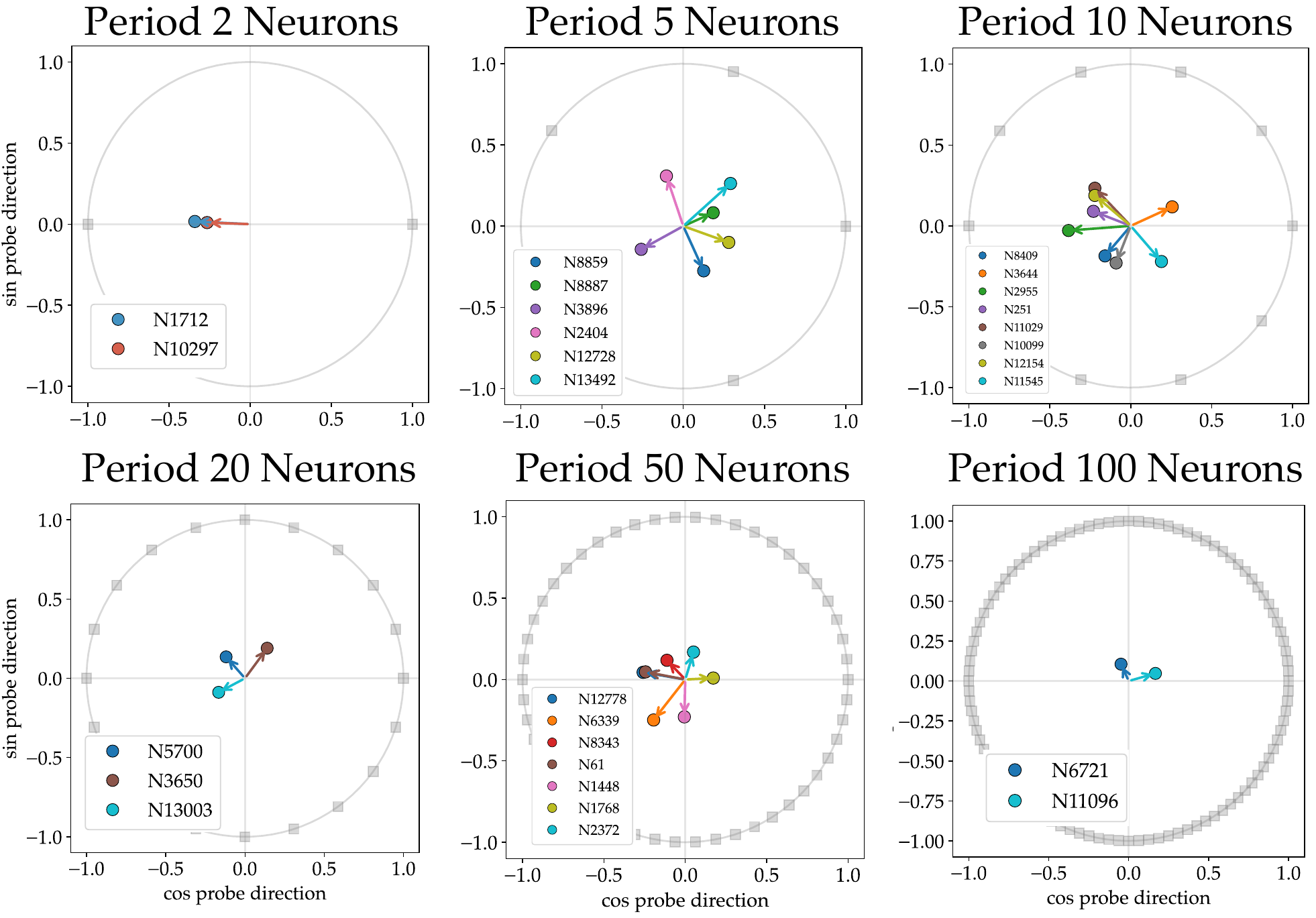}
    \caption{Down projection rows $\mathbf{d}_i$ for all addition neurons, projected onto the Fourier plane that corresponds to their activation period from Figure~\ref{fig:unclipped-ribbons}. We normalize probes before projecting. We do not orthogonalize cosine and sine probes for each period, as most are orthogonal (see Figure~\ref{fig:cosine}.}
    \label{fig:all-static-downprojs}
\end{figure}

\begin{figure}[h]
    \centering
    \includegraphics[width=0.9\linewidth]{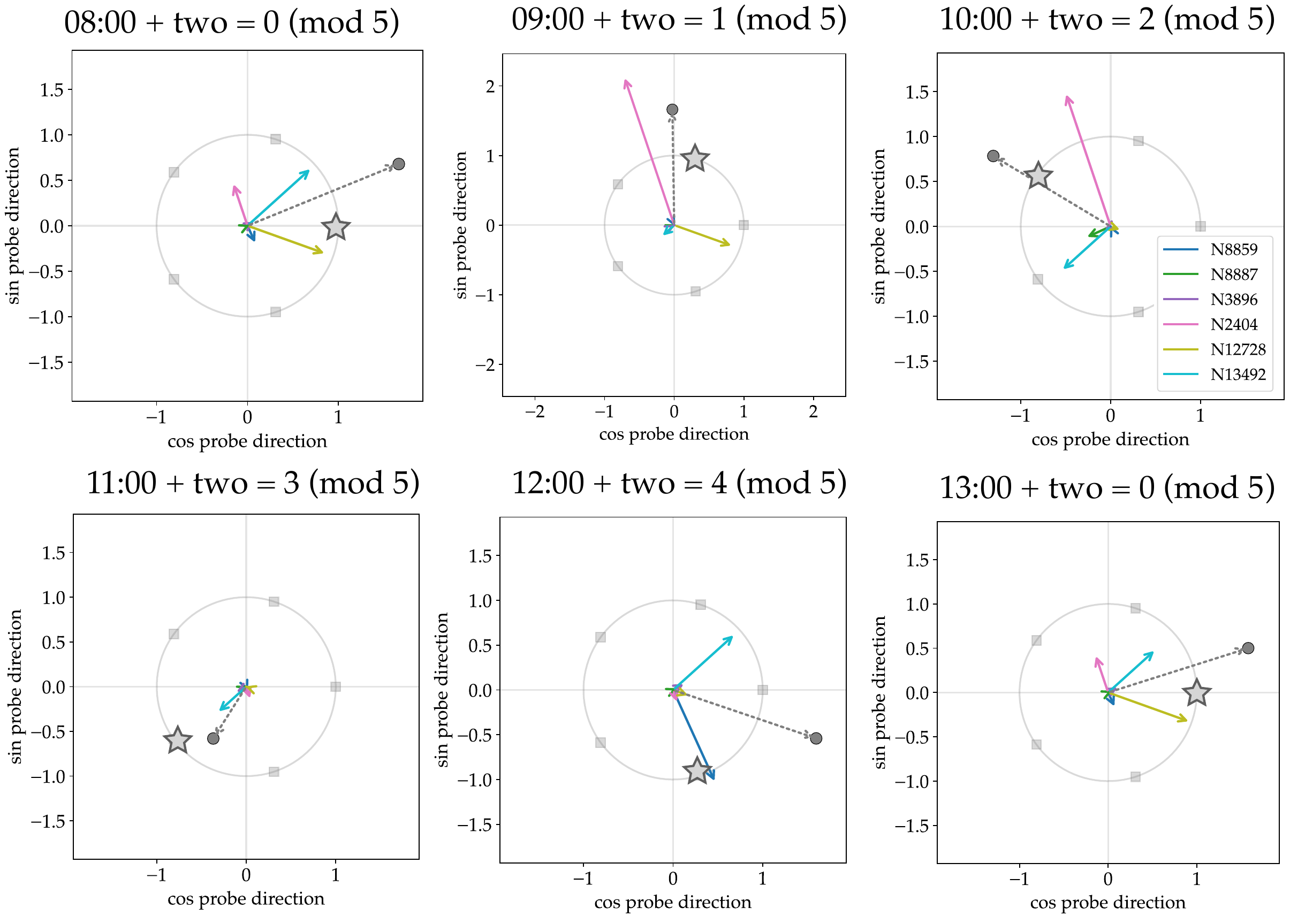}
    \caption{Period 5 neuron behavior when scrubbing across the \inputconcept\ for the \texttt{hours} task, holding \inputnumber\ constant. The gray dashed line indicates the sum of all the period five neuron activations, and the gray star indicates the ideal location this vector should point to. From left to right, the sum of our six neurons creates a vector with an angle roughly corresponding to the sum of \inputconcept\ + \inputnumber\ modulo 5. }
    \label{fig:scrub-hours}
\end{figure}

\begin{figure}[h]
    \centering
    \includegraphics[width=0.9\linewidth]{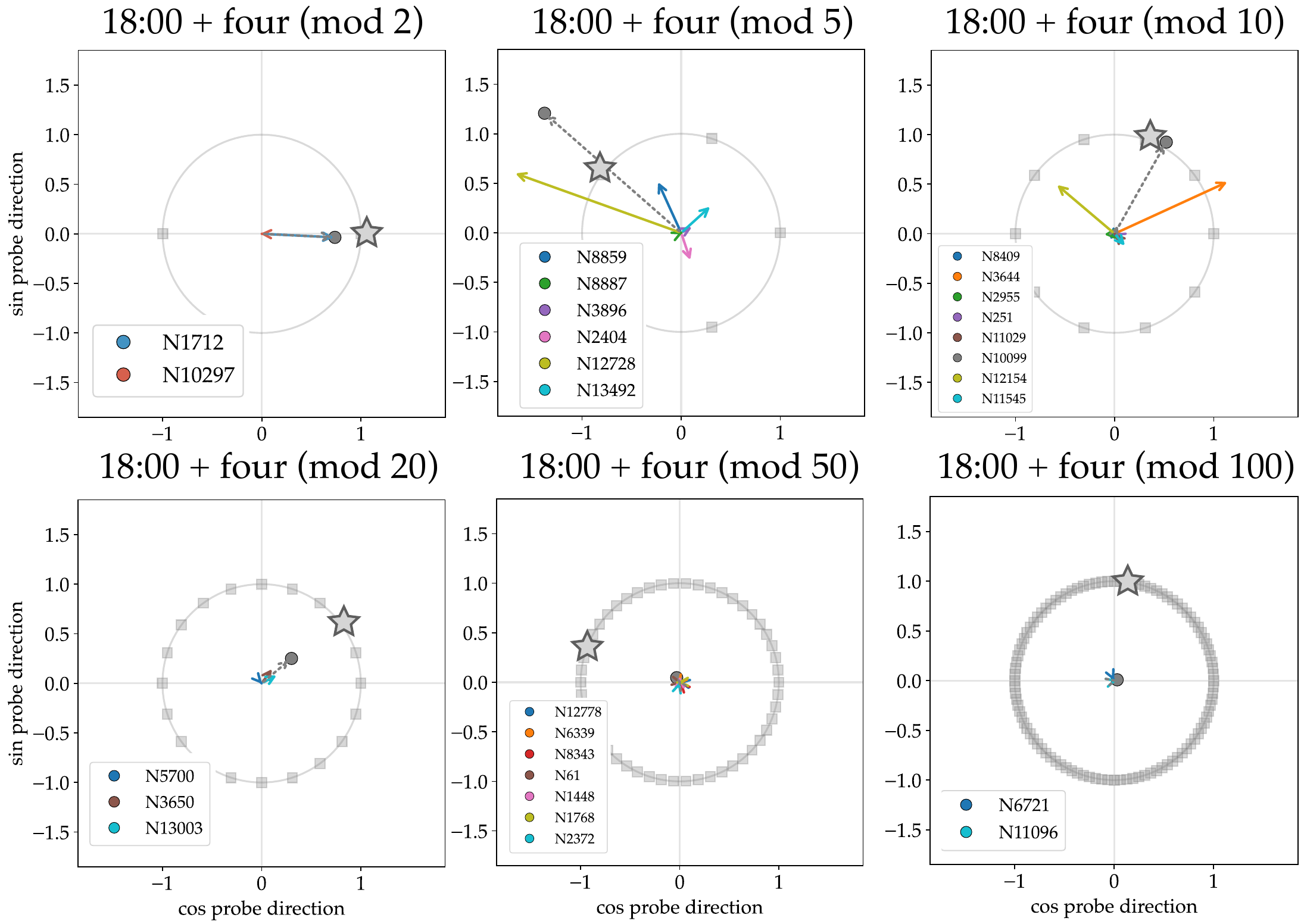}
    \caption{Neuron outputs for the \texttt{hours} prompt \textit{four hours after 18:00 = 22:00} projected onto the Fourier plane for each period. $T\in[2,5,10,20]$ activate strongly in the correct directions, while $T\in[50,100]$ activations are close to zero for this prompt.}
    \label{fig:18plusfour-allmods}
\end{figure}

\end{document}